\documentclass[10pt,twocolumn,letterpaper]{stanford}

\usepackage{bbm}
\usepackage[pagenumbers]{cvpr} 

\usepackage{tcolorbox}
\usepackage{longtable}
\definecolor{cvprblue}{rgb}{0.21,0.49,0.74}
\definecolor{stanfordblue}{HTML}{006CB8}
\definecolor{todo}{rgb}{0.9,0.1,0.1}
\definecolor{row_color}{HTML}{F3F8FC}
\definecolor{human_color}{rgb}{0.9,0.9,0.9}
\usepackage[pagebackref,breaklinks,colorlinks,allcolors=stanfordblue]{hyperref}

\usepackage{xcolor}
\usepackage{booktabs}
\usepackage{colortbl}
\usepackage{multirow}
\usepackage{caption}

\definecolor{QuantiDark}{RGB}{169,183,163}   
\definecolor{QuantiLight}{RGB}{227,232,224} 
\definecolor{QuantiDarkYellow}{RGB}{227,209,169}   
\definecolor{QuantiLightYellow}{RGB}{249,242,228} 
\definecolor{QuantiDarkRed}{RGB}{181,106,69}   
\definecolor{QuantiLightRed}{RGB}{239,225,216} 
\definecolor{QuantiDarkBlue}{RGB}{34,59,61}   
\definecolor{QuantiLighBlue}{RGB}{210,216,215} 

\newcommand{\best}[1]{\cellcolor{QuantiDark}\textbf{#1}}
\newcommand{\bestopen}[1]{\cellcolor{QuantiLight}\textbf{#1}}
\newcommand{\bestsub}[1]{\cellcolor{QuantiDarkYellow}\textbf{#1}}
\newcommand{\bestopensub}[1]{\cellcolor{QuantiLightYellow}\textbf{#1}}

\def\paperID{****} 
\def\confName{CVPR}
\def\confYear{2026}

\addtocontents{toc}{\protect\setcounter{tocdepth}{-1}}

\usepackage[draft]{minted}

\usepackage{graphicx}

\title{
    \raisebox{-0.0em}{\includegraphics[height=0.75em]{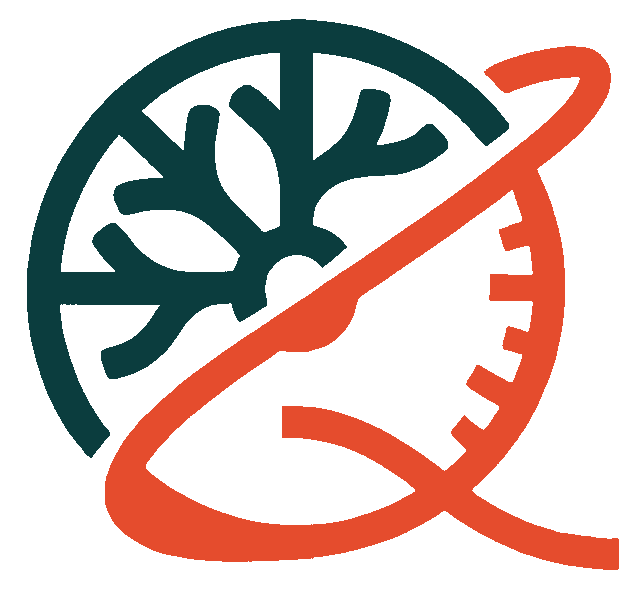}}
    \textsc{QuantiPhy}: A Quantitative Benchmark Evaluating Physical Reasoning Abilities of Vision-Language Models
}
\author[1,*]{\text{Li Puyin}}
\author[1,*]{\text{Tiange Xiang}}
\author[1,*]{\text{Ella Mao}}
\author[1]{\text{Shirley Wei}}
\author[1]{\text{Xinye Chen}}
\author[2]{\text{Adnan Masood}}
\authorbreak
\author[1,\dagger]{\text{Li Fei-Fei}}
\author[1,\dagger]{\text{Ehsan Adeli}}
\affiliation[1]{Stanford University}
\affiliation[2]{UST}
\contribution{\textsuperscript{*}Equal First Authorship$\quad$\textsuperscript{$\dagger$}Equal Last Authorship}

\dataset{\url{https://huggingface.co/datasets/PaulineLi/QuantiPhy-validation}}

\project{\url{https://quantiphy.stanford.edu/}}
\code{\url{https://github.com/Paulineli/QuantiPhy}}

\metadata[Correspondence to]{\texttt{\{puyinli, xtiange, eadeli\}@stanford.edu}}

\begin{document}

\abstract{Understanding the physical world is essential for generalist AI agents. However, it remains unclear whether state-of-the-art vision perception models (e.g., large VLMs) can reason physical properties quantitatively. Existing evaluations are predominantly VQA-based and qualitative, offering limited insight into whether these models can infer the kinematic quantities of moving objects from video observations. To address this, we present QuantiPhy, the first benchmark designed to quantitatively measure a VLM's physical reasoning ability. Comprising more than 3.3K video–text instances with numerical ground truth, QuantiPhy evaluates a VLM's performance on estimating an object's size, velocity, and acceleration at a given timestamp, using one of these properties as an input prior. The benchmark standardizes prompts and scoring to assess numerical accuracy, enabling fair comparisons across models. Our experiments on state-of-the-art VLMs reveal a consistent gap between their qualitative plausibility and actual numerical correctness. We further provide an in-depth analysis of key factors like background noise, counterfactual priors, and strategic prompting and find that state-of-the-art VLMs lean heavily on pre-trained world knowledge rather than faithfully using the provided visual and textual inputs as references when reasoning kinematic properties quantitatively. QuantiPhy offers the first rigorous, scalable testbed to move VLMs beyond mere verbal plausibility toward a numerically grounded physical understanding.}

\twocolumn[{%
    \renewcommand\twocolumn[1][]{#1}%
    \maketitle
    \begin{center}
        \centering
        \includegraphics[width=0.99\textwidth]{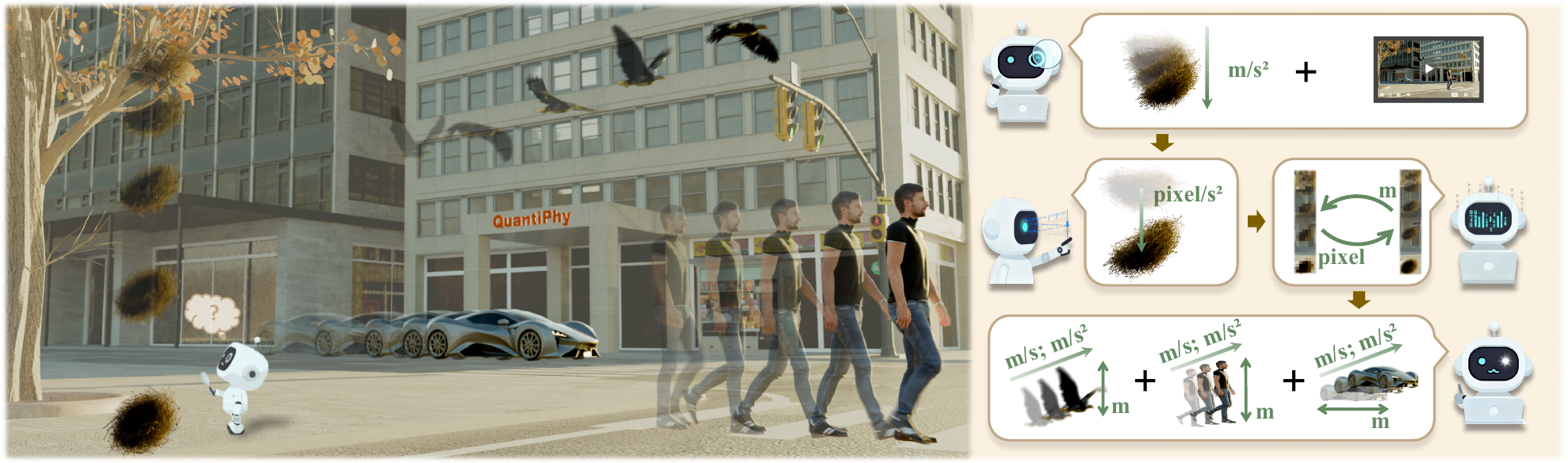}
        \captionof{figure}{On a crowded city street, a bird's nest falls from a branch, a car rushes by, an eagle flits over a building, and a person walks in a crosswalk — the real world is full of complex physical motion. To enable AI to understand and navigate this environment, it is essential for generalist embodied systems to reason about physical properties quantitatively. Because objects obey common laws of physics, their kinematic properties (such as size, velocity, and acceleration) are interrelated. This interdependence makes it possible for visual AI to systematically reason about these properties with respect to available priors. In this work, we present \textsc{QuantiPhy}, the first benchmark to evaluate the reasoning ability of AI models on quantitative kinematic inference tasks.}
        \label{fig:teaser}
    \end{center}
    \vspace{0.0cm}  
}]

\section{Introduction}
\label{sec:intro}
Understanding the physical world has long been a challenge for artificial intelligence. Humans inhabit a world governed by physical laws. From an apple falling from a tree to the trajectory of a thrown ball, we have developed mathematical tools to measure and calculate physical attributes quantitatively. This quantitative understanding forms the foundation for all modern scientific advances.

General visual intelligence systems, such as Vision-Language Models (VLMs), are developed in a manner that differs significantly from that of humans. They are trained to fit vast amounts of real-world data, which implicitly contain the abstract physical principles behind visual observations. Thus, assessing the numerical accuracy of VLMs’ reasoning about physical properties is a necessary next step. It is crucial for deploying applications such as embodied AI \cite{vuong2023open, driess2023palm}, AR/VR \cite{xiang2023wild2avatar, xiang2023rendering, grauman2022ego4d, xiang2025neuhmr, mangalam2023egoschema}, and autonomous driving \cite{tian2024drivevlm}.

Reasoning about quantitative physical properties from visual data can be challenging. For example, advanced methods like FoundationPose~\cite{wen2024foundationpose} require extensive prior information, including color, depth, object meshes, and camera parameters, to accurately localize an object in 3D space. Unfortunately, most of these priors are unavailable in ``in-the-wild'' captures. These challenges make it natural to ask whether large VLMs can leverage their rich implicit priors to reason end-to-end about precise kinematic and geometric properties. Moreover, a robust evaluation of VLM reasoning has the potential to help assess and improve the physical realism of videos created by generative models.

Studies of VLM physical understanding are not new. A variety of related benchmarks exist, spanning kinematics~\cite{yiCLEVRERCOLLISIONEVENTS2020} and dynamics~\cite{xu2025deepphy} to relationships~\cite{chowPhysBenchBenchmarkingEnhancing2025} and scene understanding~\cite{vsi}. However, almost all existing benchmarks are \textit{Visual Question Answering (VQA)-based and qualitative}. In this paradigm, ground-truth answers are effectively constrained by the prompt, and models are typically evaluated with multiple-choice questions. However, this VQA paradigm does not provide fine-grained evaluations of physical understanding. For example, if a model is asked to infer the size of a car from a video (ground truth: 3 meters), the incorrect answers 3.1 meters and 31 meters would be treated as equally wrong in a multiple-choice format. Quantitatively, however, the 31-meter answer is 10$\times$ worse. To truly push VLMs toward real-world applications, it is crucial to capture this numerical gap.

Our contributions in this work are four-fold:
(I) We propose a new quantitative paradigm for evaluating the physical reasoning ability of VLMs, moving beyond the limitations of qualitative VQA.
(II) We define a kinematic inference task for VLM physical reasoning that explicitly targets the challenge of understanding dynamics in videos: since kinematic properties of an object, such as size, velocity, and acceleration, are mutually correlated quantities, our task formalizes how visual agents could, in principle, transform a single physical prior into a family of numerically grounded predictions in real-world units that are useful for understanding and acting in the physical world.
(III) We present \textsc{QuantiPhy}, the first benchmark to systematically evaluate a VLM’s quantitative reasoning on object kinematic properties in videos, spanning 2D/3D motion, static and dynamic priors, and diverse scene conditions, together with a standardized metric, prompting protocol, and leaderboard over 21 state-of-the-art models. (IV) We provide a detailed analysis of factors affecting VLM reasoning, including scene complexity, video availability, counterfactual priors, and chain-of-thought prompting. Summarizing the analysis, we conclude with one interesting finding: \emph{when estimating kinematic quantities, existing VLMs hallucinate by relying heavily on pre-trained world knowledge while hardly \textbf{inferring from} the actual reference video and text.}
\section{Related Work}
\label{sec:related_work}

\begin{figure*}[t]
    \centering
    \includegraphics[width=0.99\linewidth]{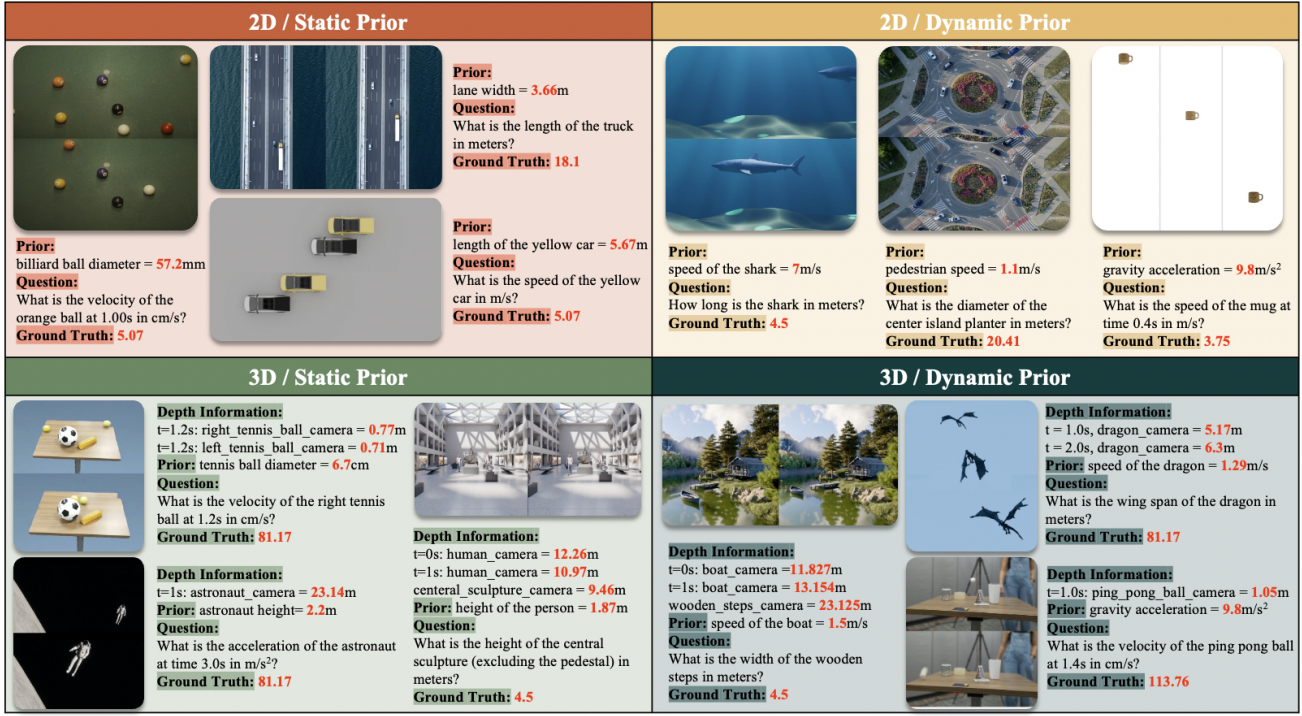}
    \caption{Sample examples from \textsc{QuantiPhy}, illustrating the four core task combinations defined by \texttt{Dimensionality} \{\texttt{2D, 3D}\} and \texttt{Physical Prior} \{\texttt{Static, Dynamic}\}, as described in \sectionautorefname~\ref{sec:overview}. Our collected data is diverse in nature, and each video is paired with multiple (prior, question, ground truth) triplets. Please see the supplementary materials for more data examples.}
    \label{fig:data}
\end{figure*}

\subsection{Benchmarks for VLM Physical Understanding}
For over a decade, researchers have recognized the need for AI to comprehend real-world physics. A series of benchmarks and evaluation protocols has been developed to assess the qualitative physical reasoning abilities of VLMs. Early efforts primarily focused on evaluating a model’s capacity to describe, predict, and explain basic physical events, such as collisions, falling, and rebounding in controlled synthetic environments~\cite{yiCLEVRERCOLLISIONEVENTS2020,bakhtinPHYRENewBenchmark2019}. Subsequent works expanded this scope to include reasoning about inherent physical properties of objects, such as mass, friction, elasticity, and deformability~\cite{tungPhysionEvaluatingPhysical}. More recently, comprehensive benchmarks like PhysBench~\cite{chowPhysBenchBenchmarkingEnhancing2025} and STAR~\cite{wu2024star} have been proposed to assess broader aspects of physical reasoning. These include object relationships, physical scene understanding, complex physics-based dynamics, and physics-situated actions in more realistic or diverse environments~\cite{grunde2021agqa,zheng2024contphy}, aiming to simulate the real-world reasoning demands that embodied AI systems will face.

However, despite these advances, the vast majority of existing benchmarks adopt the VQA framework~\cite{antol2015vqa,xiao2021next,tapaswi2016movieqa,yi2018neural}. In this framework, models are asked to provide multiple-choice selections or descriptive explanations of physical events. One relevant work, VSI-Bench~\cite{vsi}, provides preliminary studies on basic spatial understanding using numerical metrics, but is limited to static objects, focusing more on the models' perceptual ability than on emergent reasoning. Following that, Super-VSI~\cite{yang2025cambrian} was recently proposed, demonstrating that VLMs empowered with the ability for numerical spatial understanding have strong potential in empowering embodied AI. Therefore, a need still exists for benchmarking VLM’s ability to perform quantitative reasoning over the geometric and kinematic properties of moving objects (e.g., estimating size, velocity, or acceleration in metric terms).

\subsection{Physical Reasoning Models} Although AI models have achieved impressive results on general tasks, their ability to reason about the physical world remains limited. For VLMs, state-of-the-art models such as ChatGPT-4o and Gemini-1.5 Pro achieve only around 60\% accuracy on the PhysBench benchmark, far below human-level performance ($\sim$95\%)~\cite{chowPhysBenchBenchmarkingEnhancing2025}, indicating persistent challenges in physical reasoning. Generative models also struggle. Recent work shows that generated videos often violate basic physical laws~\cite{bansal2025videophy}, as these models are more sensitive to low-level visual cues (e.g., color, shape) than to underlying physical properties~\cite{kangHowFarVideo2024}. In embodied AI, recent systems have demonstrated basic physical reasoning abilities. They can interact with the environment~\cite{azzolini2025cosmos}, but their reasoning remains largely qualitative, lacking a precise and quantitative understanding of physical dynamics. These gaps underscore the need for benchmarks and models that transcend qualitative judgment and move toward accurate quantitative reasoning in physical contexts.

\subsection{Physical Understanding and Reasoning} In computer vision, techniques such as optical flow~\cite{horn1981determining,beauchemin1995computation}, combined with object detection and tracking models like YOLO~\cite{redmon2016you} and ByteTrack~\cite{zhang2022bytetrack}, enable accurate analysis of object motion and are widely used in broad applications~\cite{giachetti2002use,sangsuwan2024video}. These methods demonstrate that quantitative information (e.g., velocity, displacement) can be reliably inferred from video inputs. Meanwhile, research in cognitive science shows that humans possess strong abilities to reason about scale, from microscopic to astronomical, by leveraging relational cues and prior knowledge~\cite{resnick2017using}. Even in visually unfamiliar or simulated environments, people can extract physical rules and perform causal and predictive reasoning about object motion and interactions~\cite{gerstenberg2021counterfactual,allen2020rapid,li2022learning}. This highlights the potential for robust physical reasoning ability to emerge in large AI models from visual inputs, a crucial step toward grounded quantitative understanding.

\section{Methods}

We introduce \textbf{\textsc{QuantiPhy}}, the first benchmark designed to \emph{quantitatively} evaluate VLMs' physical reasoning ability on moving objects. In \autoref{sec:overview}, we first define the basic object kinematic properties to be assessed in \textsc{QuantiPhy}, and present key statistics of it. Then, \autoref{sec:construction} provides details on the construction of \textsc{QuantiPhy}, focusing on data collection.

\subsection{Overview of \textsc{QuantiPhy}}
\label{sec:overview}

\noindent\textbf{The task of kinematic inference.}
The purpose of \textsc{QuantiPhy} is to evaluate whether VLMs can utilize prior knowledge to reason about objects' kinematic properties with numerical accuracy, which forms the foundation for a more sophisticated understanding of physics. Specifically, we focus on the \emph{translational movements} of various objects.\footnote{We do not include rotational movements in this work; we defer this to future work and provide a discussion in \autoref{sec:discussion}.}

\begin{tcolorbox}
Given a video, we provide the VLM with a single physical prior for a source object (from the set $\{\mathbf{S}^{\mathrm{world}}, \mathbf{V}^{\mathrm{world}}_{t}, \mathbf{A}^{\mathrm{world}}_{t}\}$, in real-world units) as textual input. The model is then prompted to \emph{quantitatively determine} requested kinematic properties for a target object (which may be the same as or different from the source object) in world space.
\end{tcolorbox}

We distinguish \emph{pixel space}, where quantities are measured in pixels ([pixel], [pixel/s], [pixel/s$^2$]), from \emph{world space}, where they are expressed in physical units (e.g., [m], [m/s], [m/s$^2$]). Consider a video capturing the translational movement of an object with a fixed camera. At any time $t$, the object's location in pixel space, $\mathbf{X}^{\mathrm{pixel}}_{t}$, can be obtained from the frames. From the discrete trajectory, we compute pixel-space velocity and acceleration via finite differences:
\[
\resizebox{\columnwidth}{!}{$
\mathbf{V}^{pixel}_{t}\approx\frac{\mathbf{X}^{\mathrm{pixel}}_{t+\mathrm{d}t}-\mathbf{X}^{\mathrm{pixel}}_{t}}{\mathrm{d}t};\;
\mathbf{A}^{pixel}_{t}\approx\frac{\mathbf{X}^{\mathrm{pixel}}_{t+2\mathrm{d}t}-2\mathbf{X}^{\mathrm{pixel}}_{t+\mathrm{d}t}+\mathbf{X}^{\mathrm{pixel}}_{t}}{\mathrm{d}t^{2}}
$}.
\]

The video thus defines the kinematics only in pixel units. To relate them to world space, we assume an unknown scalar scale factor $\gamma>0$ (with units [world length/pixel]) such that, along the motion direction, $\mathbf S^{\mathrm{world}} = \gamma\mathbf S^{\mathrm{pixel}}$, $\mathbf{V}^{\mathrm{world}}_{t} = \gamma\mathbf{V}^{\mathrm{pixel}}_{t}$, $
\mathbf{A}^{\mathrm{world}}_{t} = \gamma\,\mathbf{A}^{\mathrm{pixel}}_{t}$, where $\mathbf S^{\mathrm{pixel}}$ is an object size measured in pixels and $\mathbf S^{\mathrm{world}}$ the same size in a physical unit. When a single prior in world space is provided (object size $\mathbf S^{\mathrm{world}}$, velocity $\mathbf{V}^{\mathrm{world}}_{t}$, or acceleration $\mathbf{A}^{\mathrm{world}}_{t}$ at some time $t$), together with the corresponding pixel-space quantity from the video, $\gamma$ is determined, and any other kinematic property in world space follows by rescaling its pixel-space counterpart.

See \autoref{fig:data} for examples of the kinematic tasks included in the benchmark. The performance of VLMs is measured by the numerical error between the VLM's prediction and the annotated ground truth.

\noindent\textbf{Benchmark setup.}
For a comprehensive evaluation of the kinematic movements above, \textsc{QuantiPhy} is designed to include video-question pairs categorized along three primary axes. We first describe the two axes that define the core reasoning task:

\noindent\ $\bullet$\ \ \texttt{Dimensionality}: \{\texttt{2D}, \texttt{3D}\}. \texttt{2D} movement assumes the object moves strictly in the x-y plane with no change in depth relative to the camera. \texttt{3D} movement includes the z-axis, resulting in varying depth, which is intrinsically more challenging\footnote{We provide additional depth prior on \texttt{3D} scenes.}. 

\noindent\ $\bullet$\ \ \texttt{Physical prior}: \{\texttt{Static}, \texttt{Dynamic}\}. The \texttt{Static} prior indicates the provision of the object size $\mathbf{S}^{\mathrm{world}}$, which is constant throughout the video. While \texttt{Dynamic} prior indicates velocity $\mathbf{V}^{\mathrm{world}}_{t}$ or acceleration $\mathbf{A}^{\mathrm{world}}_{t}$ at a given timestamp $t$.

Together, these two axes divide the Benchmark into four task categories: \texttt{2D-Static}, \texttt{2D-Dynamic}, \texttt{3D-Static}, and \texttt{3D-Dynamic}.\footnote{In the graphs and tables, we denote the four kinematic inference task categories as 2S (\texttt{2D-Static}), 2D (\texttt{2D-Dynamic}), 3S (\texttt{3D-Static}), and 3D (\texttt{3D-Dynamic}) for short.}

\begin{figure}[t]
    \centering
    \includegraphics[width=1.0\linewidth]{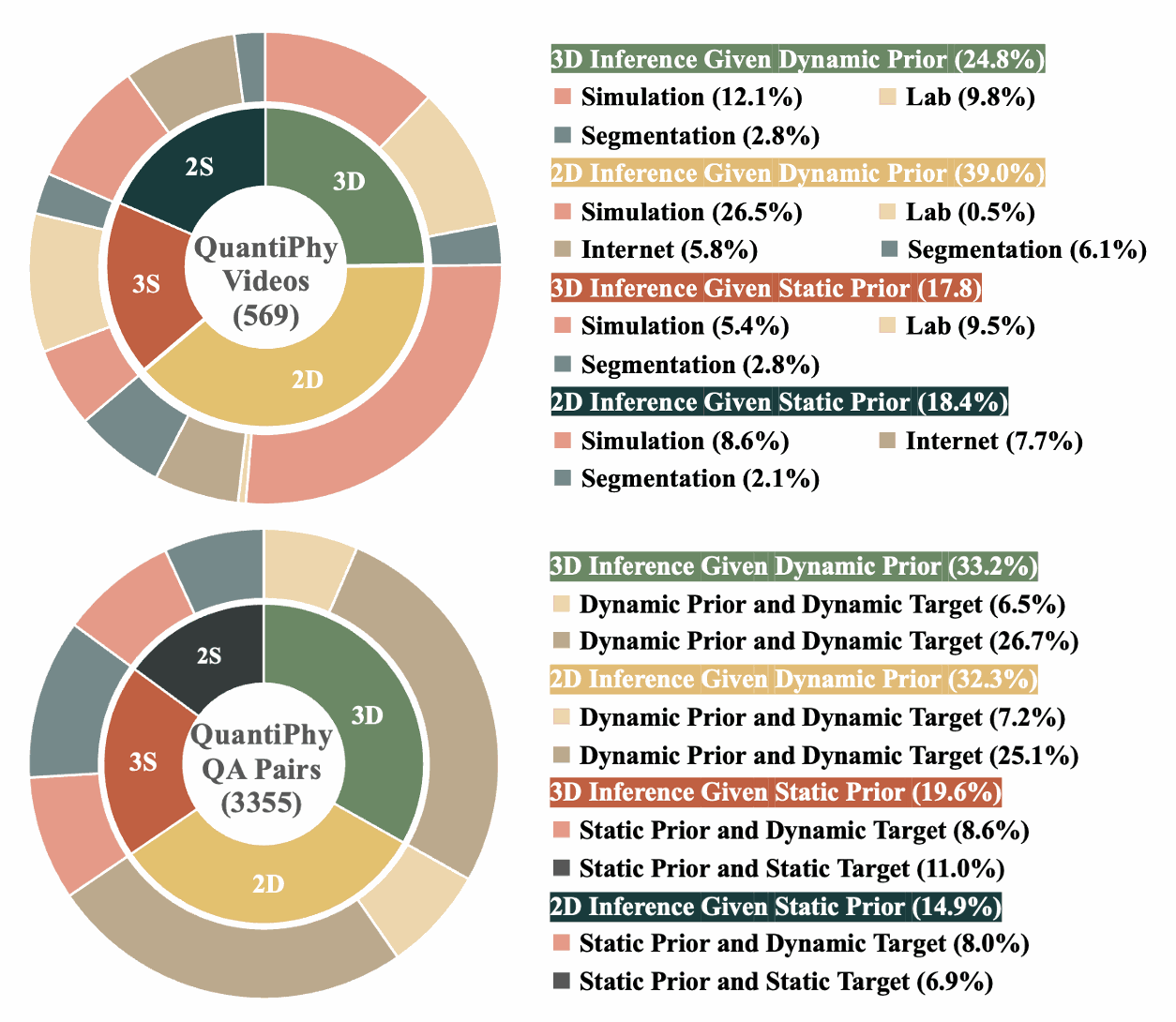}
    \vspace{-0.5em}
    \caption{\textbf{\textsc{QuantiPhy} Statistics.} The collected data and curated QA pairs are among four main setups with further breakdowns.}
    \vspace{-0.5em}
    \label{fig:statistics}
\end{figure}

\noindent\textbf{Data statistics.}
Similar to~\cite{chowPhysBenchBenchmarkingEnhancing2025,vsi}, data in \textsc{QuantiPhy} is organized into triplets of (question/prompt, video, numerical ground truth). Each video may be paired with multiple questions and corresponding ground truth annotations. To ensure diversity, the four data setups were collected from various sources, yielding a total of 569 unique videos and 3355 questions. With proper post-processing, our collected videos typically have a duration of 2-3 seconds, occupying approximately 115MB of disk storage, making the benchmark suitable for use across various hardware settings. See \autoref{fig:statistics} for a detailed breakdown of the data statistics.

\begin{figure*}[ht!]
    \centering
    \includegraphics[width=0.99\linewidth]{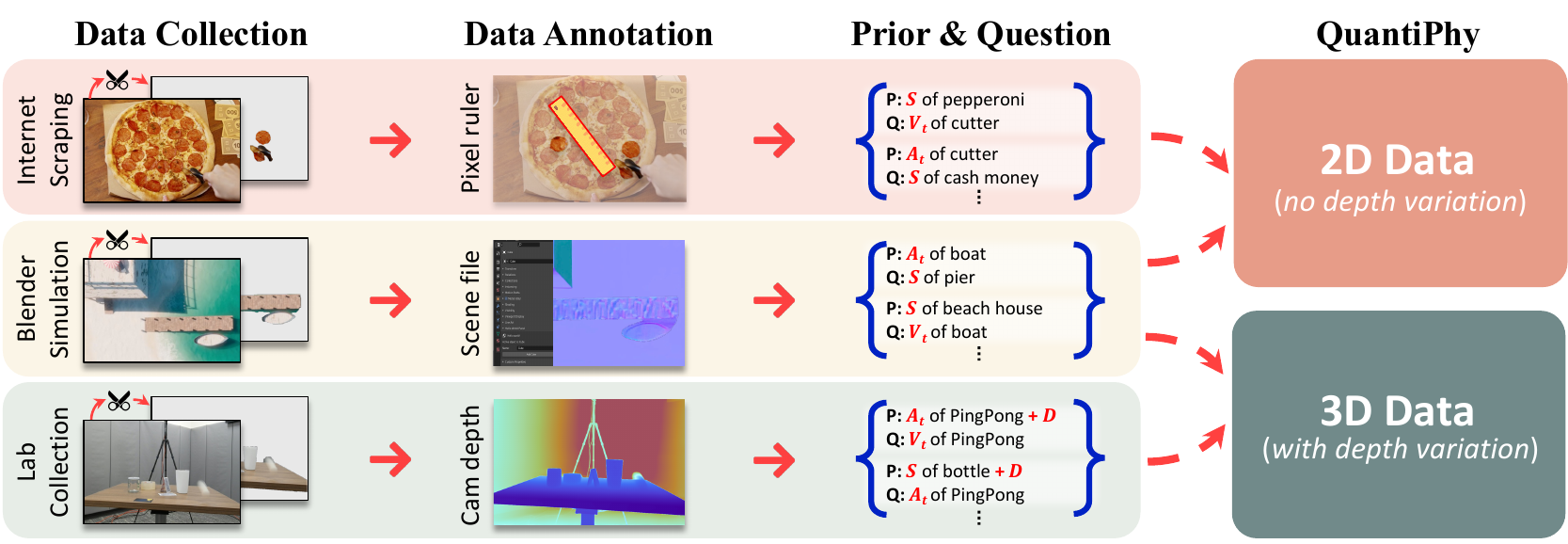}
    \caption{The construction of \textsc{QuantiPhy} proceeds in three sequential stages. First, we collect diverse raw videos from three different sources. Additionally, we segment these videos with solid plain background (described in \autoref{sec:additionalseg}). Second, we obtain high-quality annotations, employing distinct labeling methods tailored to each data source to accurately capture the object's physical properties. Finally, we formulate the benchmark tasks by associating each video with multiple (prior, question, ground truth) triplets. Each triplet is then categorized as either \texttt{2D} or \texttt{3D}, depending on the object's movement relative to the camera.}
    \label{fig:pipeline}
\end{figure*}

\subsection{Benchmark Construction}
\label{sec:construction}
In \textsc{QuantiPhy}, we constructed a large-scale dataset containing diverse video data from multiple sources.

\noindent\textbf{Data collection.} Our data combines simulated and real-world sources for controllability and practical applicability. The data collection process is visualized in \autoref{fig:pipeline}.

\noindent\ $\bullet$\ \ \texttt{Blender simulation}: Blender allows us to render scenes that are both visually realistic and physically plausible, encompassing \texttt{2D} and \texttt{3D} motion. Simulating object movements in Blender provides complete control over the environment and guarantees precise ground-truth annotations. This control allows us to probe VLM's reasoning capabilities by placing objects with different motion types within a single scene. It also enables us to study model robustness by holding motions fixed while systematically varying the background and visual noise. To enhance data diversity, we include simulations of scenes that are difficult or impossible to record in the real world, ranging from microscopic scales (e.g., red blood cells in vessels) to astronomical scales (e.g., galaxy movements).

\noindent\ $\bullet$\ \ \texttt{Lab capturing}: Reconstructing objects in 4D (3D space + time) lets us annotate real-world object motion using a multi-view stereo setup. Our captures feature a wide diversity of movements, including free fall, sliding down slopes, pendulum motion, and bouncing. We also varied the physical properties of the objects, including those that are rollable and deformable. Further details on our setup can be found in the supplementary materials.

\noindent\ $\bullet$\ \ \texttt{Internet scraping}: To extend our dataset to more out-of-distribution scenarios, we leverage the internet as a source for in-the-wild data. However, not all internet data is suitable for our benchmark, as our benchmark specifically requires consistent coverage of moving objects captured by a relatively static camera, along with reference objects of known dimensions (e.g., a standard coin) to serve as priors. Hence, we narrowed our scraping scope to websites hosting high-quality monocular videos and then manually inspected them to select those meeting our stringent requirements.

We provide a detailed description of the data collection process in the supplementary materials \autoref{appendix:data_collection}.

\label{sec:additionalseg}
\noindent\textbf{Additional segmented data.}
We apply SAM~\cite{ravi2024sam2} to segment moving objects against solid backgrounds, doubling our dataset without additional annotation while enabling controlled analysis of background complexity effects.

\noindent\textbf{Data annotation and question design.}
Blender simulations provide direct extraction of sizes, velocities, and accelerations via automation scripts. We construct Blender scenes using open-source 3D assets from Sketchfab~\cite{Sketchfab} and BlenderKit. Automation scripts were developed for each scene to extract accurate, time-specific physical properties for any given object.

However, data annotation is non-trivial for the other two sources. For lab captures, we utilize metric depth directly from depth cameras combined with multi-view stereo to achieve a full 4D reconstruction of the scene. We then apply per-pixel segmentation masks to outline the objects of interest in each video. For each video, we select a primary camera and use the object's metric depth from that viewpoint as the depth prior provided to the VLM. Object movements are then computed in world coordinates. For internet data, since the videos are monocular and lack multi-view information, we manually annotate sizes and displacements in pixel space and use reference objects with known priors to obtain the mapping to world scale. 

We provide a detailed description of the data annotation process in the supplementary materials \autoref{appendix:data_annotation}.

\noindent\textbf{Prompt design.} As mentioned, the input to the VLM consists of a video paired with a single physical prior. For simplicity and effective integration, this prior is provided as a textual description (see \autoref{fig:data} for examples). In addition, we include textual cues in the prompt, such as `analyze the video and calculate the answer carefully'. We also add output constraints, instructing the model to `output only the numerical answer and unit' in its final response.

We provide a detailed description of the prompt design process in the supplementary materials \autoref{appendix:prompt_design}.

\section{Evaluation on \textsc{QuantiPhy}}
\label{sec:evaluation}
\subsection{Evaluation Setup}

\noindent\textbf{Benchmark models.} With different architecture designs, training data and protocals, different VLMs may have varying ability of physical understanding from videos. For better comprehensiveness, we include the evaluations of a total number of 21 state-of-the-art VLMs and variants in this work, consisting of 6 proprietary models include ChatGPT-5.1~\cite{OpenAIgpt5.1}, ChatGPT-5~\cite{OpenAIgpt5}, Gemini-2.5 Flash~\cite{google2025gemini25flash}, Gemini-2.5 Pro~\cite{google2025gemini25pro}, Grok-4.1~\cite{grok41}, and Claude-4.5 Sonnet~\cite{anthropic2025claude45sonnet} and 15 open-sourced models (as listed in \autoref{tab:main-results}).
The code supports multiple providers (OpenAI, Gemini, xAI, Anthropic, and Replicate API) with provider-specific parameters. Temperature is typically 0–0.1 for deterministic outputs. Token limits vary: OpenAI models use up to 10,000 tokens due to longer thinking steps, and open-sourced models with around 500–2,048.

\noindent\textbf{Human studies.} 
To complement our model evaluation and establish a reference point for human-level performance on quantitative physical reasoning, we conducted a survey study. Participants watched 18 videos and answered 1–3 quantitative kinematic questions using the same priors and task definitions as our VLM evaluation.

Importantly, humans and VLMs receive fundamentally different forms of input. VLMs operate directly on pixel-accurate video tensors, while human participants rely on visual perception, intuition, and coarse approximations. 

\noindent\textbf{Metric design.}
All tasks in \textsc{QuantiPhy} require models to output numerical values. Following VSI-Bench~\cite{vsi}, we use \emph{Mean Relative Accuracy} (MRA) as the primary metric to measure the proximity between model predictions and ground-truth answers. While accuracy based on exact matching is a simple baseline but too brittle for continuous, noisy physical measurements in \textsc{QuantiPhy}, MRA evaluates whether the \emph{relative error} falls below a set of tolerance thresholds and averages the resulting accuracies, offering a more calibrated and robust notion of when models are “accurate enough’’ for physically grounded reasoning (see \autoref{appendix:metric_design} in the supplementary material for further discussion).

Concretely, we consider a set of confidence thresholds $\mathcal{C} = \{0.1, 0.2, \ldots, 0.9, 0.95\}$, and define MRA for a prediction $\hat{y}$ with ground truth $y$ as
\[
\mathrm{MRA} = \frac{1}{10} \sum_{\theta \in \mathcal{C}}
\mathbbm{1} \left( \frac{|\hat{y} - y|}{|y|} < 1 - \theta \right),
\]
where $\mathbbm{1}(\cdot)$ is the indicator function. Intuitively, larger $\theta$ corresponds to a stricter tolerance $1-\theta$ on the relative error, and MRA averages accuracy across a spectrum of such tolerances, providing a more informative measure than a single-threshold accuracy.

Our benchmark organizes questions into four kinematic categories, \texttt{2D-Static}, \texttt{2D-Dynamic}, \texttt{3D-Static}, and \texttt{3D-Dynamic}. For each model and each category, we compute the category-level score by averaging the question-level MRA over all questions in that category for which the model produces a \emph{valid numerical answer}. The overall score of a model is then obtained as the unweighted mean of its four category-level scores.

A practical challenge is that not all VLMs consistently output sensible numerical predictions for our reasoning tasks. For every video–question pair, we query a model up to five times with the same prompt, stopping early if any response contains a parseable numerical value. If none of the five responses yields a valid number, we regard the model as \emph{failing to answer} that question.

\subsection{Main Results}

\autoref{tab:main-results} summarizes performance on \textsc{QuantiPhy} across four task types (\texttt{2D-Static}, \texttt{2D-Dynamic}, \texttt{3D-Static}, and \texttt{3D-Dynamic}).
Overall, we find that quantitative kinematic inference remains challenging for current VLMs: even the best systems do not yet reach human performance, despite in principle having access to more precise information.

\noindent\textbf{Human baseline.}
Human annotators achieve an average MRA of $55.6$ across all categories, with scores between $50.0$ (\texttt{2D-Static}) and $59.1$ (\texttt{2D-Dynamic}).
 This range is consistent with the fact that humans do not have direct access to pixel-level measurements and must instead rely on coarse visual estimation (e.g., counting grid lines or comparing to familiar objects), but still demonstrates that the tasks are solvable with reasonable accuracy.

\noindent\textbf{Proprietary models.}
ChatGPT-5.1 attains the highest overall score with $53.1$ MRA, followed by Gemini-2.5-Pro at $49.6$.
ChatGPT-5.1 slightly surpasses humans on the \texttt{2D-Dynamic} category, yet none of them surpass the human average.
Other closed models such as GPT-5 and Claude Sonnet~4.5 are substantially weaker, with overall MRA around $32.6$ and $22.8$, respectively.

\noindent\textbf{Open-weight models.}
Open-weight models exhibit a wide performance spread.
The best open-weight system, Qwen3-VL-Instruct-32B, reaches $46.0$ overall MRA, with strong \texttt{2D-Dynamic} and \texttt{3D-Dynamic} scores, followed by InternVL-3.5-30B ($40.7$) and Qwen3-VL-Instruct-8B ($38.8$).
These models are clearly below the top proprietary ones, but already comparable to mid-tier closed models.
Smaller open-weight models such as Phi-4 Multimodal and SmolVLM-256M still achieve non-trivial MRA, yet remain far from both human and large-model performance.

\noindent\textbf{Scaling effects.}
To better understand the role of model scale, we compare models within the same family across different parameter sizes.
Within the Qwen3-VL family, average MRA increases from $29.0$ at 2B parameters to $38.8$ at 8B and further to $46.0$ at 32B. A similar trend is observed for InternVL, where InternVL-3.5-30B ($40.7$) substantially outperforms its 8B ($35.4$) and 2B ($25.0$) variants. Notably, scaling benefits are most pronounced on dynamic categories (\texttt{2D-Dynamic} and \texttt{3D-Dynamic}), suggesting that larger models are better able to integrate temporal information for quantitative inference. However, these gains exhibit diminishing returns and do not close the gap to either top proprietary models or human performance, indicating that scale alone is insufficient for faithful physical reasoning.

\noindent\textbf{Gap to super-human performance.} By collecting results from both humans and VLMs, we observe an interesting pattern. Importantly, human performance does not represent the theoretical ceiling for \textsc{QuantiPhy}. While humans must rely on coarse visual approximations, an ideal agent with precise frame-level access to pixel coordinates could recover the world--pixel scale and compute target quantities exactly. In theory, VLMs should be capable of significantly outperforming humans by leveraging this exact pixel information to perform precise algebraic operations. The fact that the best current systems cluster around $50\%$ MRA, comparable to or below human baselines, suggests they still fundamentally under-utilize visual precision and physical priors, leaving robust and accurate quantitative reasoning for kinematic inference tasks as an open challenge.

\begin{table*}[t]
\centering
\small
\setlength{\tabcolsep}{4pt}

\begin{minipage}[t]{0.61\textwidth}
\centering
\resizebox{\linewidth}{!}{
\begin{tabular}{
r|
>{\centering\arraybackslash}p{1.2cm}|
>{\centering\arraybackslash}p{1.2cm}
>{\centering\arraybackslash}p{1.2cm}
>{\centering\arraybackslash}p{1.2cm}
>{\centering\arraybackslash}p{1.2cm}|
>{\centering\arraybackslash}p{1.2cm}
}
\toprule
\multicolumn{1}{c}{\multirow{2}{*}{Models}}  & \multicolumn{1}{c}{\multirow{2}{*}{Size}}   &
\multicolumn{4}{|c|}{\textbf{Kinematic Categories}} & \multicolumn{1}{c}{\multirow{2}{*}{Avg.}} \\
 \multicolumn{1}{c}{}&  \multicolumn{1}{c}{} &  \multicolumn{1}{|c}{2S}  & \multicolumn{1}{c}{2D}  & \multicolumn{1}{c}{3S}  & \multicolumn{1}{c|}{3D} &  \\
\midrule
\rowcolor{row_color}
\multicolumn{7}{l}{\textit{Proprietary models}} \\
\includegraphics[height=0.75em]{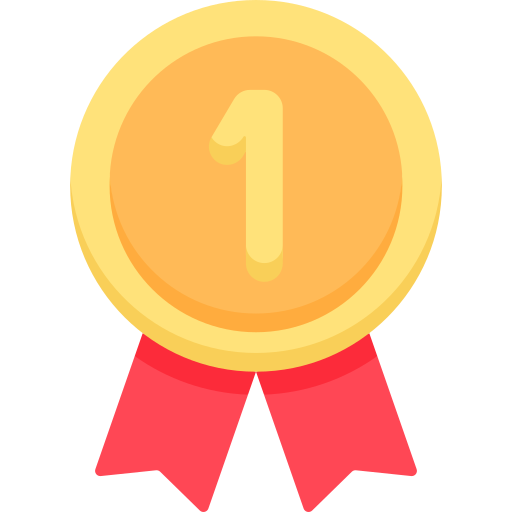}\ ChatGPT-5.1~\cite{OpenAIgpt5.1} & -- & \bestsub{46.3} & 56.2 & \bestsub{51.5} & \bestsub{58.3} & \best{53.1} \\
\includegraphics[height=0.75em]{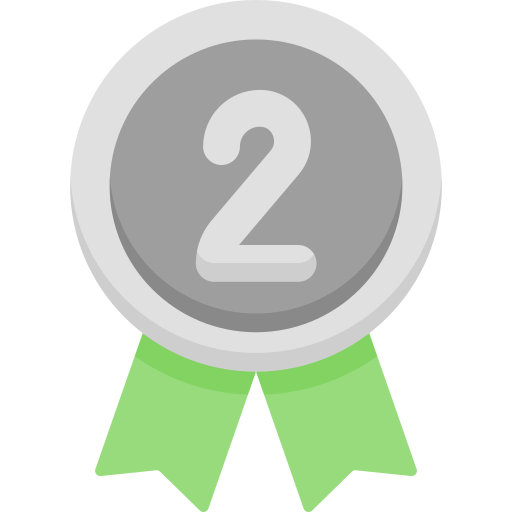}\ Gemini-2.5 Pro~\cite{google2025gemini25pro} & -- & 44.8 & \bestsub{57.5} & 42.4 & 53.7 & 49.6 \\
\includegraphics[height=0.75em]{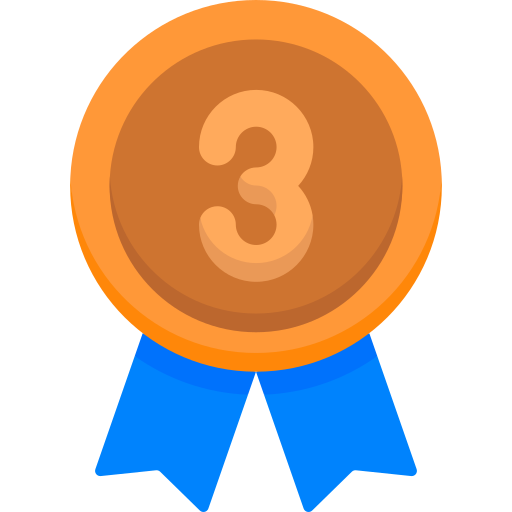}\ Gemini-2.5 Flash~\cite{google2025gemini25flash} & -- & 40.3 & 53.2 & 43.6 & 57.4 & 48.6 \\
Grok 4.1 (Fast Reasoning)~\cite{grok41} & -- & 39.4 & 49.5 & 42.4 & 48.6 & 45.0 \\
ChatGPT-5~\cite{OpenAIgpt5} & -- & 36.6 & 35.0 & 25.9 & 33.1 & 32.6 \\
Claude Sonnet 4.5~\cite{anthropic2025claude45sonnet} & -- & 19.6 & 23.0 & 19.6 & 29.1 & 22.8 \\
\midrule
\rowcolor{row_color}
\multicolumn{7}{l}{\textit{Open-weight models}} \\
\includegraphics[height=0.75em]{sec/images/icons/medal1.png}\ Qwen3-VL-Instruct-32B~\cite{bai2025qwen25vl} & 32B & 35.8 & \bestopensub{51.6} & \bestopensub{43.2} & \bestopensub{53.4} & \bestopen{46.0} \\
\includegraphics[height=0.75em]{sec/images/icons/medal2.png}\ InternVL-3.5-30B~\cite{chen2024internvl} & 30B & \bestopensub{36.7} & 45.4 & 38.6 & 42.0 & 40.7 \\
\includegraphics[height=0.75em]{sec/images/icons/medal3.png}\ Qwen3-VL-Instruct-8B~\cite{bai2025qwen25vl} & 8B & 26.0 & 47.8 & 35.1 & 46.3 & 38.8 \\
InternVL-3.5-8B~\cite{chen2024internvl} & 8B & 27.3 & 41.8 & 34.4 & 38.3 & 35.4 \\
Molmo-7B~\cite{allenai2025molmo} & 7B & 30.0 & 43.1 & 24.4 & 36.6 & 33.5 \\
Phi-4-Multimodal-Instruct~\cite{microsoft2025phi4multi} & 5.6B & 33.4 & 42.3 & 25.4 & 28.4 & 32.4 \\
Qwen3-VL-Instruct-2B~\cite{bai2025qwen25vl} & 2B & 27.1 & 39.0 & 17.6 & 32.1 & 29.0 \\
SmolVLM-Instruct~\cite{marafioti2025smolvlm} & 0.26B & 31.6 & 34.4 & 20.0 & 27.8 & 28.5 \\
InternVL-3.5-2B~\cite{chen2024internvl} & 2B & 25.0 & 31.1 & 16.6 & 27.4 & 25.0 \\
VILA-7B~\cite{lin2023vila} & 7B & 23.0 & 29.8 & 14.4 & 23.0 & 22.6 \\
CogVLM2 Video~\cite{wang2024cogvlm2} & 12B & 19.4 & 28.7 & 12.7 & 27.9 & 22.2 \\
Phi-3-Mini-128K-Instruct-3.8B~\cite{microsoft2024phi3} & 3.8B & 17.3 & 14.7 & 19.5 & 18.6 & 17.5 \\
LLaVA-13B~\cite{liu2023visual} & 13B & 14.4 & 22.1 & 8.0 & 16.5 & 15.2 \\
MiniCPM-V 4.5~\cite{yao2024minicpmv} & 8B & 27.6 & 26.3 & 0.4 & 0.0 & 13.6 \\
Fuyu-8B~\cite{adept2023fuyu} & 8B & 9.5 & 14.7 & 9.5 & 16.2 & 12.5 \\
\midrule
\textcolor{gray}{Human Baseline} & \textcolor{gray}{--} &
\textcolor{gray}{50.0} & \textcolor{gray}{59.1} &
\textcolor{gray}{55.2} & \textcolor{gray}{57.9} & \textcolor{gray}{55.6} \\
\bottomrule
\end{tabular}
}
\caption{
Evaluation results on \textsc{QuantiPhy}. We report Mean Relative Accuracy (MRA \%) on four kinematic categories (2S, 2D, 3S, 3D) and their average. \best{Dark} cell marks the best overall model and \bestopen{light} cell marks the best open-weight model.
}
\label{tab:main-results}
\end{minipage}
\hfill
\begin{minipage}[t]{0.35\textwidth}
\centering
\resizebox{\linewidth}{!}{
\begin{tabular}{
>{\centering\arraybackslash}p{1.7cm}|
>{\centering\arraybackslash}p{1.7cm}|
>{\centering\arraybackslash}p{1.9cm}|
>{\centering\arraybackslash}p{1.4cm}
}
\toprule
\multicolumn{1}{c}{\multirow{2}{*}{Video + Prior}} &
\multicolumn{1}{c}{\multirow{2}{*}{Prior only}} &
\multicolumn{1}{c}{\multirow{2}{*}{Counterfactual}} &
\multicolumn{1}{c}{\multirow{2}{*}{CoT}} \\
\multicolumn{1}{c}{} &
\multicolumn{1}{c}{} &
\multicolumn{1}{c}{} &
\multicolumn{1}{c}{} \\
\midrule
\rowcolor{row_color}
\multicolumn{4}{l}{\textit{}} \\

56.1 & 39.0 & 15.4 & 27.7 \\ 
60.9 & 46.1 & 29.9 & 49.8 \\ 
49.8 & 36.1 & 14.4 & 22.4 \\ 
47.5 & 44.3 & 31.6 & 39.5 \\ 
34.2 & 50.8 & 29.6 & 53.7 \\ 
25.4 & 16.6 & 11.6 & 25.9 \\ 

\midrule
\rowcolor{row_color}
\multicolumn{4}{l}{\textit{}} \\

50.1 & 37.2 & 34.0 & 23.1 \\ 
45.4 & --   & 12.1 & 17.6 \\ 
40.5 & 24.9 & 12.0 & 21.0 \\ 
37.0 & 19.3 & 29.7 & 18.2 \\ 
39.8 & --   & 14.7 & 15.9 \\ 
40.0 & 20.1 & 9.2  & 23.5 \\ 
34.9 & 28.2 & 25.2 & 25.6 \\ 
38.9 & --   & 14.3 & 17.8 \\ 
32.7 & 25.1 & 22.5 & 21.5 \\ 
31.8 & --   & 14.1 & 10.0 \\ 
28.5 & --   & 9.5  & 26.4 \\ 
11.1 & 10.3 & 8.4  & 7.2  \\ 
20.2 & --   & 13.9 & 14.4 \\ 
29.7 & --   & 19.9 & 24.1 \\ 
14.3 & --   & 9.0  & 21.1 \\ 

\midrule
\textcolor{gray}{--} & \textcolor{gray}{--} &
\textcolor{gray}{--} & \textcolor{gray}{--} \\
\bottomrule
\end{tabular}
}
\caption{Extensive results on an analysis subset. We report Mean Relative Accuracy (MRA) in \%. Rows follow the same model order as in Table~\ref{tab:main-results}.}

\label{tab:more-results}
\end{minipage}

\end{table*}

\section{Dissecting Quantitative Reasoning in VLMs}
\label{sec:analysis}

Beyond directly evaluating VLMs as black boxes on kinematic inference tasks, we next take an inside look at \emph{how} they arrive at their quantitative conclusions. Specifically, we investigate three aspects: (i) how scene context, such that background complexity and the number of objects, modulates task difficulty; (ii) the extent to which VLMs faithfully use the provided video and priors, rather than relying on memorized world knowledge; and (iii) whether structured prompting with step-by-step guidance can systematically improve kinematic inference.

\subsection{Effect of Scene Context}
\label{subsec:scene_difficulty}

For a more fine-grained analysis, we further categorize each video-question pair along a third axis describing the visual environment:
\emph{Scene Difficulty} $\in \{\texttt{SX}, \texttt{MX}, \texttt{SS}, \texttt{MS}, \texttt{SC}, \texttt{MC}\}$.
The first letter indicates whether there is a single (\texttt{S}) or multiple (\texttt{M}) moving objects.
The second letter indicates the background:
a solid plain color (\texttt{X})\footnote{Some \texttt{X}-type scenes are constructed by segmenting the target object with SAM2 and compositing it onto a uniform background, i.e., a fully ``denoised'' variant of the original video.}, a simple but textured scene (\texttt{S}), or a visually complex scene (\texttt{C}).

\begin{figure}
    \centering
    \includegraphics[width=0.99\linewidth]{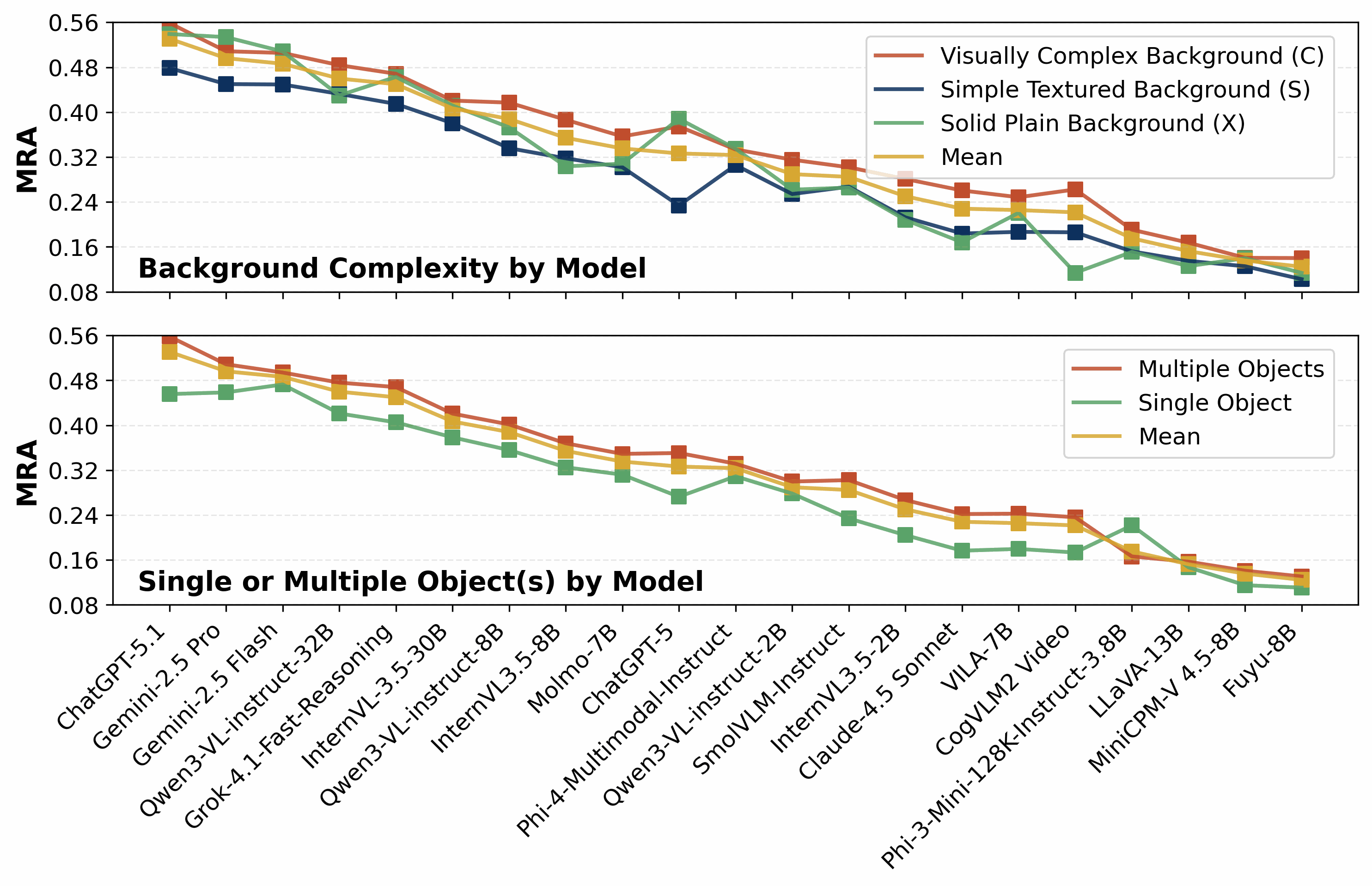}
    \vspace{-0.5em}
    \caption{\textbf{Effect of scene context.} We plot the MRA (\%) scores for all benchmark models on different categories, sorted in descending order according to their average MRA performance.}
    \label{fig:mra_plot}
\end{figure}

\autoref{fig:mra_plot} summarizes MRA across these categories.
Overall, we observe that background complexity has only a mild effect on model's performance on the task.
Models perform slightly \emph{better} in the SAM-denoised condition than in the simple-texture (\texttt{S}) background, suggesting that removing irrelevant clutter reduces distractions and stabilizes quantitative estimates.
Interestingly, performance in visually complex scenes (\texttt{C}) is above the other two background conditions for most models.
A plausible explanation is that realistic backgrounds provide additional reference cues (e.g., tiles, windows, or road markings) that help the model infer scale and motion.

In contrast, the number of objects in the scene exhibits a clearer trend:
setups with multiple objects (\texttt{MX}, \texttt{MS}, \texttt{MC}) consistently yield higher MRA than their single object counterparts (\texttt{SX}, \texttt{SS}, \texttt{SC}).
Having more objects gives the model extra reference targets (e.g., another ball, a ruler-like structure), which can be used as implicit comparison standards for both size and speed. 

These observations highlight that VLMs robustly benefit from richer background information and relational structure in the scene.

\subsection{Do VLMs Use Videos and Priors Faithfully?}
\label{subsec:faithful_use}

\textsc{QuantiPhy} is designed to assess whether VLMs can truly understand physical events captured by visual sensors. It is therefore crucial to investigate whether these systems genuinely comprehend the provided visual signals, rather than merely producing \emph{plausible} guesses. Here, we re-evaluate all models on a controlled subset of $161$\texttt{2D} video-prior pairs,\footnote{Due to resource constraints, the experiments in \autoref{subsec:faithful_use} and \autoref{subsec:cot} are run on a 2D-only subset of 161 instances, and MRA values therefore differ slightly from those in \autoref{tab:main-results}.} using two complementary probes.

\paragraph{Key finding: VLMs rely more on learned prior knowledge than visual inputs for physical reasoning.}
We first compare the default \emph{video+prior} setting against a \emph{prior-only} setting in which the video is removed but the prompt (including the physical prior, object description, and question) is kept unchanged.
Hypothetically, without the reference video, we would expect an \emph{evident performance drop}, since VLMs would be forced to guess; even if such guesses are reasonable, they may not correspond to the specific instance.
However, across most models we observe only modest differences between the two conditions: on the 161-pair subset, MRA in the \emph{prior-only} setting is often close to, and sometimes only slightly below, the \emph{video+prior} setting, even for strong systems such as ChatGPT-5.1, Gemini~2.5~Pro/Flash, and Qwen3-VL-Instruct-32B.
In other words, models can already obtain reasonably high scores by relying on their \emph{internal} prior knowledge about typical object sizes and speeds, with limited added value from the actual video frames.
This suggests that, in our tasks, many VLMs behave less like visual measurers and more like powerful guessers conditioned on textual hints.

\paragraph{Key finding: VLMs (mostly) do not reason but memorize.}
To further test whether models faithfully utilize the conditioning prior and video to infer the target physical quantity, we perform a counterfactual analysis. For each of the 161 instances, we construct a family of counterfactual prompts by multiplying the original physical prior by a scalar factor $\alpha \in \{0.001, 0.01, 0.1, 0.2, 5, 50, 100, 200, 500, 700\}$, while keeping the video and question unchanged. Hypothetically, if a model is capable of correct kinematic reasoning, its prediction should scale accordingly, tracking the \emph{counterfactual ground truth} $y^{\text{cf}} = \alpha \cdot y$. As a result, we would expect no, or at least only modest, performance drop in this test.
However, in practice, we find that even the best models obtain very low scores in this setting: most models' MRA drops by $80\%$, and even the strongest model drops by $70\%$. Despite being given a numerically precise but altered prior, the outputs remain close to the original physical magnitudes implied by real-world experience, rather than those dictated by the provided priors.

Taken together, these results lead to a consistent conclusion: \textbf{\emph{existing VLMs are not yet input-faithful quantitative reasoners.}}
They only weakly exploit pixel-level information in videos, and they do not reliably condition on the exact numerical priors provided in the prompts.
Instead, their quantitative kinematic inferences are dominated by internal, pre-trained world knowledge, with visual evidence and explicit priors acting more as soft hints than hard constraints.

\subsection{Do Structured Prompts Help VLMs Reason?}
\label{subsec:cot}

Based on the above observation that VLMs rely more heavily on text prompts, we therefore further investigate whether a more structured ``chain-of-thought" reasoning pattern can improve quantitative physical predictions.
Instead of directly asking for the target quantity, we decompose each question into a four-step chain:

\noindent\texttt{(I) Pixel-level source property:} ``What is [the ground-truth object's property] in pixels?''

\noindent\texttt{(II) Scale estimation:} ``What is the proportional relationship between pixels and [a kinematic scale]?''

\noindent\texttt{(III) Pixel-level target property:} ``What is [the inference target's property] in pixels?''

\noindent\texttt{(IV) World-level target property:} ``What is [the inference target's real-world property]?''

We query each model sequentially on the same analysis subset as in \autoref{subsec:faithful_use}. If the model produces a parseable numerical answer at a given step, we append both the question and the extracted answer as additional context for the next step. If the model fails to generate a valid number at a step, subsequent prompts omit that step and its answer. We evaluate models' final answers with MRA.

As shown in \autoref{tab:more-results}, chain-of-thought (CoT) prompting is less helpful than anticipated. Among the 21 models we examined, only three show any improvement, with ChatGPT-5 and Fuyu-8B exhibiting a noticeable increase in MRA. For the remaining 19 models, including several strong open-weight systems, performance under our CoT protocol is \emph{worse} than under direct zero-shot prompting, sometimes by a large margin. In other words, under our setup, explicitly spelling out pixel measurement, scale estimation, and rescaling does \emph{not} systematically improve current VLMs' quantitative reasoning. Many models appear unable to reliably solve the intermediate numeric subproblems, so decomposing the task mainly amplifies and propagates early errors.

Our analysis suggests that existing VLMs can exploit visual cues, priors, and structured prompts to reason quantitatively, but do so in a rather brittle and often inconsistent manner. \textsc{QuantiPhy} thus provides not only a benchmark for aggregate performance, but also a diagnostic tool for probing where quantitative physical reasoning succeeds, fails, and how it might be improved. We provide additional studies in the supplementary material \autoref{appendix:more_studies} to further support our findings.

\section{Discussion and Future Work}
\label{sec:discussion}

\noindent\textbf{Summary of Findings.} Our key finding is that state-of-the-art VLMs have not yet established a reliable link between visual observations and quantitative physical facts. This disconnect manifests as a critical lack of \emph{input faithfulness}: although models process video inputs, our extensive studies show that they hardly infer kinematic properties from the actual pixel-level information. Instead, they exhibit a strong reliance on parametric priors, often overriding explicit user inputs and visual evidence in favor of memorized world knowledge. Consequently, current systems act more as approximate guessers'' based on semantic context rather than precise visual measurers,'' limiting their reliability for real-world embodied agents.

\noindent\textbf{Limitations.} Our study has several limitations that suggest avenues for future work. The dataset focuses exclusively on translational movement, omitting rotational dynamics, and utilizes a fixed camera perspective, which simplifies the task compared to real-world scenarios with dynamic viewpoints. Additionally, we examined only rigid objects, excluding soft bodies and deformable materials. Finally, our dataset is relatively simplified, featuring isolated movements rather than complex, multi-object interactions.

\noindent\textbf{Conclusion and Future Work.} These limitations highlight promising directions for future research. First, a more comprehensive video dataset is needed to evaluate VLMs on more diverse physics, incorporating the complexities we omitted: rotational dynamics, deformable objects, varied camera perspectives, and complex multi-body interactions. Second, our findings can inform new VLM training methodologies, such as physics-informed objectives or specialized pre-training on physics-rich data. Ultimately, this research aims to advance the development of generalist embodied AI agents capable of sophisticated reasoning and interaction with the physical world.

\section*{Acknowledgment}
The authors thank Juze Zhang and Heng Yu for their help with lab data capturing. This work was partially funded by the NIH Grant R01AG089169, Stanford HAI Hoffman-Yee Award, Stanford HAI graduate fellowship, and UST.

{
    \small
    \bibliographystyle{ieeenat_fullname}
    \bibliography{main}
}

\clearpage
\setcounter{page}{1}

\clearpage
\onecolumn          

\appendix
\maketitlesupplementary


\addtocontents{toc}{\protect\setcounter{tocdepth}{3}}
\hypersetup{linkcolor=black}
{\small \tableofcontents} 
\newpage

\clearpage
\twocolumn  

\section{More Studies and Results}
\label{appendix:more_studies}
\subsection{Additional Case Studies}

To better understand how Vision-Language Models (VLMs) solve kinematic inference tasks beyond aggregate scores, we conduct a qualitative case study on the top-performing model, ChatGPT-5.1, using its \texttt{Thinking} mode in the standard user-facing interface.
For each selected video–text pair, we repeatedly query the model and inspect the tool-augmented chain-of-thought until we obtain representative traces that are syntactically well-formed and numerically valid.\footnote{We prompt the model multiple times and observe substantial variability: on the same instance, some runs produce accurate and well-structured reasoning, while others fail to parse the question or ignore key inputs.
We therefore collect several responses per instance and manually select representative traces that are syntactically coherent and numerically valid for detailed analysis.
A similar instability is also present in our API-based evaluation, even with temperature fixed to $0$; in the main benchmark, we address this by running multiple trials and recording a failure rate.}
We then analyze both the final numerical answers and the intermediate reasoning steps.\footnote{We emphasize that this analysis is diagnostic rather than evaluative: we study one specific model’s internal behavior to illustrate broader patterns of (un)faithful quantitative reasoning.}

Overall, we observe a sharp contrast between successful instances, where the model follows a textbook-like “measure pixels $\rightarrow$ apply prior $\rightarrow$ compute target” pipeline,  and failure modes, where it largely ignores the video and the provided prior, and instead falls back to pre-trained world knowledge or generic heuristics.
Below we discuss four representative cases.

\begin{figure}
    \centering
    \includegraphics[width=0.99\linewidth]{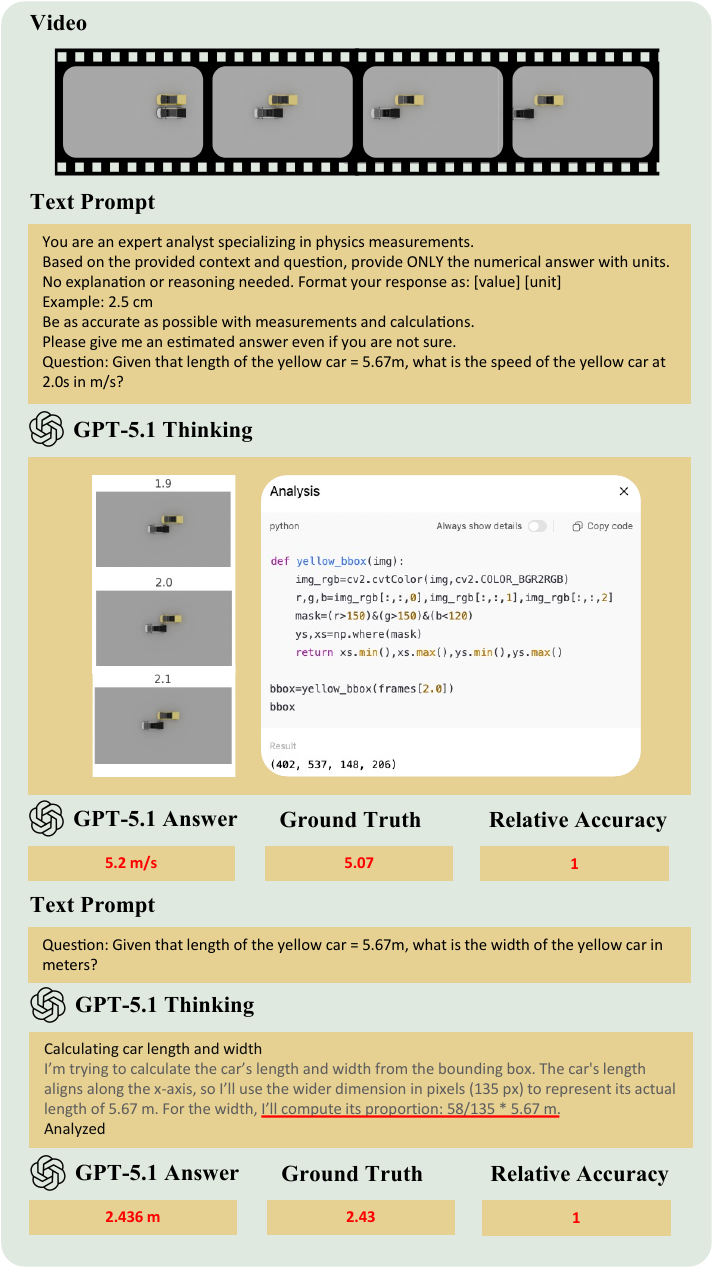}
    \vspace{-0.5em}
    \caption{\textbf{Case 1: Faithful pixel–prior reasoning.}}
    \label{fig:case1}
\end{figure}

\noindent\textbf{Case 1: Faithful pixel–prior reasoning.}
\autoref{fig:case1} shows a 2D scene with a yellow car moving laterally.
The model is asked two questions: (i) given that the car’s length is $5.67$\,m, what is its speed at $2.0$\,s; and (ii) what is the car’s width in meters.
In this instance, ChatGPT-5.1’s chain-of-thought closely matches the intended reasoning procedure.
The model first identifies the relevant frames around $t=2.0$\,s, uses OpenCV-style tools to obtain bounding boxes, and explicitly treats the longer side of the box (135\,px) as the car’s length in pixel space.
It then calibrates a pixel-to-meter scale from the given length prior (5.67\,m), and computes the width as
\[
\text{width} \approx \frac{58}{135} \times 5.67\text{m}\approx 2.44\text{m}\,,
\]
 which is close to the ground truth width and achieves high relative accuracy.
Here the model behaves as an input-faithful visual measurer: it grounds both the prior and the kinematic target in pixel space and performs the correct proportional reasoning.
When this pipeline is followed, the resulting numerical answers are often near the ground truth.

\begin{figure}
    \centering
    \includegraphics[width=0.99\linewidth]{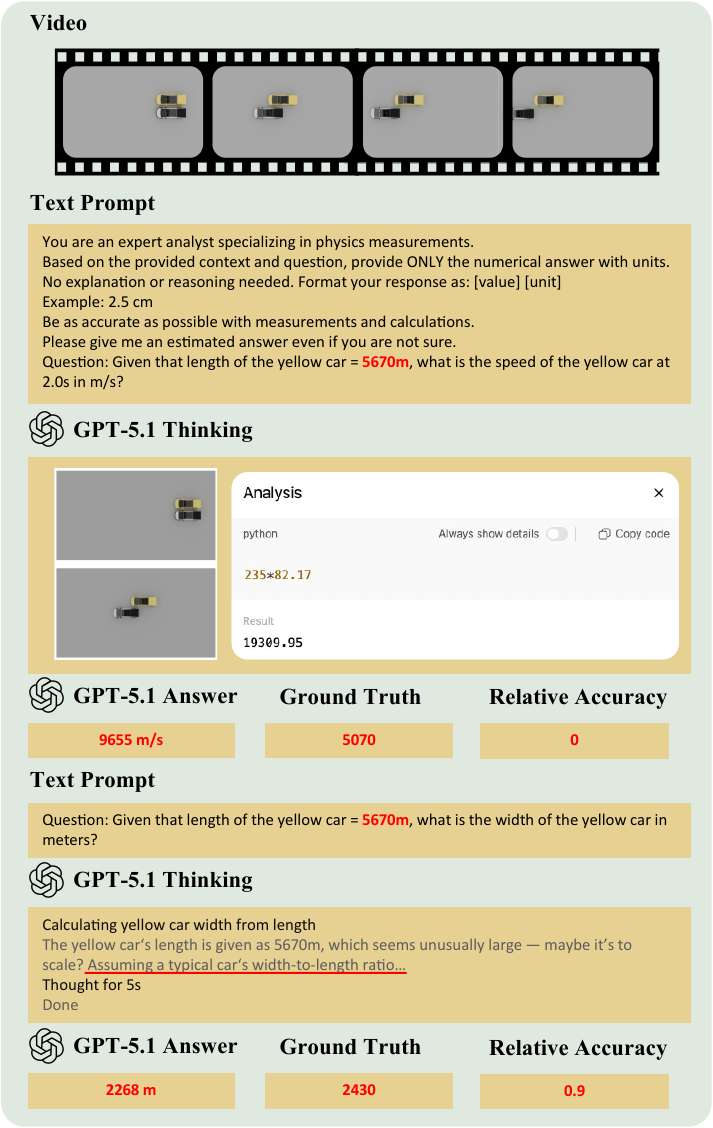}
    \vspace{-0.5em}
    \caption{\textbf{Case 2: Counterfactual prior breaks faithfulness.}}
    \label{fig:case2}
\end{figure}

\noindent\textbf{Case 2: Counterfactual prior breaks faithfulness.}
In the second case (\autoref{fig:case2}), we reuse the same video but multiply the car-length prior by a counterfactual factor of 1000, changing the input to “length of the yellow car = 5670\,m.”
The task is again to infer speed at $2.0$\,s and the car’s width.
In its \texttt{Thinking} trace, the model explicitly notes that “5670\,m” is an implausible car length and expresses confusion.
Crucially, instead of continuing to rely on pixel measurements and the (counterfactual) prior, it effectively abandons the video and the numeric input.
For the width question, it switches to a generic heuristic, assuming a “typical car’s width-to-length ratio” and hallucinating a plausible-looking width independent of the actual scene.
The final width prediction happens to have high relative accuracy (close to $0.9$), but this success is not input-faithful; it arises from pre-trained knowledge about cars rather than from the specific video or the given prior.
This case highlights a key risk that purely outcome-based metrics can judge an answer as “good,” while the underlying reasoning ignores the provided evidence.

\begin{figure}
    \centering
    \includegraphics[width=0.99\linewidth]{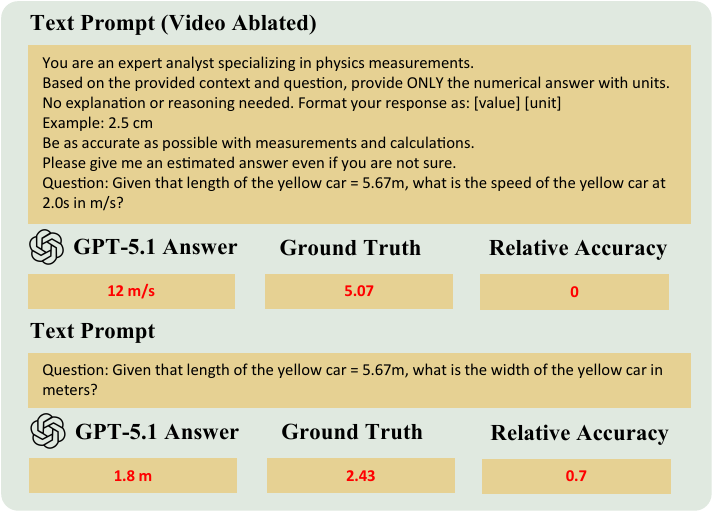}
    \vspace{-0.5em}
    \caption{\textbf{Case 3: Video ablation reveals reliance on priors.}}
    \label{fig:case3}
\end{figure}

\noindent\textbf{Case 3: Video ablation reveals reliance on priors.}
Case 3 (\autoref{fig:case3}) uses a video-ablation setting. The model receives only the text prompt (including “length of the yellow car = 5.67\,m”), without access to the video.
When asked for the car’s speed at $2.0$\,s, ChatGPT-5.1 produces an answer ($12$\,m/s) that is far from the ground truth, confirming that motion estimation is difficult without visual evidence.
However, when asked for the car’s width (still without video), the model outputs a numerically reasonable value with relatively high accuracy (relative accuracy $\approx 0.7$).
Since no pixel information is available in this ablated setting, this behavior can only be explained by the model’s internal prior over typical car dimensions.
Combined with Case 1 and 2, this suggests a pattern that even when video is available, much of the “size” inference can be driven by pre-trained world knowledge rather than by explicit pixel measurements.

\begin{figure}
    \centering
    \includegraphics[width=0.99\linewidth]{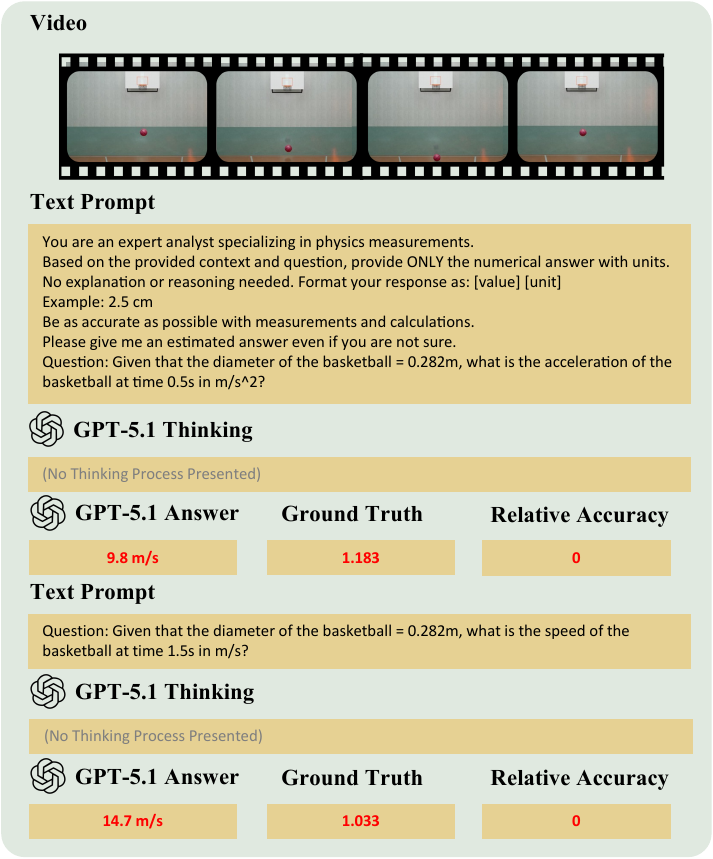}
    \vspace{-0.5em}
    \caption{\textbf{Case 4: Strong gravitational prior overrides counterfactual physics.}}
    \label{fig:case4}
\end{figure}

\noindent\textbf{Case 4: Strong gravitational prior overrides counterfactual physics.}
The fourth case (\autoref{fig:case4}) involves a Blender-simulated basketball scene that visually resembles a realistic indoor court, but with counterfactual physics. The ball’s acceleration is time-varying and close to $1\,\text{m/s}^2$, rather than standard gravity. The model is asked for the ball’s acceleration at $0.5$\,s and its speed at $1.5$\,s, given the ball’s diameter as a prior.
Here, ChatGPT-5.1 completely ignores both the video and the non-standard trajectory implied by the simulation.
It directly outputs the canonical gravitational acceleration $9.8\,\text{m/s}^2$, and for speed simply multiplies $g$ by time (i.e., $9.8 \times 1.5 = 14.7\,\text{m/s}$), leading to relative accuracy equals to $0$ on both queries.
No pixel measurements or scale computations appear in the \texttt{Thinking} trace.
This illustrates how strong pre-trained physical priors (e.g., “objects fall with acceleration $g$”) can dominate the model’s behavior, even when they contradict the actual visual input and the provided prior.

\noindent\textbf{Discussion.}
These four cases collectively sharpen our main quantitative findings.
\begin{itemize}[leftmargin=*]
  \item When everything works, ChatGPT-5.1 can execute an impressive, tool-augmented pipeline that does read pixel trajectories, apply the physical prior, and compute accurate kinematic quantities.
  \item However, this behavior is fragile. As soon as the prior becomes counterfactual, the video is ignored, or the underlying physics departs from familiar regimes, the model quickly reverts to pre-trained world knowledge or rough heuristics, often ignoring the provided inputs.
  \item High numerical accuracy does not guarantee input-faithful reasoning. Specifically, Case 2 demonstrates that a model can get the “right” answer for the wrong reasons, while Cases 3 \& 4 show that it can stick to canonical physical constants even when the scene violates them.
\end{itemize}

These observations suggest that improving VLMs’ quantitative physical reasoning will require not only better aggregate performance, but also mechanisms that encourage faithful use of visual evidence and explicit numerical priors, rather than letting powerful, but sometimes misleading, pre-trained world knowledge dominate the inference process.

\subsection{Metric Design Justification}
\label{appendix:metric_design}

In \textsc{QuantiPhy}, we adopt Mean Relative Accuracy (MRA) as the primary evaluation metric for quantitative physical inference tasks with continuous outputs.
Each prediction $\hat{y}$ is compared to the ground truth $y$ using its relative error $|\hat{y} - y| / y$, and awarded partial credit if the error falls below confidence thresholds $\theta \in \{0.5, 0.55, \ldots, 0.95\}$.
The final MRA score is the average of binary accuracies across these thresholds
\begin{equation}
\mathrm{MRA} = \frac{1}{10} \sum_{\theta \in \mathcal{C}}
\mathbbm{1} \left( \frac{|\hat{y} - y|}{|y|} < 1 - \theta \right),
\end{equation}
where $C = \{0.5, 0.55, \ldots, 0.95\}$ is the set of confidence thresholds.

While one could alternatively compute continuous relative error (e.g., mean relative error or mean absolute percentage error), we prefer MRA for several practical and conceptual reasons.

\noindent\textbf{Discrete but calibrated.}
MRA discretizes accuracy into a finite set of thresholds, offering interpretable feedback on how often the model is ``close enough'' under increasing demands.
Rather than penalizing deviations proportionally (which can be dominated by outliers), MRA provides a graded scale of correctness, similar in spirit to the mAP@IoU\footnote{mAP@IoU: Mean Average Precision (mAP) calculated at specific Intersection over Union (IoU) thresholds.} metrics in object detection.

\noindent\textbf{Robust to ambiguity and noise.}
Many video-based physical inferences involve semantic or visual uncertainty.
For example, estimating a person’s height may vary depending on whether hair or shoes are included;
estimating a cup’s diameter may depend on whether the inner or outer rim is used.
MRA tolerates such ambiguity by granting full credit when answers fall within a reasonable margin.

Moreover, measurement noise is often unavoidable:
\begin{itemize}
  \item \textbf{Temporal aliasing:} limited frame rates restrict temporal resolution for computing velocities and accelerations;
  \item \textbf{Motion blur:} fast-moving objects introduce visual uncertainty during measurement;
  \item \textbf{Imprecise priors:} even real-world scale references (e.g., credit card dimensions) may not be perfectly visible or aligned.
\end{itemize}
MRA accommodates these natural imperfections more flexibly than a regression-style loss.

\noindent\textbf{Supported by precedent.}
The MRA metric was introduced by Yang et al.~\cite{yang2025thinking} in VSI-Bench for evaluating numerical answers in visual–spatial reasoning tasks.
Some follow-up work~\cite{sridhar2025vreason, cai2025holistic} adopted the same evaluation to benchmark multimodal models in physical and spatial settings.
These works motivate MRA as a stable and discriminative way to capture proximity between predicted and ground-truth values, especially when scale varies across examples.
Our use of MRA continues this design choice, ensuring comparability while improving robustness.

In sum, MRA balances informativeness, robustness, and interpretability.
It is well-suited for evaluating VLMs on physically grounded, numerically sensitive tasks like those in \textsc{QuantiPhy}, where small deviations are acceptable, but large errors are unacceptable regardless of scale.

\begin{figure*}[ht!]
    \centering
    \includegraphics[width=0.80\linewidth]{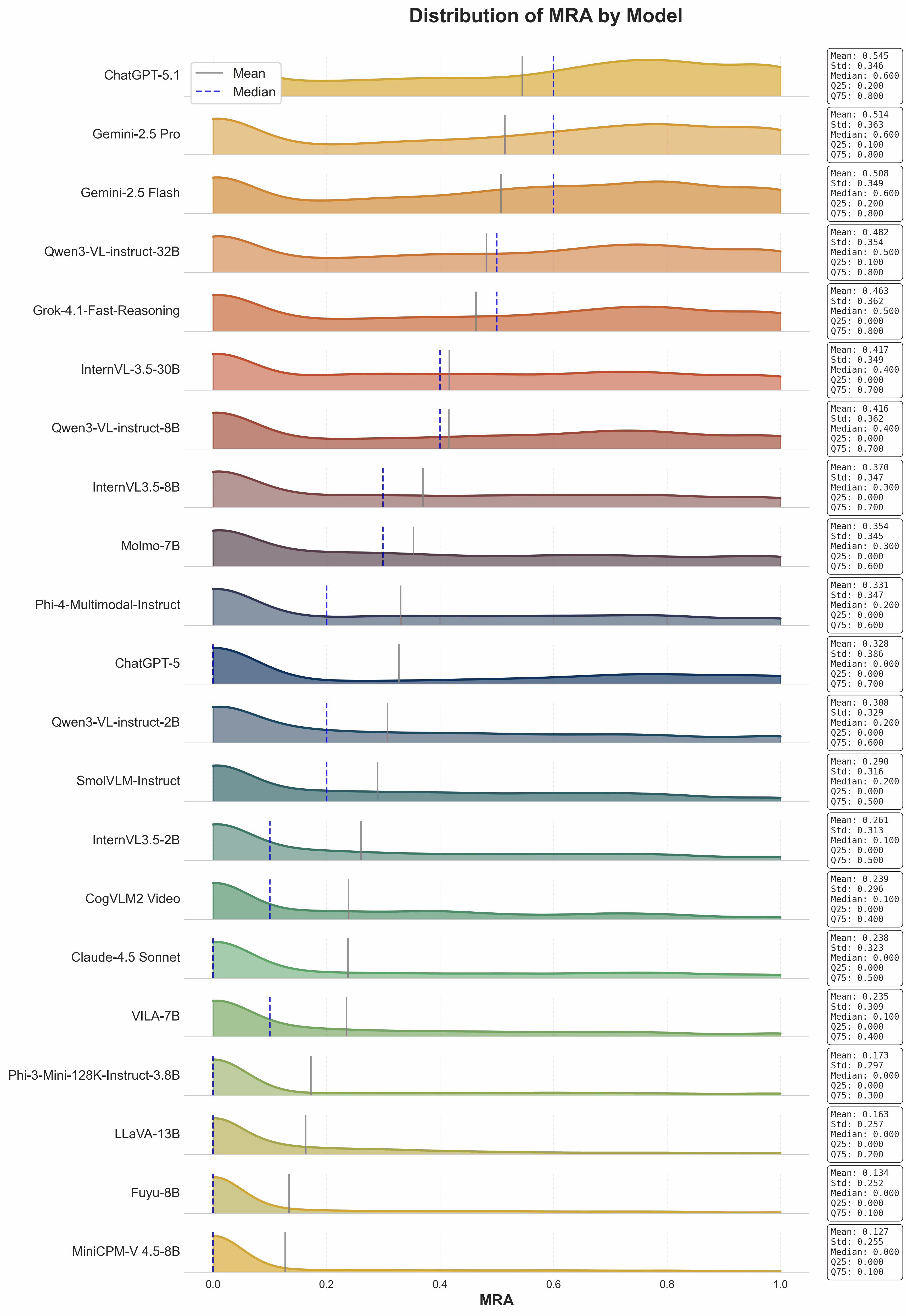}
    \caption{\textbf{Distribution of MRA by model.} One caveat to note is that the Avg. MRA in \autoref{tab:main-results} reflects the mean MRA across inference task categories for each model (i.e., the average MRA of \texttt{2D-Static}, \texttt{2D-Dynamic}, \texttt{3D-Static}, and \texttt{3D-Dynamic}). In contrast, the mean in this distribution plot represents the average MRA at the individual-question level for each model.}
    \label{fig:mra_density}
\end{figure*}

\subsection{Model MRA Distribution}

\autoref{fig:mra_density} visualizes the MRA distributions for each VLM. Each subplot shows a density curve with mean (gray solid) and median (blue dashed) lines, along with summary statistics (mean, std, median, Q25, Q75) displayed to the right.

The top-performing models, including ChatGPT-5.1, Gemini-2.5 Pro, Gemini-2.5 Flash, Qwen3-VL-instruct-32B, and Grok-4.1-Fast-Reasoning, show notably higher densities around moderate to high MRA values above 0.5, with means and medians clustering around 0.4–0.6. Their smoother, more concentrated curves indicate both higher accuracy and more consistent performance across the evaluation set.

A second tier of models, including InternVL-3.5-30B, Qwen3-VL-instruct-8B, Molmo-7B, and ChatGPT-5, exhibits slightly lower means and medians, generally around 0.3–0.4. Their broader distributions with heavier tails indicate greater variability that these models produce some strong outputs but also more low-scoring cases. The alignment of medians and means suggests errors are not extremely skewed.

Mid-tier models such as Qwen-3-VL-instruct-2B, Phi-4-Multimodal-Instruct, SmolVLM-Instruct, CogVLM2 Video, Claude-4.5 Sonnet, and VILA-7B show means of 0.2–0.3. Their distributions are heavily weighted toward low MRA values with thin right tails, indicating that while they occasionally achieve moderate scores, they rarely reach the performance levels of top-tier systems.

The weakest-performing models, for example, Phi-3-Mini-128K-Instruct-3.8B, LLaVA-13B, Fuyu-8B, and MiniCPM-V 4.5-8B, show distributions highly concentrated near zero. With means below 0.15 and medians around 0.0, these models fail to produce meaningful MRA performance in most cases.

Notably, many distributions have substantial mass centered around zero. An MRA of zero indicates either the model output zero as an answer or failed to produce proper numerical output. We also observed that some model APIs are unstable and produce errors over time, potentially due to API server error, running environment, internet traffic, batch size variations, etc. We abstract away from these in this paper and simply treat these failed cases as observations with MRA equal to zero.

Overall, modern frontier proprietary models cluster around substantially higher and more consistent MRA values, mid-tier models show moderate capability with noticeable variability, and smaller or older models yield predominantly low scores. The density patterns and comparative statistics reveal a clear performance gap between state-of-the-art systems and lightweight or earlier-generation models.
\section{Dataset Construction Guidelines} \label{appendix:data_collection}

\subsection{General Principles}
\label{sec:general_principles}
Our data collection and curation follow several general principles to ensure that \textsc{QuantiPhy} is ethically sourced, physically well-defined, and suitable for quantitative evaluation.

\noindent\textbf{Copyright and ethics.}
We carefully avoid copyright and ethical issues throughout all stages of data collection and processing.
All videos, 3D assets, and simulation resources are either open-source, licensed for research use, or explicitly verified to pose no known copyright conflicts before inclusion.
When raw videos contain personally identifiable information (e.g., human faces, license plates), we apply blurring or masking to anonymize the content.

\noindent\textbf{Video selection criteria.}
To make quantitative kinematic inference well-posed and to reduce confounding factors, we enforce the following constraints on all collected videos.
\begin{itemize}[leftmargin=*]
  \item \textbf{Static camera in world coordinates.}
  The camera remains fixed in the world frame during each clip.
  This avoids entangling camera motion with object motion, which would otherwise introduce additional ambiguity and noise into the inference problem.
    \item \textbf{At least one rigid object undergoing translational motion.}
    Each video contains at least one rigid object whose dominant motion is translation.
    This requirement ensures that we can formulate well-defined kinematic inference tasks.
    Non-rigid objects and purely rotational motions are left out of the current benchmark and deferred to future work.\footnote{In this work, we adopt a relaxed definition of rigid objects: we consider an object ``rigid'' if its motion can be consistently approximated by a stable center of mass across frames. This includes some entities that may exhibit slight non-rigid deformation (e.g., a flying bird or a walking person), as long as their motion remains locally trackable and structurally coherent.}
  \item \textbf{Planar motion for \texttt{2D} tasks.}
  For \texttt{2D} instances, the target object and the reasoning target are constrained to move in a plane parallel to the image plane, i.e., the depth relative to the camera remains (approximately) constant over time.
  This assumption guarantees a consistent mapping between pixel displacement and world-space distance within each clip, making the 2D kinematic inference problem well-defined.
\end{itemize}

\noindent\textbf{Video–text record schema.}
Each annotated instance in \textsc{QuantiPhy} is represented as a structured video–text record.
Table~\ref{tab:record-schema-example} shows a representative example.

\begin{table*}[t]
\centering
\small
\renewcommand{\arraystretch}{1.1}
\begin{tabular}{l p{0.68\linewidth}}
\toprule
Property & Example value \\
\midrule
\texttt{\textbf{video\_id}} &
\texttt{simulation\_0032} \\
\texttt{\textbf{video\_source}} &
\texttt{simulation} \\
\texttt{\textbf{video\_type}} &
\texttt{A3MC} \\
\texttt{\textbf{fps}} &
\texttt{30} \\
\texttt{\textbf{inference\_type}} &
\texttt{DD} \\
\texttt{\textbf{question}} &
\texttt{What is the acceleration of the orange car at 1.0s in m/s\textsuperscript{2}?} \\
\texttt{\textbf{ground\_truth\_prior}} &
\texttt{gravity acc = 9.8 m/s\textsuperscript{2}} \\
\texttt{\textbf{depth\_info}} &
\texttt{t=1s, distance\_ball\_camera = 13.80 m;}\\[-0.2em]
&
\texttt{t=2s, distance\_ball\_camera = 13.80 m;}\\[-0.2em]
&
\texttt{t=1.5s, distance\_orange\_camera = 10.18 m;}\\[-0.2em]
&
\texttt{t=2s, distance\_orange\_camera = 10.40 m;}\\[-0.2em]
&
\texttt{t=2.5s, distance\_green\_camera = 4.32 m;}\\[-0.2em]
&
\texttt{t=3s, distance\_green\_camera = 6.91 m} \\
\texttt{\textbf{ground\_truth\_posterior}} &
\texttt{2.86} \\
\bottomrule
\end{tabular}
\caption{\textbf{Example of a single video--text record in \textsc{QuantiPhy}.}}
\label{tab:record-schema-example}
\end{table*}

Each record contains the following fields:
\begin{enumerate}[leftmargin=*]
  \item \textbf{\texttt{video\_id}.} A unique identifier for the underlying video.
  \item \textbf{\texttt{video\_source}.} The data source from which the video was obtained (e.g., \texttt{simulation}, \texttt{lab}, or \texttt{internet}).
  \item \textbf{\texttt{video\_type}.} A four-letter code encoding the configuration of the task.
  The four characters denote, in order: (i) the type of physical prior (Size, Velocity, or Acceleration), (ii) whether the reasoning task is \texttt{2D} or \texttt{3D}, (iii) whether there is a single (\texttt{S}) or multiple (\texttt{M}) moving objects, and (iv) the background type, that is, plain (\texttt{X}), simple (\texttt{S}), or complex (\texttt{C}). More details are included in \autoref{subsec:video_type}.
  \item \textbf{\texttt{fps}.} The frame rate of the video, used to convert frame indices into time and to compute velocities/accelerations consistently.
  \item \textbf{\texttt{inference\_type}.} A two-letter code indicating whether the prior and the inference target are static or dynamic over time:
  \texttt{S} denotes a static quantity, and \texttt{D} denotes a time-dependent (dynamic) one.
  The first letter corresponds to the prior, and the second to the posterior.
  \item \textbf{\texttt{question}.} The natural-language prompt presented to the VLM.
  We ensure that each question explicitly specifies the physical unit (e.g., m, cm/s, m/s\textsuperscript{2}) and, for velocity or acceleration, clearly indicates whether the query concerns an instantaneous quantity at a given timestamp or an average quantity over an interval.
  \item \textbf{\texttt{ground\_truth\_prior}.} The physical prior provided to the model, formatted as a positive numeric value with unit (e.g., \texttt{gravity acc = 9.8 m/s\textsuperscript{2}}).
  We enforce consistent formatting to simplify parsing and downstream use.
  \item \textbf{\texttt{depth\_info}.} Depth annotations used only for 3D reasoning tasks.
  This field contains depth values (in metric units) for the prior object and, when needed, for the inference target at one or more timestamps.
  Depth information is designed so that, in principle, the depth of the inference target can be recovered from the provided entries.
  The formatting mirrors that of \textbf{\texttt{ground\_truth\_prior}}.
  \item \textbf{\texttt{ground\_truth\_posterior}.} The numeric ground-truth answer to the kinematic inference question, represented as a positive scalar without unit (the unit is part of the question text).
\end{enumerate}

\noindent\textbf{Balance and diversity.}
Finally, we design the dataset to be both balanced across task types and diverse in content.
\begin{itemize}[leftmargin=*]
  \item \textbf{Fine-grained video types.}
  Using the four-letter \textbf{\texttt{video\_type}} code, we partition all clips into 36 fine-grained categories.
  We ensure that each category contains at least four videos, so that every configuration is represented non-trivially.
  \item \textbf{Balanced core task categories.}
  We strive to keep the four core inference task categories (\texttt{2D-Static}, \texttt{2D-Dynamic}, \texttt{3D-Static}, \texttt{3D-Dynamic}) approximately balanced in terms of the number of videos and associated questions, enabling fair comparison across conditions.
  \item \textbf{Rich scenes and motion patterns.}
  We cover a broad spectrum of spatial scales and motion types.
  Scenes range from astronomical (e.g., planetary motion), to everyday macroscopic settings (e.g., traffic, sports), to microscopic phenomena (e.g., cells and bacteria).
  Motion patterns include uniform motion, accelerated and decelerated linear motion, projectile motion, pendulum-like oscillations, and centripetal motion, among others.
  This diversity is crucial for probing whether VLMs’ quantitative reasoning generalizes beyond narrow, highly stylized scenarios.
\end{itemize}

\subsection{Video Types}
\label{subsec:video_type}
\subsubsection{Video Categories Definition}
As described in the section , we assign each video a four-character code that encodes its physical prior, dimensionality, object setting, and background type.

\begin{figure*}[ht!]
    \centering
    \includegraphics[width=0.99\linewidth]{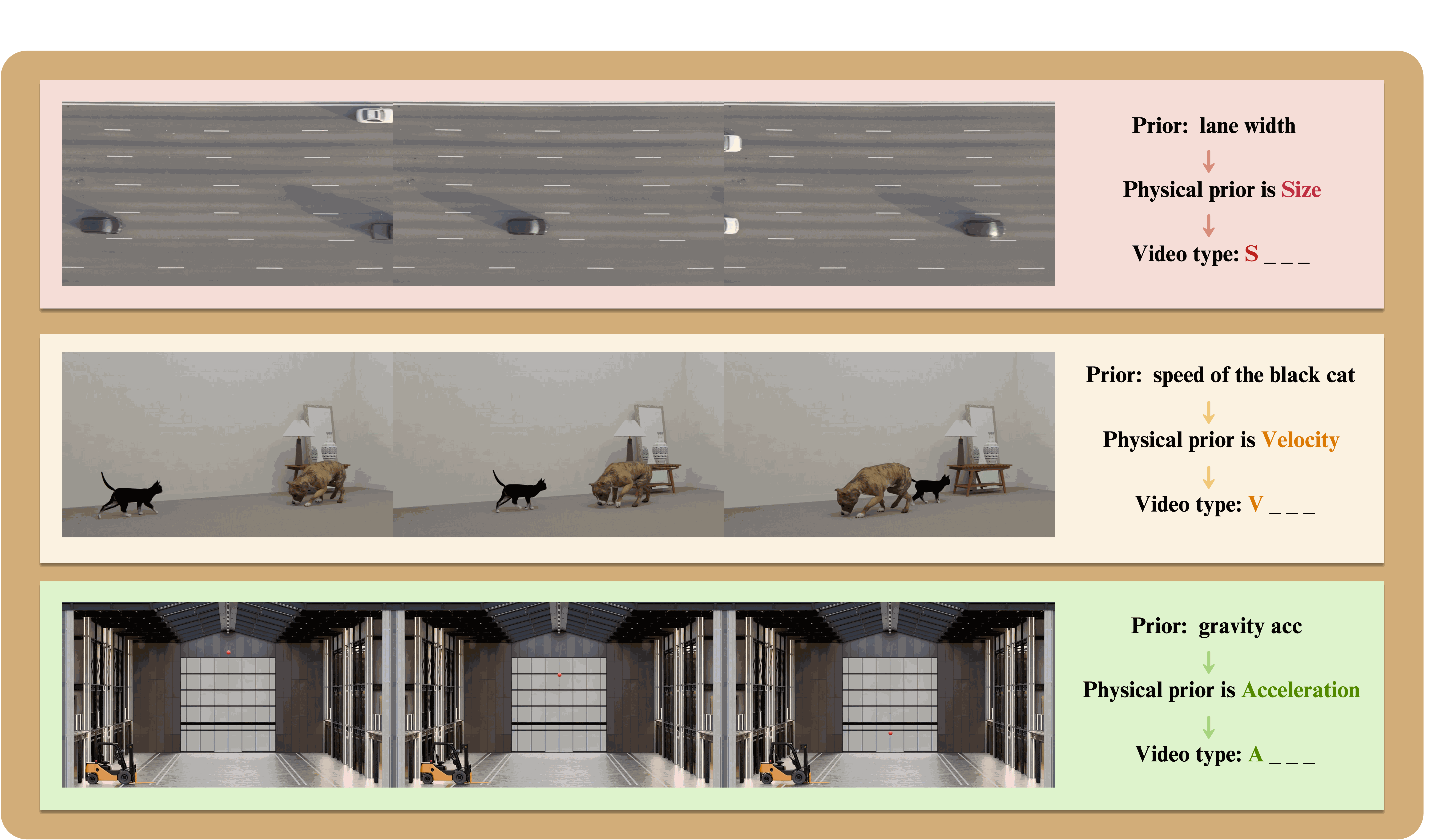}
    \caption{\textbf{Examples of S/V/A scene.}}
    \label{fig:SVA_scene}
\end{figure*}

\noindent\textbf{First character (\texttt{S} / \texttt{V} / \texttt{A}).}
The first character specifies which physical prior the video is constructed to probe. For a designated object in the scene, we use
\begin{itemize}[leftmargin=1.6em,topsep=0.2em,itemsep=0pt]
  \item \texttt{S} = size
  \item \texttt{V} = velocity
  \item \texttt{A} = acceleration
\end{itemize}
Thus, the first character takes one of $\{\texttt{S}, \texttt{V}, \texttt{A}\}$, indicating whether the ground-truth physical quantity is size, velocity, or acceleration for that object, as illustrated in \autoref{fig:SVA_scene}.

\begin{figure*}[ht!]
    \centering
    \includegraphics[width=0.99\linewidth]{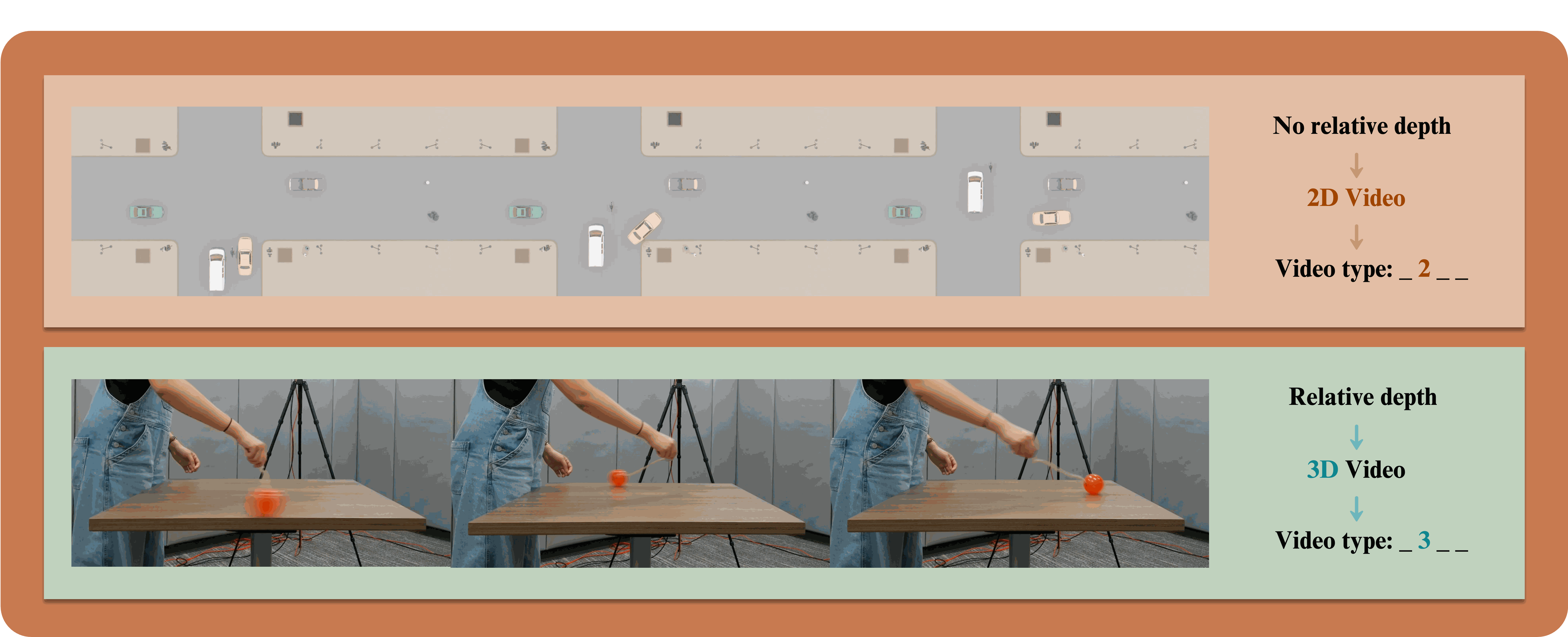}
    \caption{\textbf{Examples of 2/3 scene.}}
    \label{fig:2/3_scene}
\end{figure*}

\noindent\textbf{Second character (\texttt{2} / \texttt{3}).}
The second character indicates whether the video is 2-Dimensional(\texttt{2D}) or 3-Dimensional(\texttt{3D}).
\begin{itemize}[leftmargin=1.6em,topsep=0.2em,itemsep=0pt]
  \item \texttt{2D} = planar video
  \item \texttt{3D} = volumetric video
\end{itemize}
More precisely, the second character is set to \texttt{2} when, at every frame, the distances from the moving object, the ground-truth reference object, and every object referenced in the inference question to the camera are exactly equal, so they occupy a single depth layer with no relative depth or parallax among them in the rendered image, and to \texttt{3} otherwise. Video examples are shown in \autoref{fig:2/3_scene}.

\begin{figure*}[ht!]
    \centering
    \includegraphics[width=0.99\linewidth]{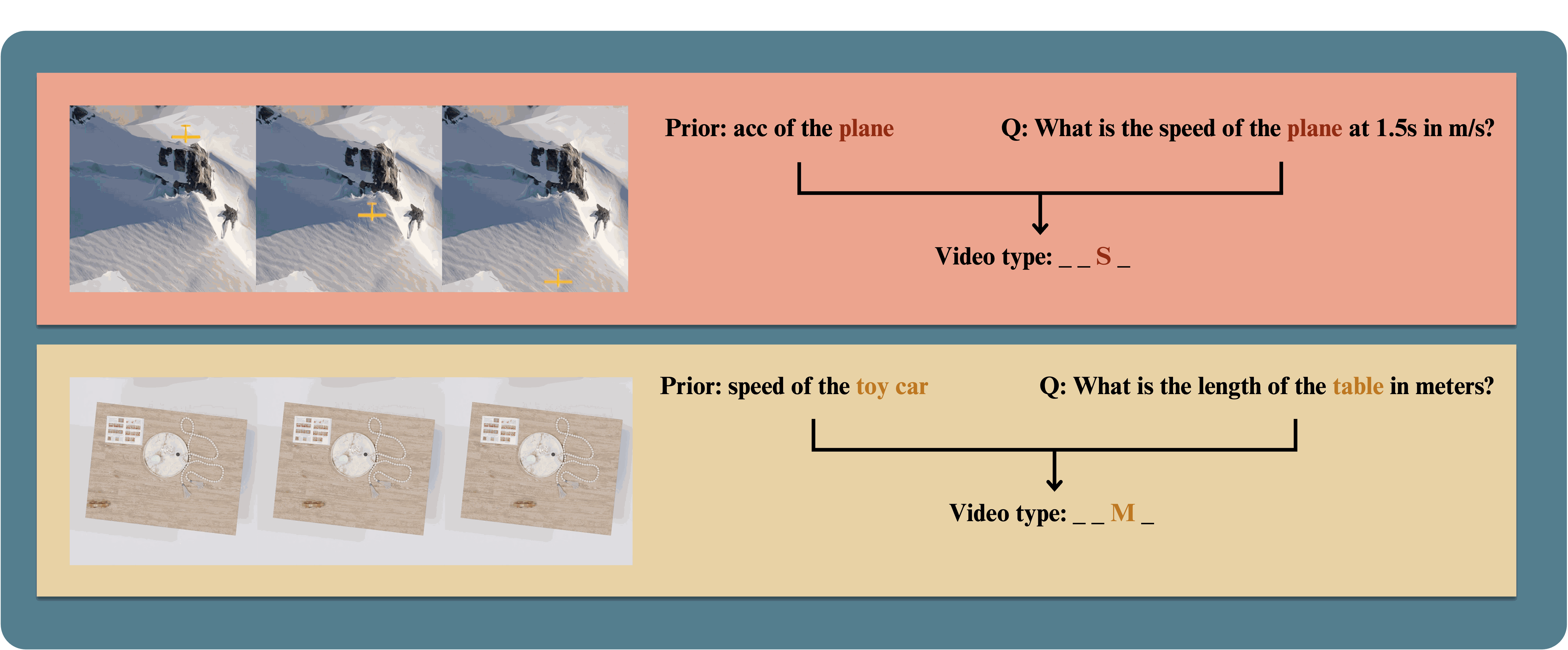}
    \caption{\textbf{Examples of S/M scene.}}
    \label{fig:s/m_scene}
\end{figure*}

\noindent\textbf{Third character (\texttt{S} / \texttt{M}).}
The third character distinguishes between “single-object” and “multiple-object” settings, relative to the queried object(s) rather than the sheer number of objects visible in the scene:
\begin{itemize}[leftmargin=1.6em,topsep=0.2em,itemsep=0pt]
  \item \texttt{S} = single-object
  \item \texttt{M} = multiple-object
\end{itemize}
As shown in \autoref{fig:s/m_scene}, this label depends on how many objects the viewer needs to reason about. We use \texttt{S} (single-object) when the reasoning process only involves one object. Other objects may appear, but they only serve as background or generic distractors. We use \texttt{M} (multiple-object) when answering any of the questions requires reasoning about two or more objects, for example by comparing their sizes or speeds, or by using one object as an explicit reference for another.

\begin{figure*}[ht!]
    \centering
    \includegraphics[width=0.99\linewidth]{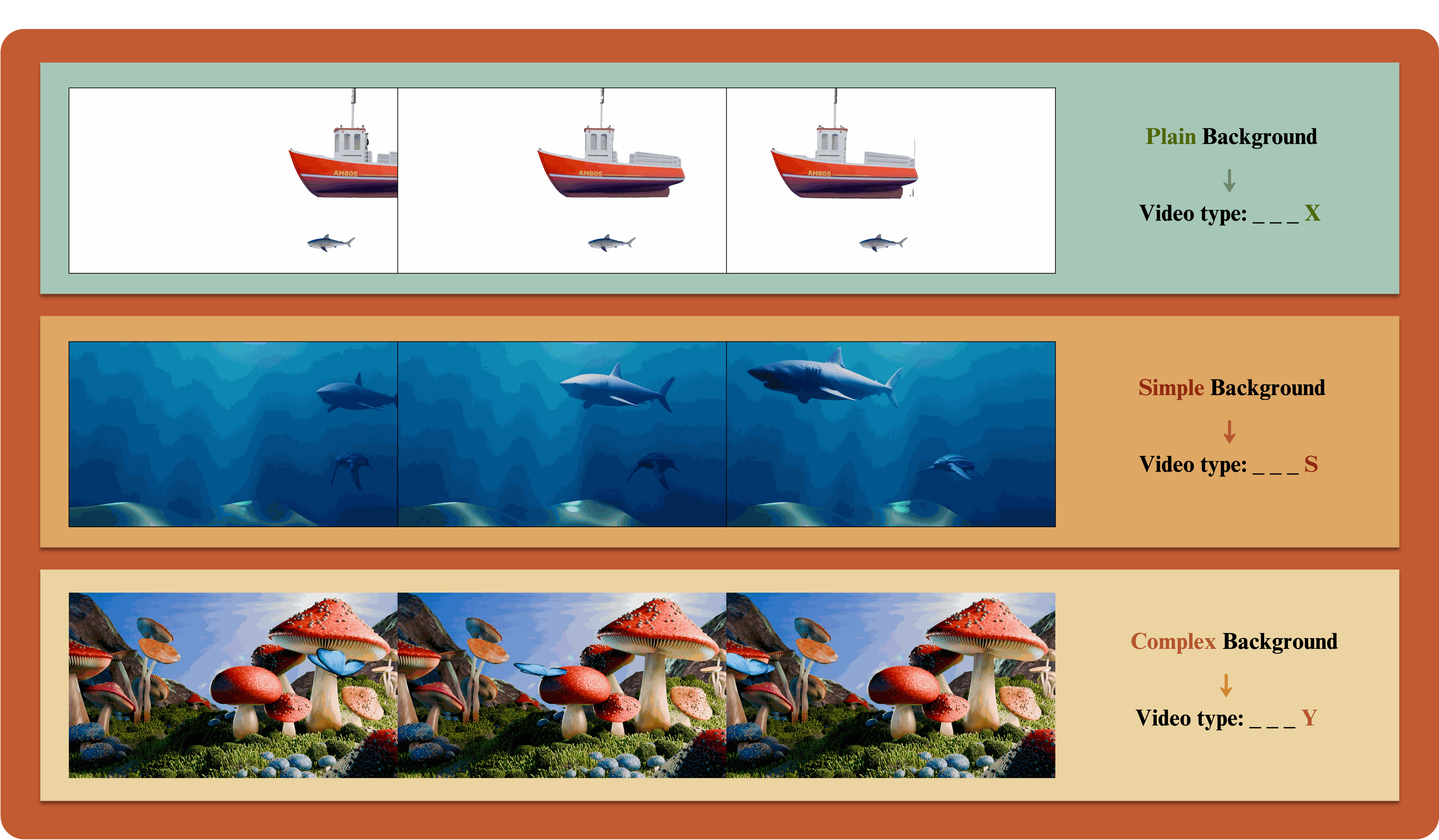}
    \caption{\textbf{Examples of X/S/C scene.}}
    \label{fig:X/S/C_scene}
\end{figure*}

\noindent\textbf{Fourth character (\texttt{X} / \texttt{S} / \texttt{C}).}
The fourth character's examples are illustrated in the Figure~\ref {fig:X/S/C_scene}. encodes the complexity of the background.
\begin{itemize}[leftmargin=1.6em,topsep=0.2em,itemsep=0pt]
  \item \texttt{X} = plain background, typically a single uniform RGB color with essentially no texture, clutter, or noise;
  \item \texttt{S} = simple background, which may contain mild lighting or shading variations but remains visually uncluttered;
  \item \texttt{C} = complex background, with rich textures, multiple visible objects, more intricate lighting, and substantial visual ``noise''.
\end{itemize}
The boundary between “simple” and “complex” backgrounds is somewhat subjective, but during dataset construction we deliberately separated these categories and designed scenes so that their visual difference is as clear as possible.

Putting these components together, for example, a code such as \texttt{A3MC} indicates that the video (i) targets the acceleration prior (\texttt{A}), (ii) is rendered as a 3D video (\texttt{3}), (iii) is labeled as a multiple-object setting (\texttt{M}) because at least one inference question asks about an object different from the one whose acceleration prior is defined, and (iv) uses a complex background.

\subsubsection{Quantitative breakdown of video types.}
To complement the qualitative description above, we now provide a quantitative summary of how clips are distributed across the four-character codes. The benchmark contains 569 videos in total, of which 328 are \texttt{2D} and 241 are \texttt{3D}, yielding an approximately 4:3 split between planar and volumetric setups (about 58\% \texttt{2D} and 42\% \texttt{3D}). This ensures that both \texttt{2D} and \texttt{3D} configurations are substantially represented rather than the dataset being dominated by a single family of scenes. \autoref{tab:2d3d-summary} reports the corresponding counts for each individual code and for the \texttt{2D} versus \texttt{3D} groups.
By construction, combining three physical priors (\texttt{S} / \texttt{V} / \texttt{A}), two dimensionalities (\texttt{2} / \texttt{3}), two object settings (\texttt{S} / \texttt{M}), and three background types (\texttt{X} / \texttt{S} / \texttt{C}) yields 36 distinct four-character codes, and all 36 appear in the dataset. For \texttt{2D} videos, the codes are
\texttt{A2SX, A2SS, A2SC, A2MX, A2MS, A2MC,
S2SX, S2SS, S2SC, S2MX, S2MS, S2MC,
V2SX, V2SS, V2SC, V2MX, V2MS, V2MC};
for \texttt{3D} videos, the codes are
\texttt{A3SX, A3SS, A3SC, A3MX, A3MS, A3MC,
S3SX, S3SS, S3SC, S3MX, S3MS, S3MC,
V3SX, V3SS, V3SC, V3MX, V3MS, V3MC}.
Each code is instantiated by at least 4 clips, so even the smallest categories have non-trivial support. The largest corpus (\texttt{V2MC} in our current dataset) contains 51 clips. Across all 36 codes, per-code counts range from 4 to 51 clips, with the majority lying between 5 and 35. This distribution avoids both extremely rare ``one-off'' configurations and a few overwhelmingly frequent ones. The precise per-code counts are given in \autoref{tab:2d3d-summary}.
\autoref{tab:2d3d-summary} also details how each video is obtained. The \texttt{Blender} column counts fully synthetic clips rendered directly in \texttt{Blender} from our own scenes, contributing 300 videos. The \texttt{Internet} column counts 72 clips sourced from existing online footage that we curate and annotate. The \texttt{Captured} column counts 112 clips that we record ourselves (for example, using handheld cameras or screen recordings). The \texttt{Segmented} column counts 85 clips created by segmenting foreground objects from source footage and compositing them into new backgrounds. For each four-character code, the Total column gives the number of clips that realize that configuration. Taken together, these statistics in \autoref{tab:2d3d-summary} show that the dataset spans all intended configurations, with each category populated by multiple clips and supported by a mix of blender-rendered, internet-sourced, captured, and segmented videos.

\begin{table*}[t]
\centering
\small
\renewcommand{\arraystretch}{1.1}
\begin{tabular}{l | l r r r r r r}
\toprule
2D/3D & \texttt{Video Type} & \texttt{Blender} & \texttt{Internet} & \texttt{Captured} & \texttt{Segmented} & \texttt{Total} & 2D/3D Total \\
\midrule

2D & \texttt{A2SX} & 0 & 0 & 0 & 11 & 11 &  \\
   & \texttt{A2SS} & 10 & 6 & 0 & 0  & 16 &  \\
   & \texttt{A2SC} & 10 & 5 & 0 & 0  & 15 &  \\
   & \texttt{A2MX} & 6 & 0 & 1 & 8  & 15 &  \\
   & \texttt{A2MS} & 14 & 0 & 1 & 0 & 15 &  \\
   & \texttt{A2MC} & 17 & 2 & 1 & 0 & 20 &  \\
   & \texttt{S2SX} & 0 & 0 & 0 & 10 & 10 & \\
   & \texttt{S2SS} & 10 & 5 & 0 & 0 & 15 & \\
   & \texttt{S2SC} & 7 & 8 & 0 & 0 & 15 & \\
   & \texttt{S2MX} & 8 & 0 & 0 & 8 & 16 & \\
   & \texttt{S2MS} & 11 & 3 & 0 & 0 & 14 & \\
   & \texttt{S2MC} & 16 & 20 & 0 & 0 & 36 & \\
   & \texttt{V2SX} & 2 & 0 & 0 & 9 & 11 & \\
   & \texttt{V2SS} & 10 & 7 & 0 & 0 & 17 & \\
   & \texttt{V2SC} & 10 & 8 & 0 & 0 & 18 & \\
   & \texttt{V2MX} & 13 & 0 & 0 & 7 & 20 & \\
   & \texttt{V2MS} & 13 & 0 & 0 & 0 & 13 & \\
   & \texttt{V2MC} & 43 & 8 & 0 & 0 & 51 & \\
\hline
   &       &   &   &   &   &    & 328 \\
\midrule

3D & \texttt{A3SX} & 2 & 0 & 6 & 1  & 9  &  \\
   & \texttt{A3SS} & 2 & 0 & 9 & 0  & 11 &  \\
   & \texttt{A3SC} & 3 & 0 & 7 & 0  & 10 &  \\
   & \texttt{A3MX} & 1 & 0 & 4 & 2  & 7  &  \\
   & \texttt{A3MS} & 3 & 0 & 4 & 0  & 7  &  \\
   & \texttt{A3MC} & 22 & 0 & 4 & 0 & 26 &  \\
   & \texttt{S3SX} & 1 & 0 & 8 & 2  & 11 & \\
   & \texttt{S3SS} & 2 & 0 & 8 & 0  & 10 & \\
   & \texttt{S3SC} & 2 & 0 & 8 & 0  & 10 & \\
   & \texttt{S3MX} & 0 & 0 & 9 & 13 & 22 & \\
   & \texttt{S3MS} & 1 & 0 & 10 & 0 & 11 & \\
   & \texttt{S3MC} & 24 & 0 & 10 & 0 & 34 & \\
   & \texttt{V3SX} & 3 & 0 & 2 & 0  & 5  &  \\
   & \texttt{V3SS} & 2 & 0 & 2 & 0  & 4  &  \\
   & \texttt{V3SC} & 3 & 0 & 2 & 0  & 5  &  \\
   & \texttt{V3MX} & 2 & 0 & 5 & 14 & 21 & \\
   & \texttt{V3MS} & 3 & 0 & 5 & 0  & 8  &  \\
   & \texttt{V3MC} & 24 & 0 & 6 & 0 & 30 & \\
\hline
   &       &   &   &   &   &    & 241 \\
\midrule
Total &      & 300 & 72 & 112 & 85 & 569 & 569 \\
\bottomrule
\end{tabular}
\caption{\textbf{Statistics of videos}.}
\label{tab:2d3d-summary}
\end{table*}

\section{Details of Data Collection}

\subsection{Blender Simulation}

\subsubsection{Blender Toolkits and Asset Sources}
\label{subsec:blender_toolkits}

\noindent\textbf{Asset sources and selection.}
We construct Blender scenes using 3D assets sourced from online repositories,
primarily BlenderKit and Sketchfab. We deliberately use these two libraries for
complementary purposes: BlenderKit mainly provides complex themed environments
that serve as background layouts, whereas Sketchfab mainly provides rigged and
animated foreground objects whose motions can be reused with minimal manual
keyframing.

\noindent\textbf{BlenderKit for complex themed environments.}
BlenderKit is a community-driven asset library that is tightly integrated into
Blender’s interface and offers a large collection of render-ready models,
materials, High Dynamic Range Images(HDRIs), brushes, and complete scenes that can be searched and
inserted directly from within Blender. In our pipeline, we primarily rely on
BlenderKit for complex “themed” environments, such as indoor rooms, streets, architectural spaces, and other richly cluttered layouts at both small and large spatial scales. These pre-built scenes typically include coherent lighting, materials, and background geometry (for example, furniture, buildings, and vegetation), which allows us to quickly instantiate diverse indoor and outdoor environments without having to model every object or compose every layout from scratch. This substantially reduces the authoring effort for environment design while still giving us control over camera placement, object insertion, and motion trajectories.

\noindent\textbf{Sketchfab for rigged and animated foreground objects.} Sketchfab is a large community platform for hosting and distributing 3D content, including many rigged and animated models across categories such as animals, vehicles, and articulated characters. In our work, we mainly use Sketchfab to obtain foreground objects that already come with a skeleton rig and a small set of reusable animation clips. Typical examples include wing-flapping cycles for flying birds (such as eagles spreading and flapping their wings) and swimming cycles for fish with realistic body undulation. Instead of manually keyframing these motions, we can directly reuse or lightly retarget the provided animations in our scenes. This makes it much easier to populate environments with moving agents whose motion is visually plausible, while significantly reducing the time spent on low-level animation authoring.

\noindent\textbf{Licensing and reuse.} For both BlenderKit and Sketchfab, we restrict ourselves to assets whose licenses explicitly allow reuse and modification in works. These licensing choices ensure that all assets in our dataset are used in a copyright-compliant way and that future researchers can reconstruct our pipeline using the same publicly available resources.

\begin{figure*}[h]
    \centering
    \includegraphics[width=\linewidth]{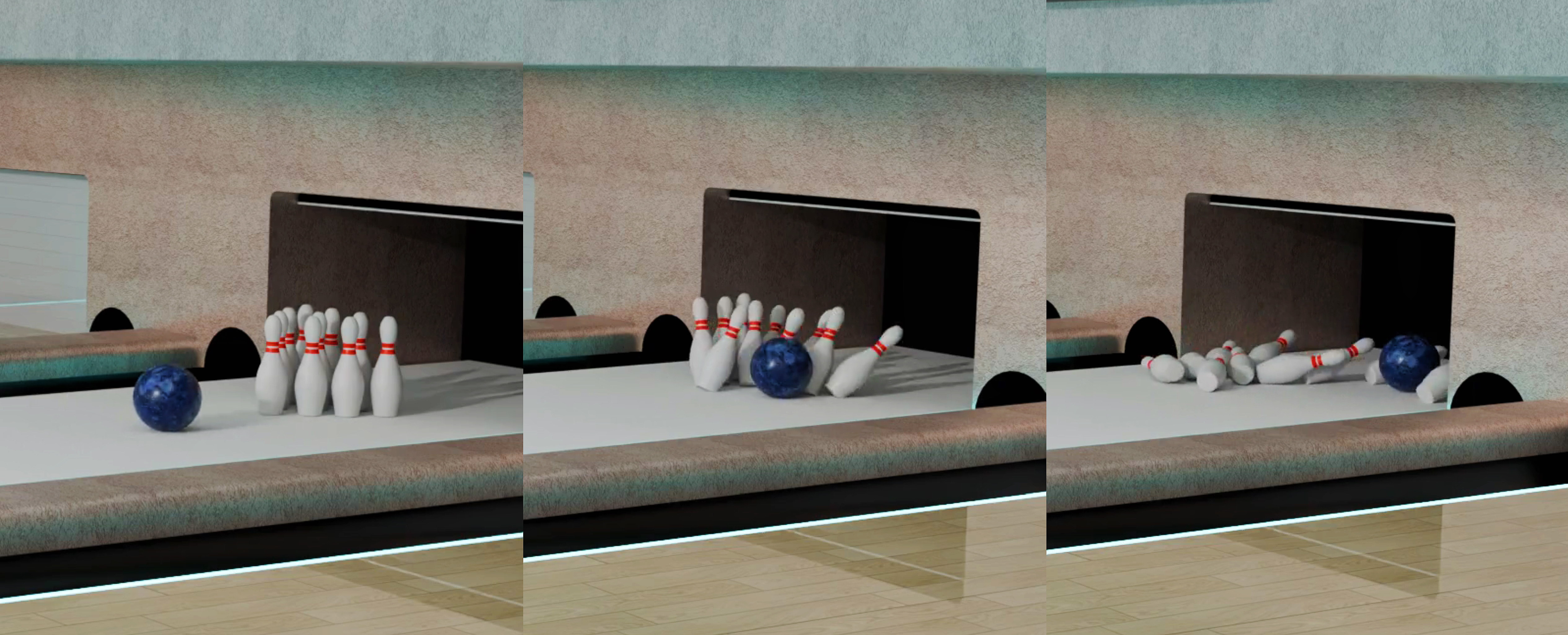}
    \caption{\textbf{Physics-driven example.} 
    A bowling ball collides with pins under Newtonian simulation. 
    The motion and resulting trajectories arise directly from rigid-body dynamics and elastic collisions, making this clip representative of our force-based motion category.}
    \label{fig:physics_example}
\end{figure*}

\begin{figure*}[h]
    \centering
    \includegraphics[width=\linewidth]{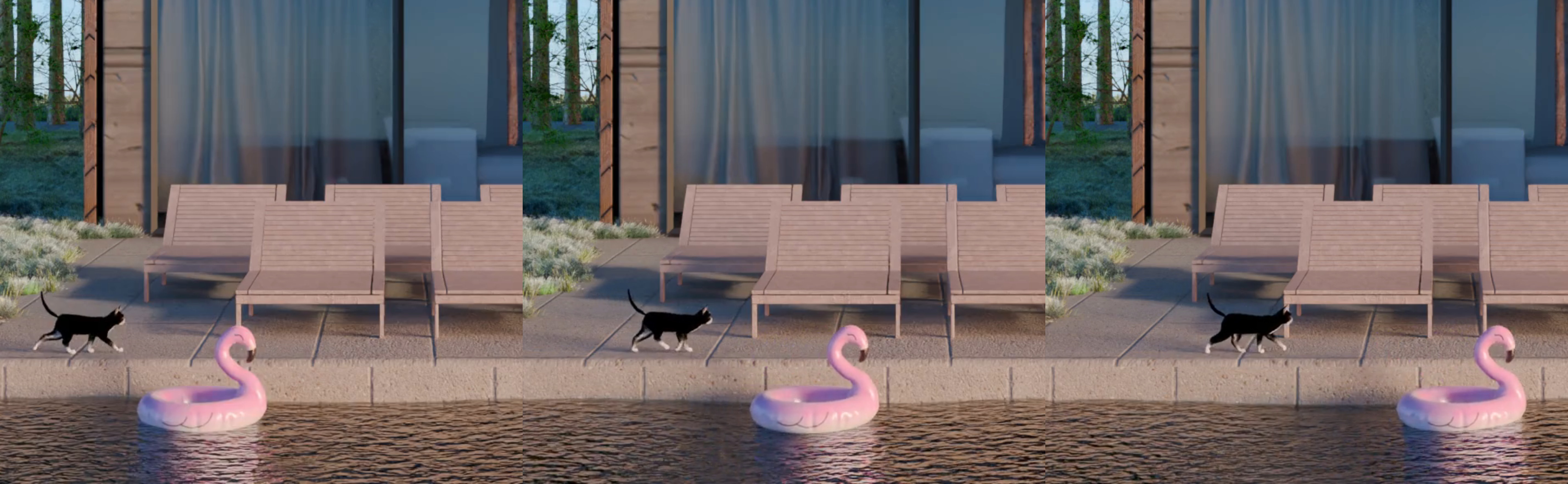}
    \caption{\textbf{Keyframed example.} 
    The floating swim ring follows a manually authored animation curve rather than buoyancy, drag, or wind forces. 
    Its trajectory is visually plausible but not physically derived, representing our keyframed motion category.}
    \label{fig:keyframe_example}
\end{figure*}

\begin{figure*}[h]
    \centering
    \includegraphics[width=\linewidth]{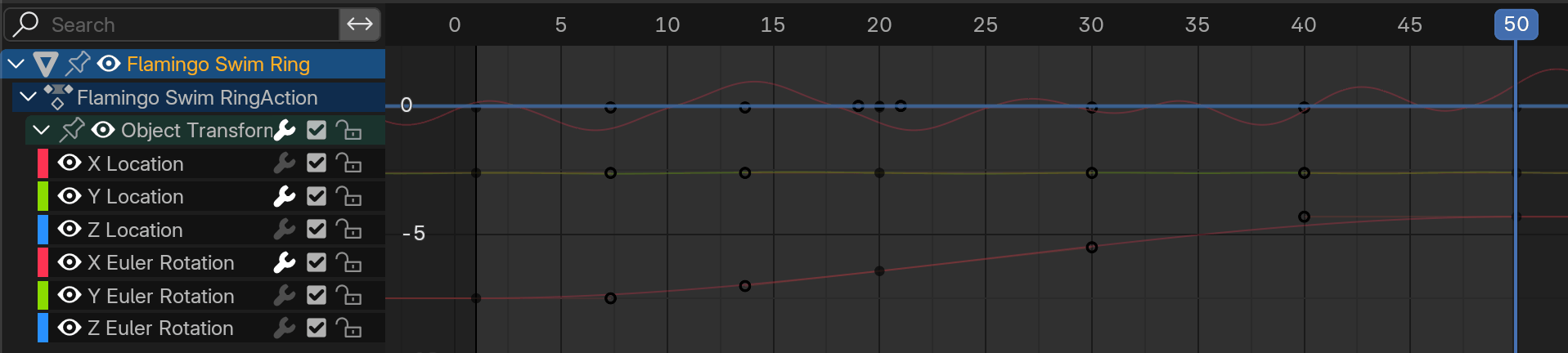}
    \caption{\textbf{Keyframed animation-curve example.} 
    The animation curve controlling the swim ring in \autoref{fig:keyframe_example}. 
    The ring moves forward along a manually authored trajectory, while small randomized perturbations in translation and rotation are added to imitate the visual appearance of floating motion.}
    \label{fig:animation_curve_example}
\end{figure*}

\begin{figure}[h]
    \centering
    \includegraphics[width=\linewidth]{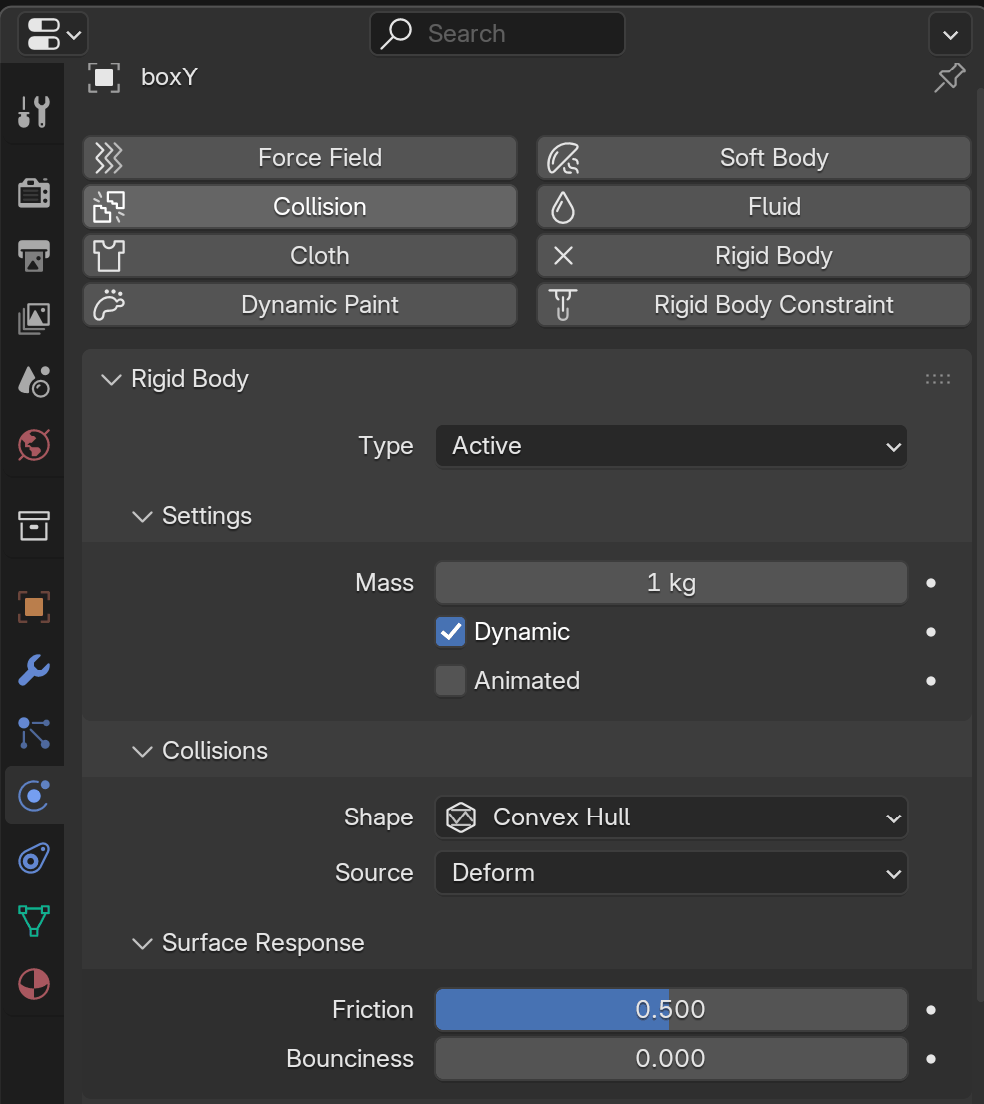}
    \caption{\textbf{Physics-driven rigid-body example.}
    Rigid-body simulation of objects such as the bowling pins reacting to the impact of an incoming ball in \autoref{fig:physics_example}.
    The pins' motion, tipping, and scattering are governed entirely by Newtonian dynamics and collision responses within the physics engine.}
    \label{fig:rigid_body_example}
\end{figure}

\subsubsection{Two Motion Simulation Types}
\label{subsec: MotionType}
The Blender-generated portion of our dataset contains two complementary categories of motion, reflecting the major paradigms of movement in computer graphics and physics-based animation.

\noindent\textbf{(1) Keyframed Motion.} 
Examples of this category are shown in \autoref{fig:keyframe_example}.
This category includes humans, animals, and other objects whose motion is defined using rigged skeletons or keyframed animation curves shown in \autoref{fig:animation_curve_example}. Because these trajectories are authored manually rather than produced through physical simulation, they are {visually plausible but not physically constrained}.

A key implication is that the resulting motion does not necessarily obey real-world physical laws. For example, in the lunar-walking scene (\autoref{fig:astronaut_scene}), the astronaut’s push-off, airtime, and landing motion are shaped by artist-edited animation curves. Although the motion is loosely inspired by reduced lunar gravity, we do not compute the animation using the Moon's gravitational constant nor derive the trajectory from force-based simulation. These sequences should therefore be interpreted as {perceptually reasonable approximations} of motion rather than physically calibrated ground truth.

\noindent\textbf{(2) Physics-Driven / Force-Based Motion.}
Examples of this category are shown in \autoref{fig:physics_example}.
This category consists of rigid bodies (shown in \autoref{fig:rigid_body_example}) moving directly under Newtonian dynamics. Their trajectories are generated by applying explicit forces (e.g., gravity, impulses) inside a physics engine or by specifying analytical kinematic profiles (e.g., constant velocity, uniform acceleration, projectile motion). These clips yield clean and physically interpretable motion, allowing us to provide exact ground-truth displacement, velocity, and acceleration.

For all videos in this category, we explicitly state the forces or kinematic parameters involved, and these values constitute part of the known prior information for each question.

\noindent\textbf{Clarifying Prior Knowledge vs.\ Visual Assumptions.}
In an era where videos may originate from real capture, simulation software, procedural animation, or generative models, visual motion alone does not guarantee adherence to physical laws. Accordingly, VLMs should reason strictly based on the prior conditions we provide, rather than relying on pretrained assumptions about how objects “should” move.

When prior information explicitly specifies a force or acceleration (e.g., “the object accelerates at $2.5\,\mathrm{m/s^2}$”), that value serves as authoritative ground truth. When such information is not given, the model should not infer physical constants, such as Earth's gravity, only from the visual appearance of the motion. Because our dataset includes diverse simulated and animated sequences, correct reasoning requires using only the provided priors, not assumed real-world physics.

\subsubsection{Blender Videos Construction}
\label{subsec:blendervideo_construction}

\noindent\textbf{Quantitative Overview.} In the Blender subset, we start from approximately 125 distinct base 3D models and 81 base scenes. From these bases, we generate 312 unique Blender video clips. Beyond simply increasing the size of the dataset, Blender enables us to construct diverse scenes and controlled trajectories of objects in simplified environments. Because these scenes are generated procedurally, we can precisely specify object sizes, positions, and kinematics via scripts and directly compute accurate ground-truth physical quantities for each clip. As a result, the Blender subset is not only large, but also covers a broad spectrum of scene scales and motions under tight experimental control.

\begin{figure*}[ht!]
    \centering
    \includegraphics[width=0.99\linewidth]{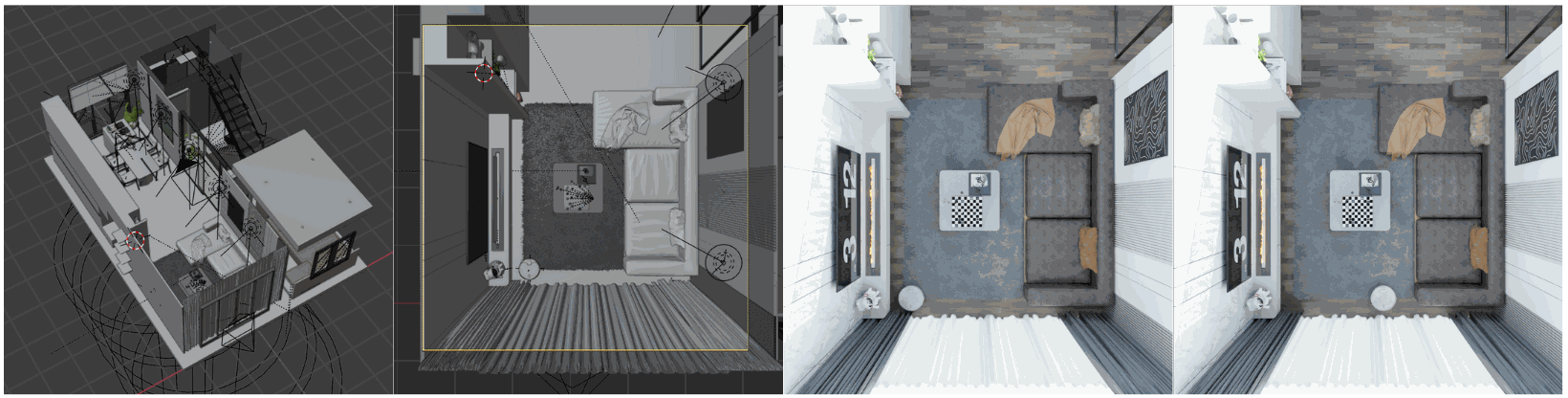}
    \caption{\textbf{Examples of indoor scene.}}
    \label{fig:indoor_scene}
\end{figure*}

\begin{figure*}[ht!]
    \centering
    \includegraphics[width=0.99\linewidth]{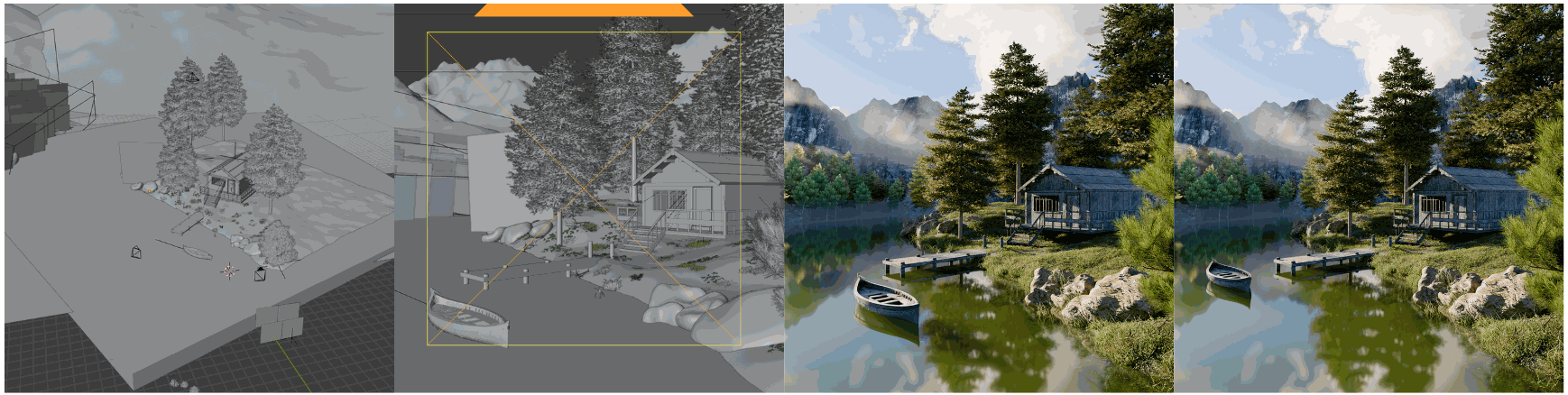}
    \caption{\textbf{Examples of outdoor scene.}}
    \label{fig:outdoor_scene}
\end{figure*}

\noindent\textbf{Scene Construction.} We construct a diverse collection of synthetic scenes to broaden the range of physical situations represented in the benchmark. Specifically, we design both everyday indoor spaces (e.g., rooms with furniture and household objects, see \autoref{fig:indoor_scene}) and outdoor spaces (e.g., natural environments with varied terrain, vegetation, and water, see \autoref{fig:outdoor_scene}). These scenes differ in layout, depth structure, and illumination, so that models must handle physical reasoning under heterogeneous visual conditions rather than overfitting to a single canonical setting.

\begin{figure*}[ht!]
    \centering
    \includegraphics[width=0.99\linewidth]{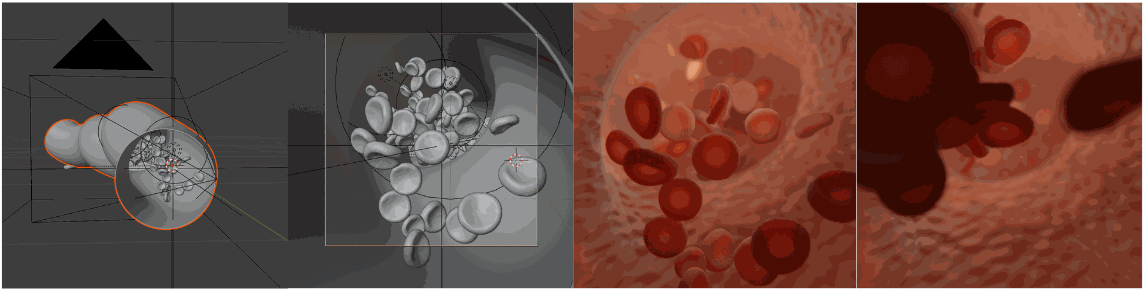}
    \caption{\textbf{Examples of microscopic scene.}}
    \label{fig:microscopic_scene}
\end{figure*}

\begin{figure*}[ht!]
    \centering
    \includegraphics[width=0.99\linewidth]{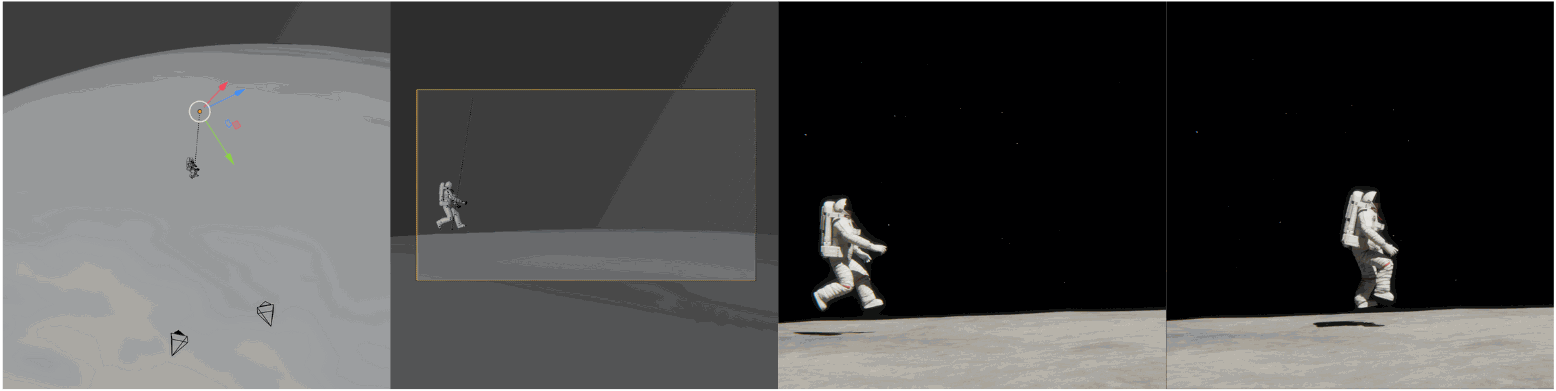}
    \caption{\textbf{Examples of extraterrestrial scene.}}
    \label{fig:astronaut_scene}
\end{figure*}

Beyond such everyday environments, we also include scenes that are difficult or impossible to capture, measure, or systematically manipulate in the real world. Examples include microscopic settings (e.g., red blood cells moving through a vessel, see \autoref{fig:microscopic_scene}), extraterrestrial scenarios (e.g., astronaut motion on the Moon, see \autoref{fig:astronaut_scene}), and more. 

\begin{figure*}[ht!]
    \centering
    \includegraphics[width=0.99\linewidth]{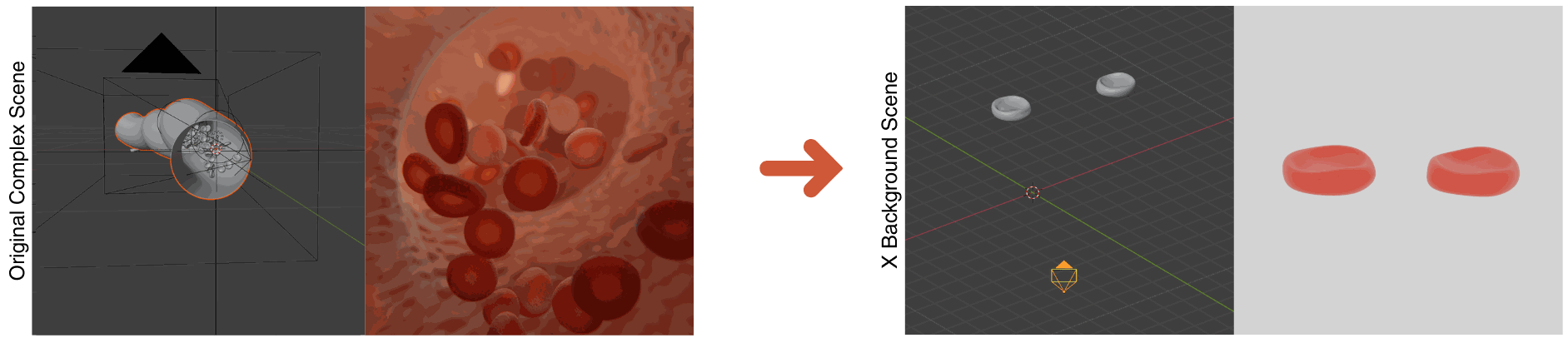}
    \vspace{-0.5em}
    \caption{\textbf{Examples of building X background scene.}}
    \label{fig:building_x_scene}
\end{figure*}

\begin{figure*}[ht!]
    \centering
    \includegraphics[width=0.99\linewidth]{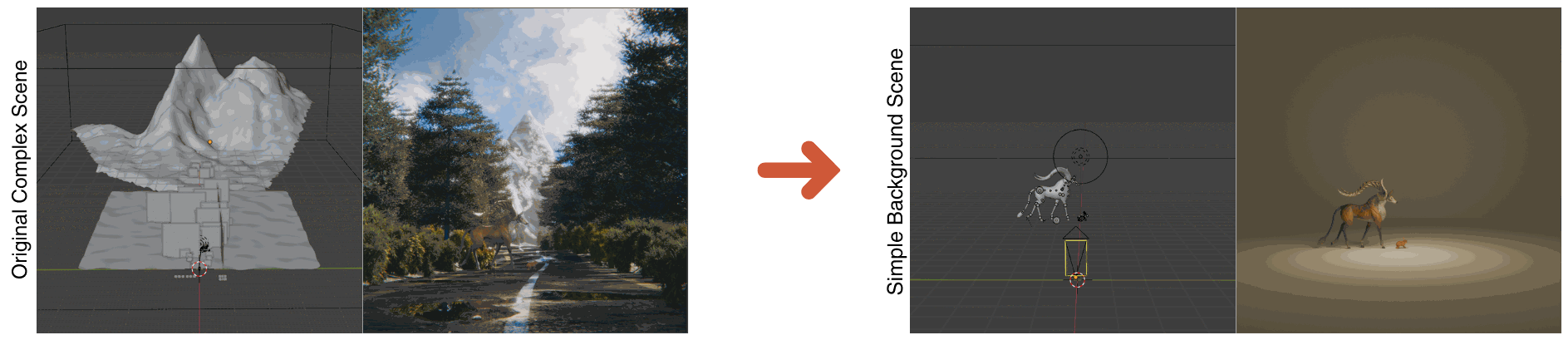}
    \vspace{-0.5em}
    \caption{\textbf{Examples of building S background scene.}}
    \label{fig:building_s_scene}
\end{figure*}

Starting from base assets and scenes, we explicitly control geometry, lighting, and background complexity to generate families of videos that fall into different categories in our taxonomy. For instance, in the red-blood-cell scene (\autoref{fig:building_x_scene}), we derive a 3MX-style variant by removing the vessel wall, discarding all but two target cells, and replacing the background with a uniform RGB field. Using the same procedure, as shown in the \autoref{fig:building_s_scene} we construct simple-background versions of more complex scenes by simplifying clutter while keeping coarse structural cues (e.g., horizon lines, major surfaces, object structures) and realistic lighting. 

It is important to note that, although we present several examples that appear to reuse the same underlying scene. In practice, not all plain-background videos and simple-background videos are reused across conditions. In addition, we also include videos that are uniquely constructed to appear only in the plain-background condition or only in the simple-background condition. This design choice is intended to increase the diversity of the video data. Reusing identical base scenes improves the efficiency of dataset construction, but it may also introduce potential biases or unwanted correlations when evaluating VLMs, so we deliberately balance such reuse with the creation of novel, non-paired scenes.

For the objects in each scene, we specify target object sizes, velocities, and accelerations using real-world statistics gathered from online references (e.g., typical dimensions and speeds of vehicles, animals, and more). Whenever possible, moving objects are modeled at real-world scale; in rare cases where real-world-scale objects yield motion trajectories that are barely discernible in the rendered videos, we apply uniform scaling to increase perceptual visibility while preserving the underlying physical relationships.

\begin{figure*}[ht!]
    \centering
    \includegraphics[width=0.99\linewidth]{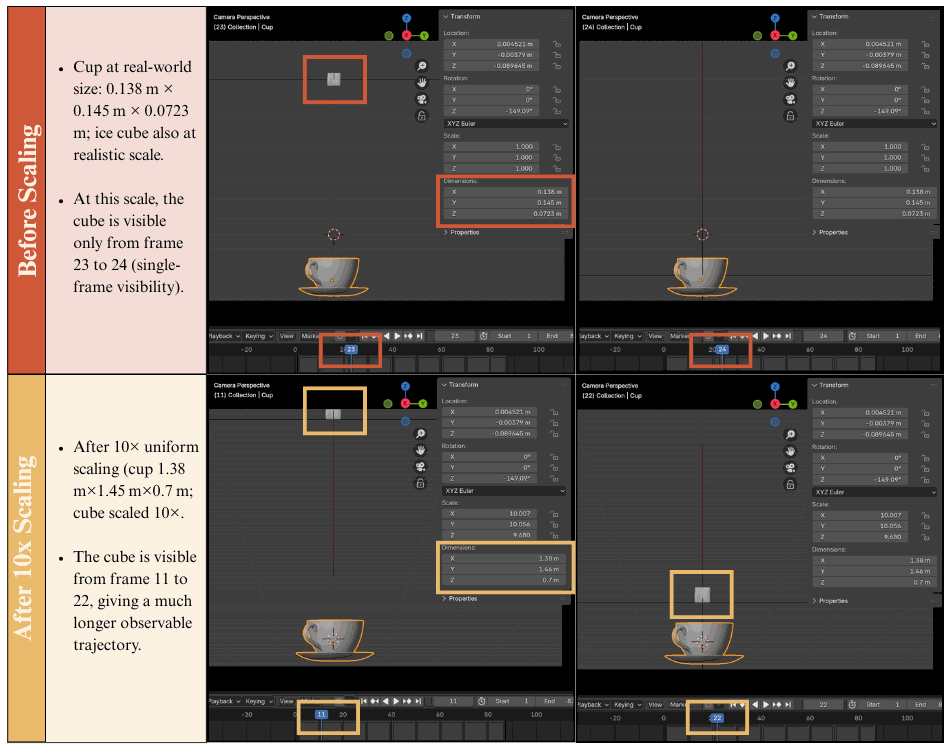}
    \caption{\textbf{Examples of scaled scene.}}
    \label{fig:scaled_scene}
\end{figure*}

An example is the “ice cube falling into a cup” scene illustrated in the \autoref{fig:scaled_scene}. If we model the cup and cube at their real-world dimensions, the cube traverses the camera frustum in essentially a single frame, making its falling trajectory almost invisible to observers and thus providing little signal for quantitative evaluation. To address this, we uniformly enlarge both the cup and the ice cube by a factor of 10. After scaling, the cube remains visible from frame 11 to 22, and the rendered video reveals a clear multi-frame trajectory. This controlled rescaling not only makes the motion measurable but also allows us to probe whether VLMs recover physical quantities from the observed kinematics given the priors specified in the prompt, or whether they instead rely predominantly on generic pre-trained knowledge and commonsense expectations about how such objects “should” behave.

\noindent\textbf{Object Motion Design and Implementation.}
As discussed in \autoref{subsec: MotionType}, our motion design follows two main paradigms: keyframed motion and physics-driven motion. The construction of physics-driven scenes is illustrated in \autoref{fig:physics_example}. For the keyframed cases, in addition to manual editing in the Graph Editor, we also use scripts to automate the animation process.

We distinguish two main classes of scripted motion: (i) analytic one-dimensional motion along a single axis, and (ii) curve-following motion along human-designed paths.

For analytic 1D motion, we explicitly encode standard kinematic equations and bake the resulting trajectories as keyframes. Given an object with initial world-space position $\mathbf{x}_0$, a target duration $T$, and a desired acceleration $a$ along the $-Y$ axis, our script iterates over frames $f$ and computes the physical time
\[
t = \frac{f - \texttt{START\_FRAME}}{\texttt{FPS}}.
\]

The corresponding displacement is computed using the constant-acceleration formula
\[
s(t) = \tfrac{1}{2} a t^{2}.
\]

\begin{minted}[
  fontsize=\footnotesize,
  linenos,
  frame=lines,
  framesep=2mm,
  baselinestretch=1.0,
  breaklines,
  bgcolor=black!2
]{python}
disp_y = 0.5 * ACCEL * (t ** 2)
new_y  = start_loc.y - disp_y
obj.location = Vector((start_loc.x, new_y, start_loc.z))
obj.keyframe_insert(data_path="location", index=-1, frame=f)
\end{minted}


This directly realizes $s(t)=\tfrac{1}{2} a t^{2}$ in world coordinates. For constant-velocity motion, we instead use the linear relation:
\[
s(t) = v t,
\]
implemented via

\begin{minted}[
  fontsize=\footnotesize,
  linenos,
  frame=lines,
  framesep=2mm,
  baselinestretch=1.0,
  breaklines,
  bgcolor=black!2
]{python}
disp_y = VEL * t
new_y  = start_loc.y - disp_y
\end{minted}

while keeping the same frame loop and keyframe insertion logic. After all keyframes are written, we programmatically set each F-curve’s interpolation mode to \texttt{LINEAR}, overriding Blender’s default Bézier interpolation to ensure that the frame-to-frame displacement matches the analytically specified velocity or acceleration profile rather than being inadvertently smoothed or distorted.

For more complex motions along curved paths, we first author an approximate trajectory in Blender (e.g., by manually keyframing a car following a road or a particle moving along a spiral), and then reparameterize this path in Python to obtain physically interpretable kinematics. Concretely, we sample the object’s world-space position at each frame of the original animation:

\begin{minted}[
  fontsize=\footnotesize,
  linenos,
  frame=lines,
  framesep=2mm,
  baselinestretch=1.0,
  breaklines,
  bgcolor=black!2
]{python}
positions = []
for f in range(frame_start, frame_end + 1):
    scene.frame_set(f)
    pos = obj.matrix_world.translation.copy()
    positions.append(pos)
\end{minted}


We then compute the cumulative arc length along this polyline.

\begin{minted}[
  fontsize=\footnotesize,
  linenos,
  frame=lines,
  framesep=2mm,
  baselinestretch=1.0,
  breaklines,
  bgcolor=black!2
]{python}
distances = [0.0]
for i in range(1, len(positions)):
    d = (positions[i] - positions[i-1]).length
    distances.append(distances[-1] + d)
total_length = distances[-1]
duration     = (frame_end - frame_start) / fps
\end{minted}
\noindent We then derive a desired kinematic profile in terms of traveled distance $s(t)$ along the curve. For constant-speed motion, we set $v = L/T$ with $L = \texttt{total\_length}$ and use
\[
s(t) = v t,
\]
implemented as

\begin{minted}[
  fontsize=\footnotesize,
  linenos,
  frame=lines,
  framesep=2mm,
  baselinestretch=1.0,
  breaklines,
  bgcolor=black!2
]{python}
target_dist = speed * t
\end{minted}
\noindent which corresponds to s(t)=vt. For constant-acceleration motion along the same curve, we instead use the quadratic form
\[
s(t) = v_0 t + \tfrac{1}{2} a t^{2},
\]
with user-specified $v_0$ and $a$. In both cases, for each frame we find the segment $[i-1,i]$ such that
\[
\texttt{distances}[i-1] \le s(t) \le \texttt{distances}[i],
\]
and linearly interpolate between the sampled positions:

\begin{minted}[
  fontsize=\footnotesize,
  linenos,
  frame=lines,
  framesep=2mm,
  baselinestretch=1.0,
  breaklines,
  bgcolor=black!2
]{python}
ratio  = (target_dist - distances[i-1]) / (distances[i] - distances[i-1])
new_loc = positions[i-1].lerp(positions[i], ratio)
obj.location = new_loc
obj.keyframe_insert(data_path="location", frame=f)
\end{minted}

Finally, as in the analytic case, all resulting keyframes are set to linear interpolation. This arc-length reparameterization scheme lets us reuse authored trajectories while imposing well-defined kinematic profiles (constant velocity, constant acceleration, or other time–distance schedules) in physical units. Together with the untouched native animations, these scripted motions give our benchmark a wide spectrum of motion patterns, from simple 1D acceleration to complex, irregular dynamics, under tight quantitative control.

\noindent\textbf{Rendering.} All Blender-generated videos in \textsc{QuantiPhy} are rendered with the Cycles and EEVEE path-tracing engine using a physically based workflow. Videos use multiple spatial resolutions, including $1920{\times}1080$ (16:9), $1080{\times}1080$ (1:1), $480{\times}960$ (vertical), as well as several additional intermediate sizes, so that models see both landscape, square, and portrait-style content. The temporal sampling is likewise heterogeneous: frame rates in the benchmark include $24$, $25$, $30$, $33$, $60$, and $120$fps. Frames are exported via FFmpeg as MP4 files with H.264 compression and a constant frame rate, without any interpolation or stabilization, so each frame aligns exactly with the simulated timeline and ground-truth annotations.

Camera parameters are also varied to increase visual diversity. Intrinsics cover a range of focal lengths from wide to normal and telephoto views, and extrinsics place the camera at different heights, offsets, and azimuth/elevation angles around the scene, producing both frontal and oblique perspectives on object motion. Lighting setups combine different environment maps and local light sources, and objects are assigned a broad set of physically based materials (e.g., diffuse, glossy, metallic, translucent), leading to diverse shading, reflection, and contrast conditions. Together, these choices give \textsc{QuantiPhy} wide coverage over resolutions, frame rates, viewpoints, and appearance statistics, while the underlying physical trajectories and numeric ground truth remain precisely controlled.

\begin{table*}[ht]
\centering
\begin{tabular}{@{}ll@{}}
\toprule
\textbf{Depth Technology}      & Time of Flight \\[3pt]
\textbf{Wavelength}            & 850\,nm \\[3pt]
\textbf{Depth Range}           & *0.25--5.46\,m (depending on depth mode) \\[3pt]
\textbf{Depth Resolution/FPS}  & Up to \(1024 \times 1024@15\) fps (WFOV), \(640 \times 576@30\) fps (NFOV) \\[3pt]
\textbf{Depth FOV}             & H \(120^\circ\) V \(120^\circ\) (WFOV), H \(75^\circ\) V \(65^\circ\) (NFOV) \\[3pt]
\textbf{RGB Resolution/FPS}    & Up to \(3840 \times 2160@25\) fps \\[3pt]
\textbf{RGB FOV}               & H \(80^\circ\) V \(51^\circ\) \\[3pt]
\textbf{Processing}            & NVIDIA Jetson Nano \\[3pt]
\textbf{IMU}                   & Supported \\
\bottomrule
\end{tabular}
\caption{\textbf{Specifications of the cameras used in our customized motion capture system.}}
\label{tab:camera}
\end{table*}

\begin{table}[ht]
\centering
\small
\setlength{\tabcolsep}{8pt}
\renewcommand{\arraystretch}{1.1}
\begin{tabular}{l l}
\toprule
Objects & Dimensions \\
\midrule
2 green tennis balls & d = 6.7 cm \\
2 pink tennis balls & d = 6.7 cm \\
2 white ping pong balls & d = 4 cm \\
1 purple yoga ball & d = 52.2 cm \\
1 soccer ball & d = 17.5 cm \\
1 basketball & d = 23.2 cm \\
1 red plastic ball & d = 7 cm \\
1 orange plastic ball & d = 5.7 cm \\
1 small whiteborad (slope) & 30.8 cm x 23.2 cm x 0.5 cm \\
1 large whiteborad & 81 cm x 60 cm x 1 cm \\
1 trashbin & 34.5 cm x 89.1 cm 59.9 cm \\
1 tape & 10.1 cm x 10.1 cm x 4.8 cm \\
1 white food box & 19.5 cm x 6.0 cm x 7.2 cm \\
stuffed toy 1 & 14.5 cm x 11 cm x 8 cm \\
stuffed toy 2 & 29.5 cm x 15.0 cm x 10.0 cm \\
stuffed toy 3 & 8.1 cm x 6.4 cm x 19.6 cm \\
1 toy cookie & 8.0 cm x 8.0 cm x 1.8 cm \\
1 cosmetic jar & 7.2 cm x 7.2 cm x 6.0 cm \\
1 glass jar & 6.8 cm x 6.8 cm x 8.5 cm \\
1 white cup & 8.6 cm x 8.6 cm x 14.5 cm \\
1 pink water bottle & 6.7 cm x 6.7 cm x 19.5 cm \\
1 marker & 13.5 cm x 0.8 cm x 0.8 cm \\
1 black pen & 14.5 cm x 1. 2 cm x 1.2 cm \\
1 green notebook & 18.6 cm x 8.8 cm x 1.5 cm \\
1 white-covered book & 12.5 cm x 19.5 cm x 2.2 cm \\
1 pencil case & 6.7 cm x 7.2 cm x 21 cm \\
1 credit card & 85.6 mm x 53.98 mm \\
1 deck of poker card & 9.8 cm x 6.3 cm x 1.8 cm \\
\bottomrule
\end{tabular}
\caption{\textbf{List of physical objects and their measured dimensions used in our lab-captured videos.}}
\label{tab:object-list}
\end{table}

\subsection{Lab Capturing}

To further complement the diversity of \textsc{QuantiPhy} with real-world data, a major part of our dataset comes from a customized motion capture system in a lab environment. Due to the nature of real-world data, lab captured videos contribute only to the category of 3D.

\noindent\textbf{MoCap setup.} 
We use four Orbbec Femto Mega cameras to construct our customized motion capture system. The specifications of the cameras are shown in \autoref{tab:camera}.

We designed two different camera setups and object arrangements in the MoCap system. The first setup covers a smaller spatial range, with the main camera placed close to the objects. In this setup, we capture motions of small objects such as a tennis ball, book, and ping pong ball moving on a desk. The second setup covers a larger range and is used to capture larger-scale motions such as basketball bouncing and accelerated motion of a trash bin. For both setups, we initialize the camera extrinsics via multi-view calibration with a checkerboard.

\noindent\textbf{Collection workflow.} 
For efficient video capture, we developed a tool with a user-friendly interface that supports synchronized recording across all four cameras. We collect videos based on a list of predefined questions. For each question, we set up the scene and objects without moving the cameras and then perform specific object motions according to the question. Three people were involved in the lab data collection workflow: one person was responsible for operating the tool (starting and pausing capture), and two people were responsible for setting up the scenes and executing the motions.

Unlike Blender simulation, capturing real-world data makes it difficult to directly manipulate specific scene parameters. We therefore manually configured the scenes to meet target conditions, such as the slope angle or the frequency of pendulum motion, using careful measurements within a set tolerance. For motions that require a controlled initial velocity, we used a motor running at s constant speed to pull the object and provide the desired initial velocity.

\noindent\textbf{Post-processing.} 
After capturing the raw videos, we obtain \texttt{.bag} files for each camera that contain the binary recordings. We follow the official camera SDK to decode the raw data. From this decoding, we obtain: (1) intrinsics for each camera, (2) timestamped RGB streams for each camera, (3) timestamped depth streams for each camera, and (4) relative extrinsics from the depth cameras to the RGB cameras. We then apply coordinate transformations to reproject the depth data into the image coordinates of the RGB videos, which allows pixel-wise depth values to be read consistently.

With aligned image coordinates, it becomes possible to read out specific depth values for target objects (with respect to the main camera). A straightforward approach is to use segmentation masks for the target object and average the depth values over the masked pixels. However, this automated approach did not work well in practice because the depth information from the camera is often incomplete and ambiguous, especially at pixels with high motion, due to hardware limitations. As a result, we could not reliably automate depth extraction for the target objects using only segmentation masks.

\begin{figure*}[ht]
    \centering
    \includegraphics[width=0.99\linewidth]{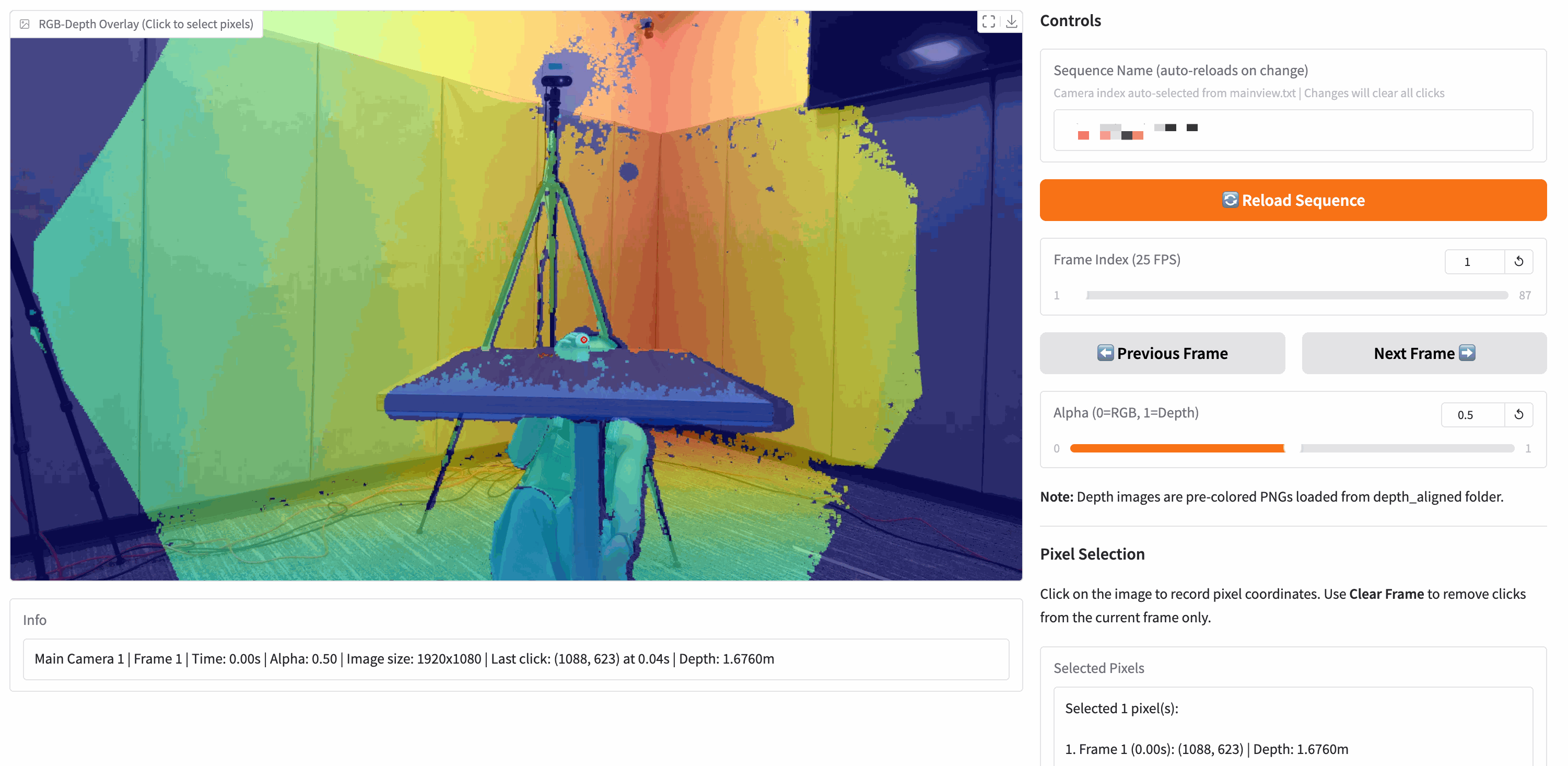}
    \caption{\textbf{The users' interface of the tool we have developed for obtaining depth value.}}
    \label{fig:depth_annotator}
\end{figure*}

To ensure high-quality depth measurements, we developed a UI-based tool that overlays the transformed depth map on top of the RGB frames. The tool allows users to click on the frame to read out exact depth values from the main camera at the selected pixels. An illustration of this UI-based tool is shown in \autoref{fig:depth_annotator}.

\noindent\textbf{List of objects used.} Table~\ref{tab:object-list} lists all physical objects used in our controlled lab videos, together with their measured dimensions.
These objects serve either as priors (with known size) or as inference targets in our kinematic tasks.
All measurements are taken with a ruler or caliper in metric units before filming.

\subsection{Internet Scraping}

\begin{figure}[t]
    \centering
    \includegraphics[width=0.99\linewidth]{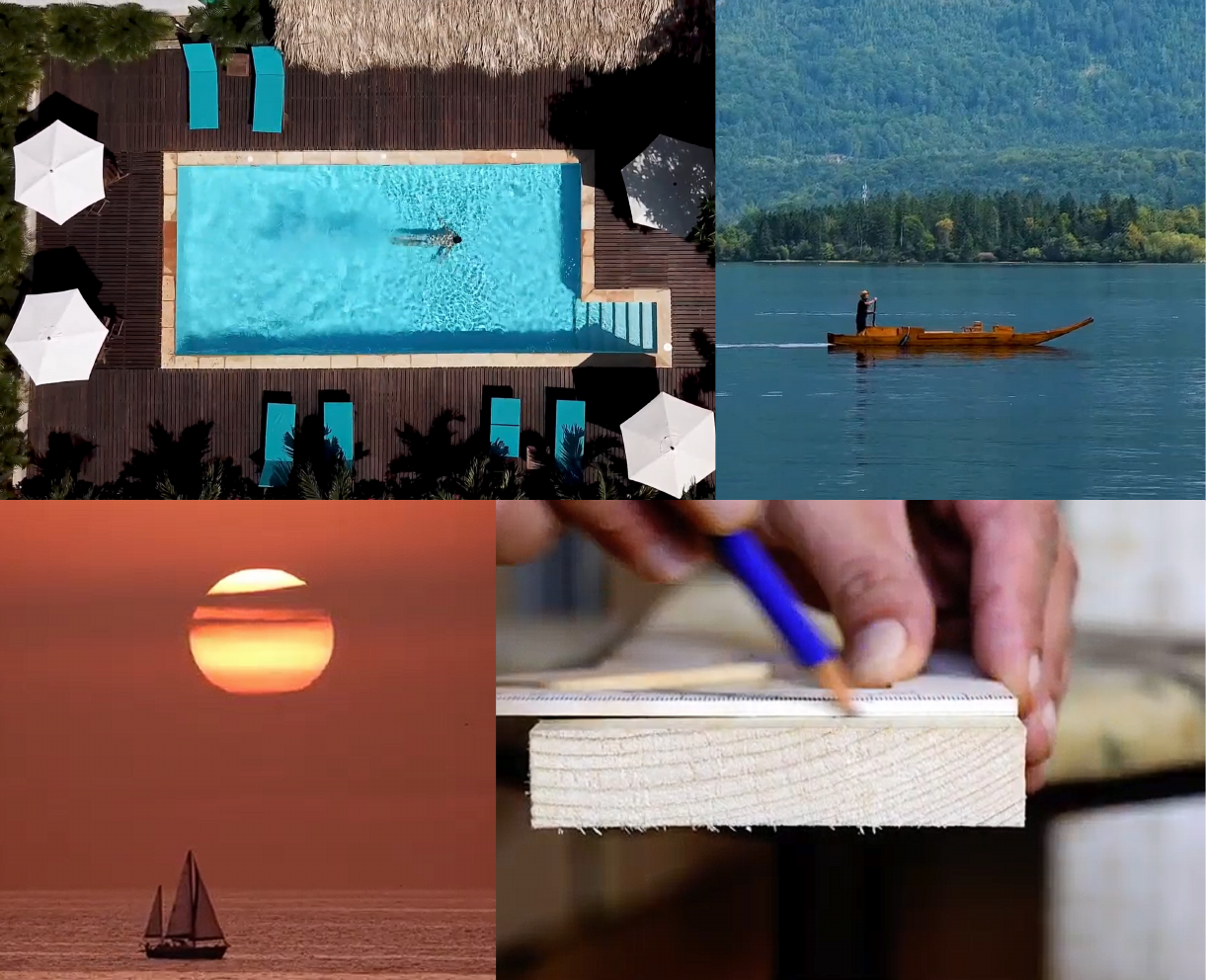}
    \caption{\textbf{Examples of videos from open-source platforms.}}
    \label{fig:internet_showcase1}
\end{figure}

\begin{figure}[t]
    \centering
    \includegraphics[width=0.99\linewidth]{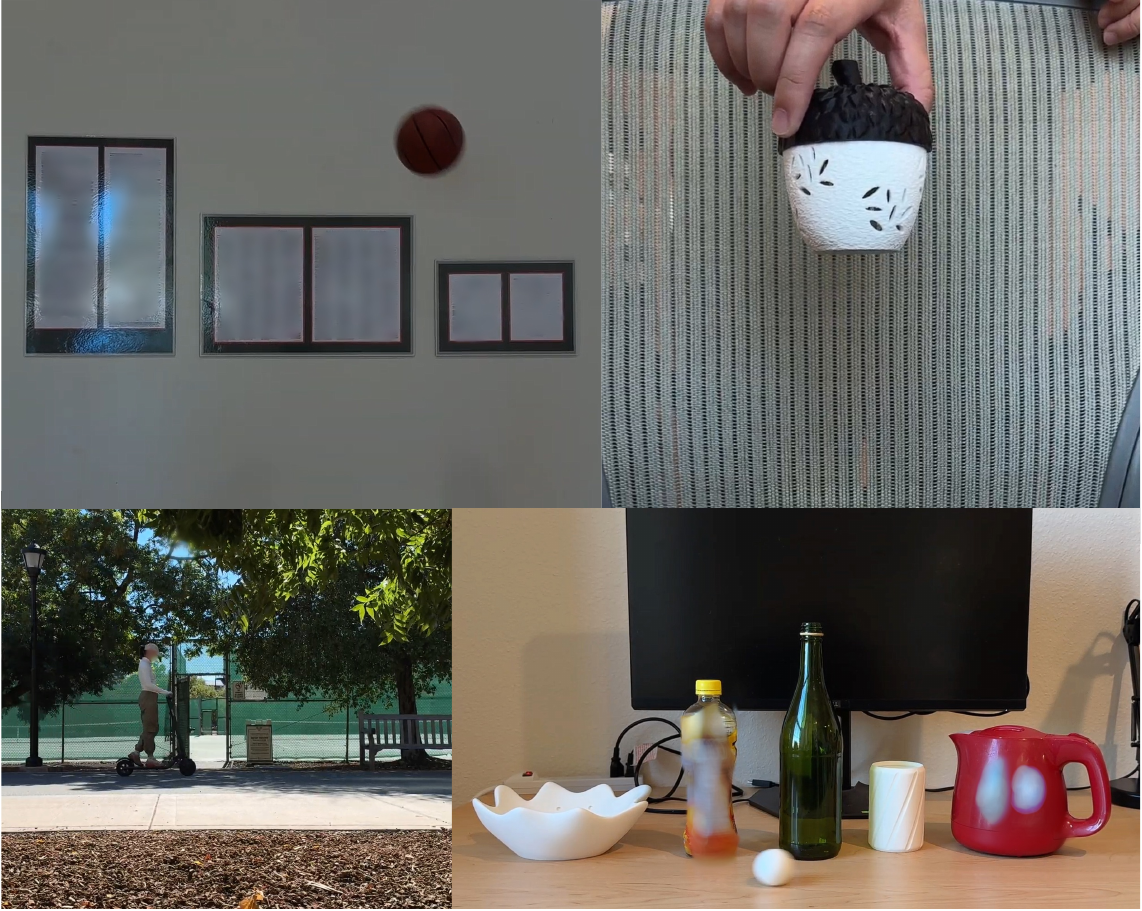}
    \caption{\textbf{Examples of self-recorded videos with identity removed.}}
    \label{fig:internet_showcase2}
\end{figure}

Our ``internet'' split in \textsc{QuantiPhy} consists of real-world videos captured by commodity cameras, and is constructed from two sources that we treat under a unified category:
(i) open-source online video platforms, contributing 42 clips;
and (ii) videos recorded by the authors using smartphone rear cameras in everyday indoor and outdoor environments, contributing 30 clips.
All of these videos are direct camera recordings of natural scenes, and thus closely reflect the statistics and imperfections of the physical world.

\noindent\textbf{Why only 2D inference from internet data.}
Unlike our simulation and controlled-lab settings, internet videos do not come with calibrated depth, camera intrinsics, or precise object geometry.
In particular, reliable metric depth is almost impossible to obtain for arbitrary internet footage.
For this reason, we restrict internet videos to 2D kinematic inference tasks.
For each selected clip, we manually construct a pixel ruler, measure pixel-level size/position trajectories of the objects of interest, and then use an approximate scale factor---derived from obvious real-world references in the scene (e.g., gravity $g = 9.8\ \text{m/s}^2$, the length of a credit card, lane width on a road, or the speed of an airport conveyor belt)---to convert these pixel measurements into world-space kinematic quantities.
While this procedure yields reasonably accurate ground truth, it is inherently less precise than the annotation pipelines used for our simulation and lab data.
Consequently, we intentionally keep the proportion of internet data moderate in the overall benchmark.

\noindent\textbf{Videos from open-source platforms.}
We choose open-source video platforms primarily because they avoid copyright issues, offer relatively high image quality, and provide diverse content.
However, videos that satisfy the three screening criteria in \autoref{sec:general_principles} (static camera, at least one rigid object undergoing translational motion, and planar motion for 2D tasks) are relatively rare.
Beyond these core constraints, we additionally require that each candidate clip contain at least one visually obvious physical prior that can be reasonably assumed or measured (e.g., gravity, a credit-card-sized object, standard lane width, or known conveyor-belt speed).
These additional constraints further narrow the pool of usable videos.
There is currently no off-the-shelf automatic pipeline for this selection process, so all internet clips are hand-picked by project members, who visually inspect candidates for compliance with our physical and annotation requirements.
Representative examples are shown in Figure~\ref{fig:internet_showcase1}.

\noindent\textbf{Author-recorded videos.}
Because suitable clips on open-source platforms are scarce, we complement them with 30 videos recorded by the authors using smartphone rear cameras in a variety of everyday scenes, including parking lots, road traffic, bedrooms, and indoor/outdoor sports venues.
During recording, we enforce a fixed camera viewpoint and ensure that at least one rigid object exhibits predominantly translational motion, so that the criteria in \autoref{sec:general_principles} are satisfied.
In many such settings, precise physical quantities (e.g., vehicle speed, basket height) cannot be directly measured.
We therefore annotate these videos using the same pixel-ruler and approximate-scale procedure as for online platform videos, again restricting them to 2D inference tasks.
Example frames from these author-recorded clips are shown in Figure~\ref{fig:internet_showcase2}.

\noindent\textbf{Privacy and anonymization.}
For all internet videos, whether sourced from open platforms or recorded by the authors, we manually inspect frames for sensitive personal information.
Whenever faces, license plates, or other identifying details appear, we apply blurring or masking before including the clip in the dataset.
This ensures that our internet data respect both copyright and privacy constraints while still providing realistic real-world scenarios for quantitative kinematic inference.

\begin{figure*}[t]
    \centering
    \includegraphics[width=0.99\linewidth]{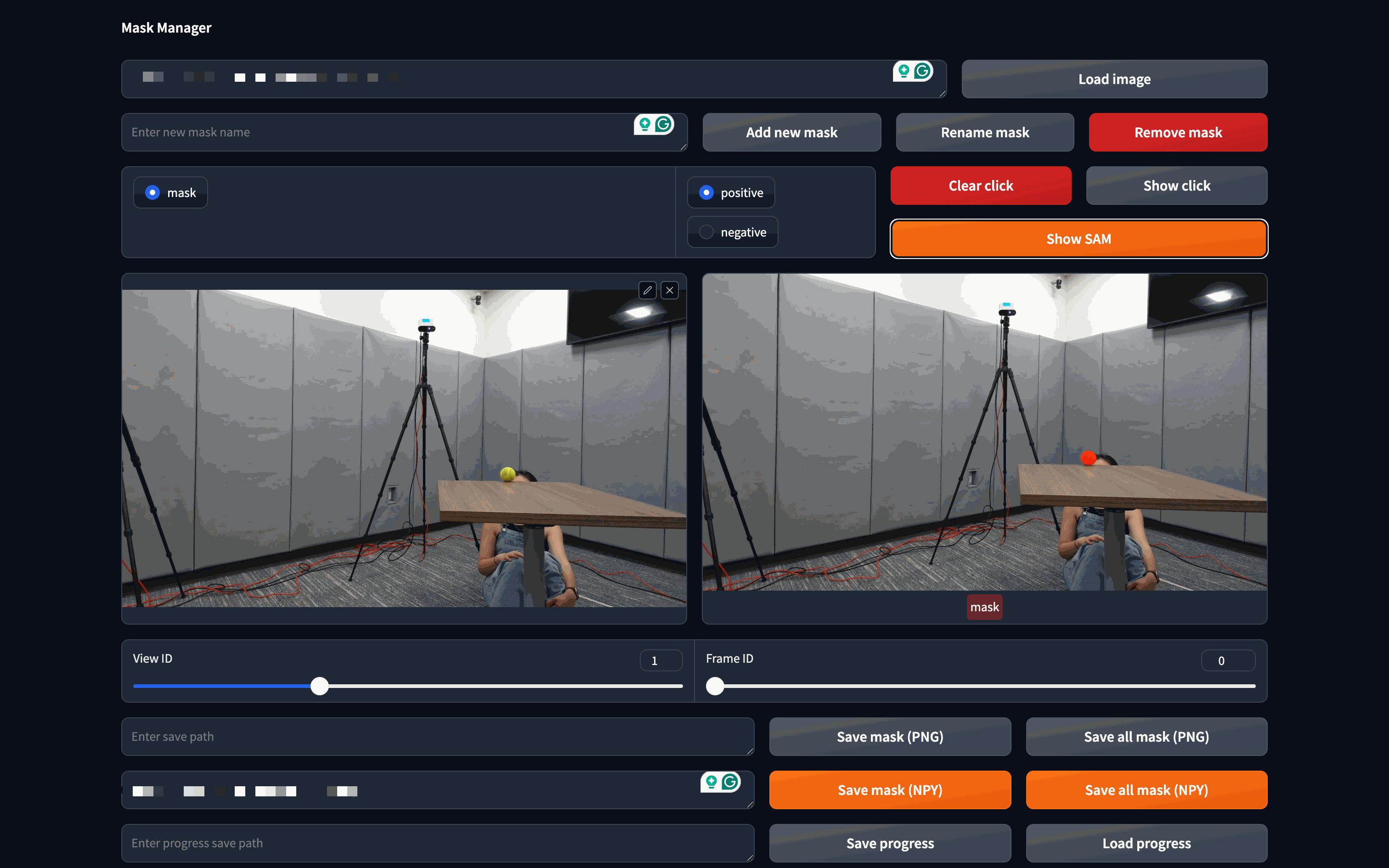}
    \caption{\textbf{The users' interface of the segmentation tool we have developed.}}
    \label{fig:segmentation_ui}
\end{figure*}

\subsection{Segmented Data}

In \textsc{QuantiPhy}, we aim to benchmark the capabilities of VLMs on videos with diverse background types. Video backgrounds often contain contextual information that may either aid or hinder the inference of a target object's physical properties. To comprehensively study this impact, we design experiments that isolate the target object by completely removing the background. Specifically, we compare the original videos against processed versions where the background has been completely denoised and removed using a segmentation model.

State-of-the-art semantic segmentation models, such as SAM 2~\cite{ravi2024sam2}, have demonstrated robust capabilities in tracking and segmenting arbitrary targets across various tasks. In our experiments, we employ SAM 2 as the backbone segmenter, utilizing a hybrid prompting strategy. To automate the pipeline, we leverage Grounding DINO 1.5~\cite{ren2024grounding} to localize target objects using textual descriptions derived from our video-question pairs. The bounding boxes generated by Grounding DINO serve as box prompts for SAM 2, enabling automatic segmentation.

\begin{figure*}[ht!]
    \centering
    \includegraphics[width=0.99\linewidth]{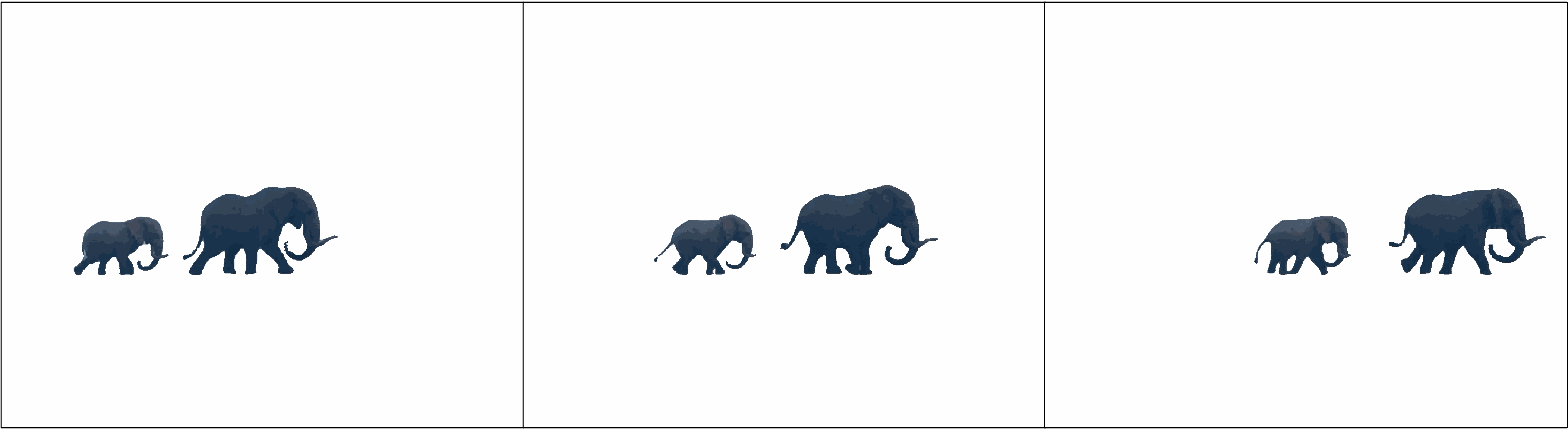}
    \caption{\textbf{Examples of segmented blender data.} After segmentation, we can replace the background image freely.}
    \label{fig:segmented_blender}
\end{figure*}

\begin{figure*}[ht!]
    \centering
    \includegraphics[width=0.99\linewidth]{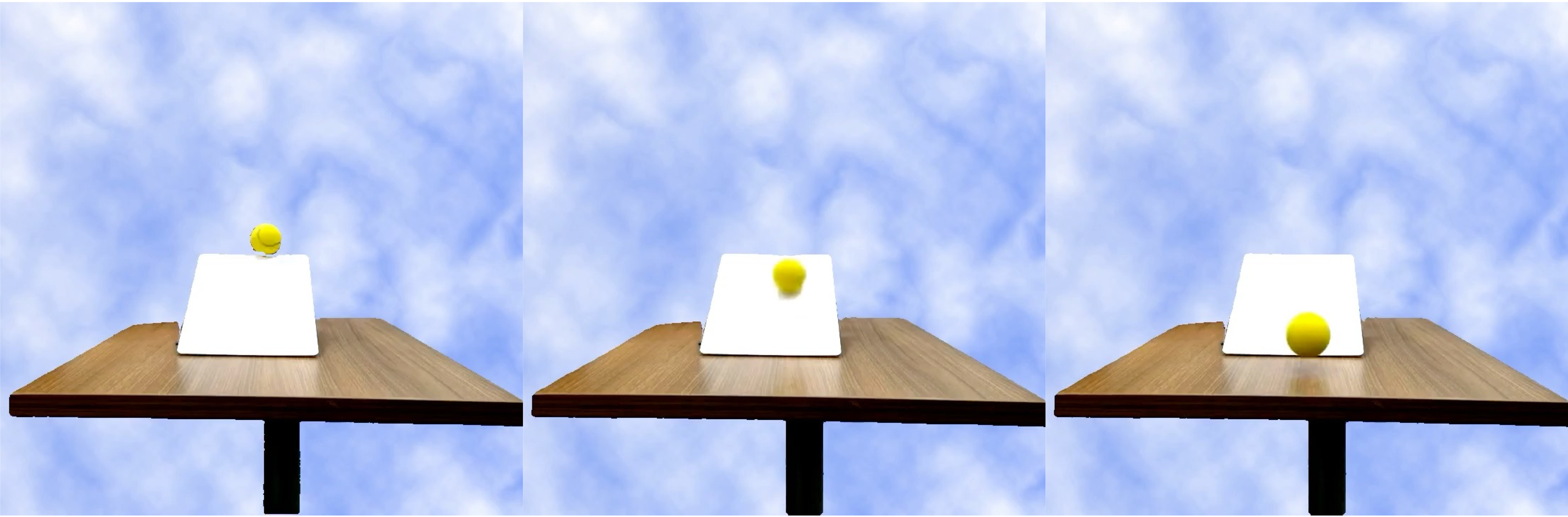}
    \caption{\textbf{Examples of segmented lab data.} After segmentation, we can replace the background image freely.}
    \label{fig:segmented_lab}
\end{figure*}

\begin{figure*}[ht!]
    \centering
    \includegraphics[width=0.99\linewidth]{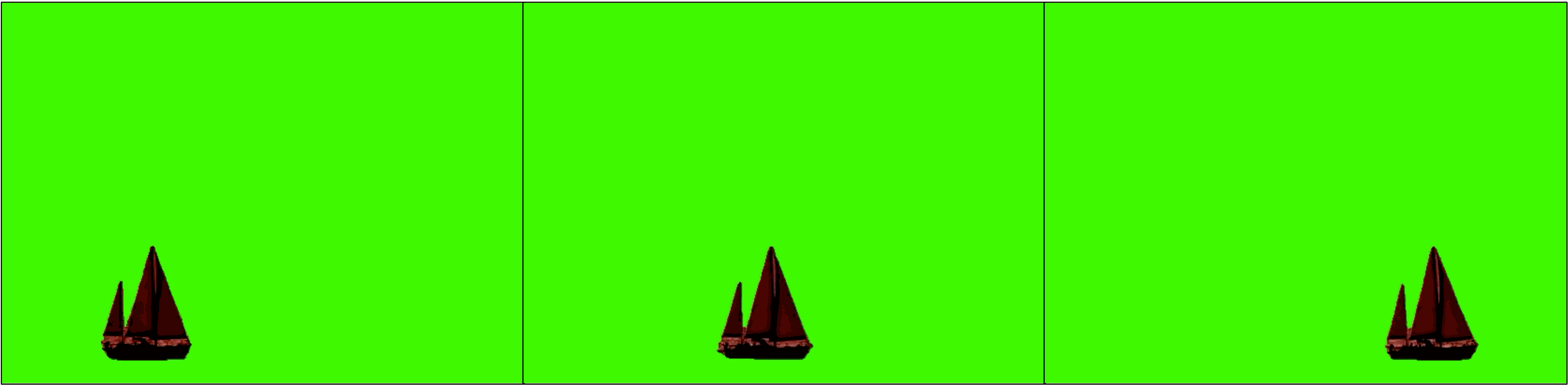}
    \caption{\textbf{Examples of segmented internet data.} After segmentation, we can replace the background image freely.}
    \label{fig:segmented_internet}
\end{figure*}

For complex scenes with multiple objects where the automated pipeline may fail, we incorporate manual intervention by providing point-based prompts to SAM 2. To streamline this workflow, we developed a custom UI-based tool that allows multiple annotators to label objects efficiently, thereby scaling the segmentation process. \autoref{fig:segmentation_ui} illustrates the user interface of this tool, and \autoref{fig:segmented_blender}, \autoref{fig:segmented_lab}, \autoref{fig:segmented_internet} provides exemplary frames of the segmented videos for reference.

\subsection{Quality Control}

The data collection process for \textsc{QuantiPhy} is highly diverse and complex. Since the data originates from heterogeneous sources with varying characteristics, establishing a universal automated protocol for quality assurance is challenging. To address this, we incorporated an additional manual review stage for all candidate data.

To maximize data quality, we manually excluded videos exhibiting excessive motion blur, severe occlusion of the target object, or objects that are difficult to model. Furthermore, we removed videos containing identifiable human subjects to ensure ethical compliance.

Following this review process, approximately 3\% of the Blender data and 30\% of the lab data were discarded, while for the videos scraped from the internet only 72 clips were retained.

\subsection{Ethical Considerations}
In compiling the \textsc{QuantiPhy} dataset, comprising both simulated Blender environments and web-sourced videos, we strictly adhere to all relevant copyright and licensing regulations. For specific data assets requiring attribution, we provide full acknowledgments in the supplementary materials.

We implemented strict privacy measures throughout the data collection and annotation phases to ensure that no personally identifiable information (PII) is retained in the dataset. Relevant Institutional Review Board (IRB) documentation is available upon request. Furthermore, \textsc{QuantiPhy} complies with ethical guidelines regarding content safety; we have rigorously screened the data to exclude biased or harmful content while prioritizing diversity to foster fairness and inclusivity.

\section{Details of Data Annotation} \label{appendix:data_annotation}

\subsection{Blender Simulation}

We annotate five basic types of physical quantities: size, displacement over a given time interval, velocity, acceleration, and depth and distance.

\subsubsection{Size} 

Blender size annotations are extracted directly from Blender's internal measurement readout to guarantee numerical accuracy. For most objects, size is obtained from the object’s axis-aligned bounding box dimensions in world space. In other special cases, the dimensions of objects may change from frame to frame.

\begin{figure*}[ht!]
    \centering
    \includegraphics[width=0.99\linewidth]{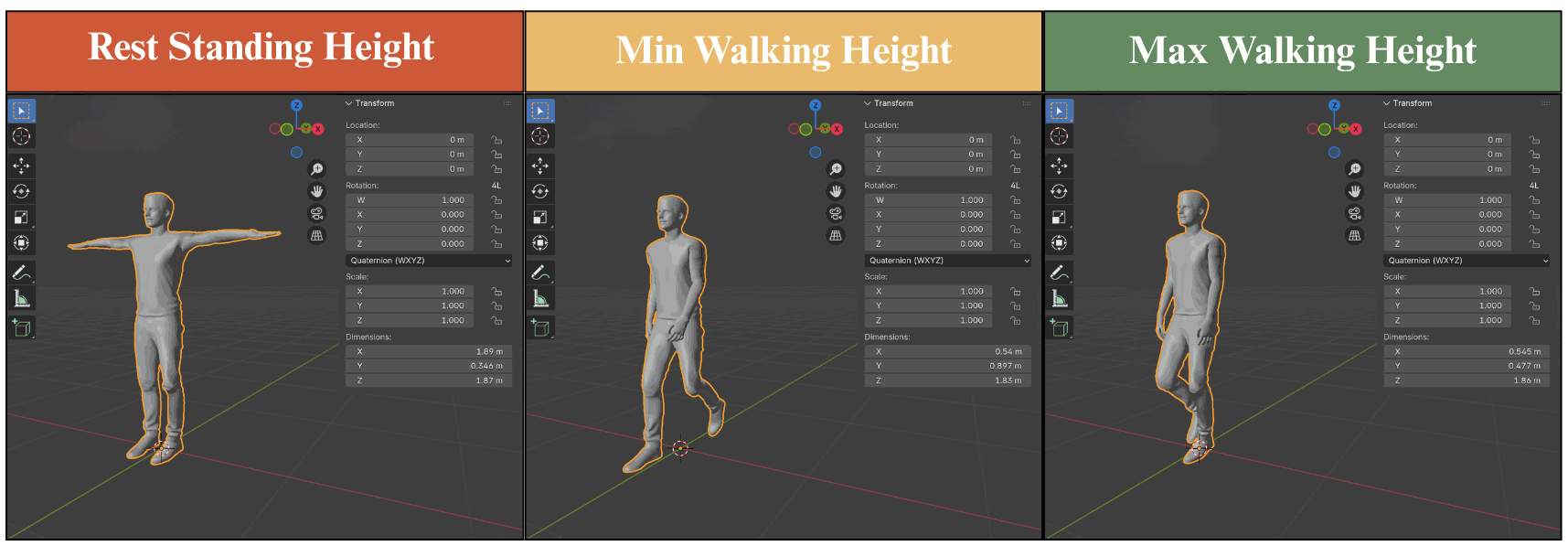}
    \caption{\textbf{Examples of human height measuring.}}
    \label{fig:human_height_measure}
\end{figure*}

For example, as shown in the \autoref{fig:human_height_measure}, articulated humans height changes while they move, thus we explicitly measure three configurations within each clip:
\begin{enumerate}
    \item \textbf{Rest standing height:} the height in a rest ``T-pose'' frame, where the character stands upright and fully extended.
    \item \textbf{Minimum apparent height during walking:} the smallest height observed over the walking segment of the clip.
    \item \textbf{Maximum apparent height during walking:} the largest height observed over the same segment.
\end{enumerate}

All heights are obtained by selecting the corresponding frames in Blender and measuring the vertical extent in world coordinates. In practice, these three measurements are usually very close (often differing by only a few centimeters), but we still record all of them to keep the annotation protocol precise and consistent.

\begin{figure*}[ht!]
    \centering
    \includegraphics[width=0.99\linewidth]{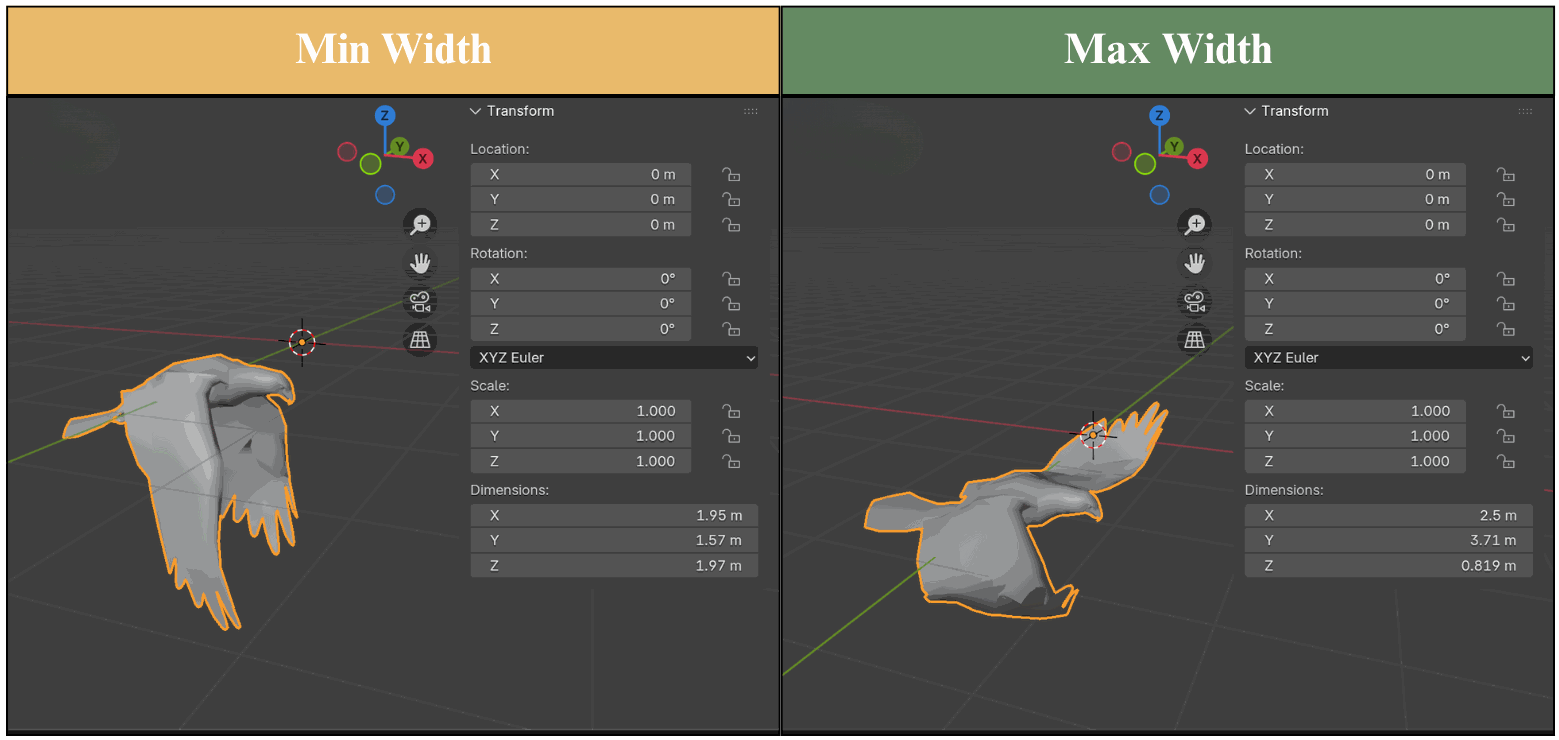}
    \caption{\textbf{Examples of flying animal measuring.}}
    \label{fig:bird_width_measure}
\end{figure*}

Similarly, for other rigged objects (such as birds and insects), we track their width frame by frame over the flight segment of the clip and only record the smallest and largest width observed. See details in \autoref{fig:bird_width_measure}. This is because, unlike the human case, these assets may not admit a clear, static “rest standing” pose in which the animal is fully extended. As a result, we do not define a separate rest measurement for flying objects.

Correspondingly, the inference questions and prior ground truth are phrased with the same level of precision. For human figures, we explicitly distinguish between the standing height at rest and the minimum and maximum height observed while walking. For flying animals, we clearly refer to the minimum or maximum width attained during flight.

\subsubsection{Displacement or Path.} 

All displacement- or path-related labels are computed from object trajectories in Blender world space using internal custom Python scripts.

\noindent\textbf{Time--frame alignment across scenes.}
In Blender, each scene specifies a render range with \texttt{scene.frame\_start} and \texttt{scene.frame\_end}, and \texttt{scene.frame\_start} is not required to be 0. We define time \(t = 0\) to occur at the render start frame, regardless of its numeric index. More precisely, if a scene has frame rate \(f\) frames per second and render start frame \(f_{\text{start}}\), then the first frame of the exported video (timestamp \(t = 0\) s) is Blender frame \(f_{\text{start}}\), the frame at time \(t = 1/f\) s is frame \(f_{\text{start}} + 1\), and so on.

When a question specifies a time interval \([t_{\text{start}}, t_{\text{end}}]\) in seconds, we convert it to Blender frame indices by
\[
i_{\text{start}} = \mathrm{round}\bigl(f_{\text{start}} + f\, t_{\text{start}}\bigr),\quad
i_{\text{end}}   = \mathrm{round}\bigl(f_{\text{start}} + f\, t_{\text{end}}\bigr).
\]

This convention works uniformly for all scenes, including those whose render ranges begin at non-zero frame numbers (e.g., \texttt{frame\_start = 20, 40, 60, \dots}). In code, the conversion has the following form:

\begin{minted}[
fontsize=\footnotesize,
linenos,
frame=lines,
framesep=2mm,
baselinestretch=1.0,
breaklines,
bgcolor=black!2
]{python}
scene   = bpy.context.scene
fps     = scene.render.fps
f_start = scene.frame_start                      # may be 0, 20, 40, ...

def time_to_frame(t_sec: float) -> int:
    return round(f_start + t_sec * fps)
\end{minted}

When an interval is specified directly in frames, we simply take \((i_{\text{start}}, i_{\text{end}})\) as given and interpret the corresponding times as
\[
t_{\text{start}} = \frac{i_{\text{start}} - f_{\text{start}}}{f},\quad
t_{\text{end}}   = \frac{i_{\text{end}}   - f_{\text{start}}}{f},
\]
so that \(t = 0\) always aligns with the first frame of the rendered video, irrespective of the absolute Blender frame index.

At any frame \(i\), the scripts query the world-space position of the target object (e.g., the animated rigid body) via

\begin{minted}[
fontsize=\footnotesize,
linenos,
frame=lines,
framesep=2mm,
baselinestretch=1.0,
breaklines,
bgcolor=black!2
]{python}
scene.frame_set(i)
loc = obj.matrix_world.translation.copy()          # (x, y, z) in world space
\end{minted}

yielding a vector $p(i) = (x(i), y(i), z(i)) \in \mathbb{R}^3$.

\noindent\textbf{Displacement annotations.}
For displacement labels, we consider a time interval \([t_{\text{start}}, t_{\text{end}}]\)  or frame interval \([i_{\text{start}}, i_{\text{end}}]\). The scripts record the world-space locations
\[
p_{\text{start}} = p(t_{\text{start}}) = p(i_{\text{start}}),\quad
p_{\text{end}}   = p(t_{\text{end}})   = p(i_{\text{end}}),
\]
and form the displacement vector
\[
\Delta p = p_{\text{end}} - p_{\text{start}} = (\Delta x, \Delta y, \Delta z).
\]

From this vector we derive two scalar quantities.

\begin{enumerate}
    
\item The full 3D displacement
\[
D_{\text{3D}} = \lVert \Delta p \rVert_2 = \sqrt{\Delta x^2 + \Delta y^2 + \Delta z^2},
\]
\item The planar displacement in a 2D plane, obtained by projecting the motion onto that plane. 
For any chosen 2D coordinate plane with axes \((u, v)\), we define the planar displacement as
\[
D_{2\text{D}} = \sqrt{(\Delta u)^2 + (\Delta v)^2}.
\]
\end{enumerate}

In our benchmark we primarily use the three canonical planes:
the horizontal ground plane
\[
D_{\text{XY}} = \sqrt{(\Delta x)^2 + (\Delta y)^2},
\]
and the two vertical planes
\[
D_{\text{XZ}} = \sqrt{(\Delta x)^2 + (\Delta z)^2}, \qquad
D_{\text{YZ}} = \sqrt{(\Delta y)^2 + (\Delta z)^2}.
\]

A compact version of the core computation is:

\begin{minted}[
fontsize=\footnotesize,
linenos,
frame=lines,
framesep=2mm,
baselinestretch=1.0,
breaklines,
bgcolor=black!2
]{python}
def displacement_world(obj, frame_start, frame_end, scene):
    # sample world-space positions at start and end
    loc_start = get_world_location_at_frame(obj, frame_start, scene)
    loc_end   = get_world_location_at_frame(obj, frame_end, scene)

    # displacement vector
    disp_vec = loc_end - loc_start
    dx, dy, dz = disp_vec.x, disp_vec.y, disp_vec.z

    # 3D displacement and horizontal (XY) displacement
    D_3D = disp_vec.length                 # sqrt(dx*dx + dy*dy + dz*dz)
    D_XY = (dx**2 + dy**2) ** 0.5          # sqrt(dx*dx + dy*dy)

    return {
        "frame_start": frame_start,
        "frame_end": frame_end,
        "loc_start":  loc_start,
        "loc_end":    loc_end,
        "dx": dx, "dy": dy, "dz": dz,
        "D_3D": D_3D,
        "D_XY": D_XY,
    }
\end{minted}

For each annotated interval, we store:
\begin{itemize}
  \item the frame indices \((i_{\text{start}}, i_{\text{end}})\);
  \item the corresponding times \((t_{\text{start}}, t_{\text{end}})\), computed relative to the render start frame as above;
  \item the start and end world-space coordinates \(p_{\text{start}}, p_{\text{end}}\);
  \item the displacement components \((\Delta x, \Delta y, \Delta z)\);
  \item the two scalar values \(D_{\text{3D}}\) and \(D_{\text{XY}}\) in Blender units.
\end{itemize}

In the question text, we explicitly state whether “displacement” refers to the full 3D displacement \(D_{\text{3D}}\) or to a planar displacement \(D_{2\text{D}}\) in a specified plane (e.g., horizontal \(D_{\text{XY}}\) or vertical \(D_{\text{XZ}}\) / \(D_{\text{YZ}}\)). 
The posterior ground-truth answer for that question is taken from the corresponding stored value.

\subsubsection{Velocity and Acceleration.}

\noindent\textbf{Velocity measurement and uniform-speed diagnostics.}
Using the time--frame convention introduced above, let \(p_i \in \mathbb{R}^3\) denote the world-space origin of the target object at frame \(i\), obtained by querying \texttt{obj.matrix\_world.translation} after calling \texttt{scene.frame\_set(i)}. For two consecutive frames \(i-1\) and \(i\) with timestamps \(t_{i-1}\) and \(t_i\), we define the temporal spacing
\[
\Delta t_i = t_i - t_{i-1},
\]
and the instantaneous 3D velocity and scalar speed at frame \(i\) as
\[
\mathbf{v}_i = \frac{p_i - p_{i-1}}{\Delta t_i},
\qquad
s_i = \|\mathbf{v}_i\|_2,
\]
where \(\mathbf{v}_i = (v_{x,i}, v_{y,i}, v_{z,i})\) is the full 3D velocity vector in world coordinates and
\[
s_i = \lVert \mathbf{v}_i \rVert_2
= \sqrt{v_{x,i}^2 + v_{y,i}^2 + v_{z,i}^2}
\]
is its Euclidean norm. All operations are performed in Blender world space and use the true frame rate, so that positions \(p_i\), velocities \(\mathbf{v}_i\), and speeds \(s_i\) are expressed in a self-consistent system of physical units determined by the underlying scene scale.

The Blender analysis script computes these quantities frame by frame and writes them into a text block in Blender's Text Editor:
\begin{minted}[
fontsize=\footnotesize,
linenos,
frame=lines,
framesep=2mm,
baselinestretch=1.0,
breaklines,
bgcolor=black!2
]{python}
prev_loc   = None
prev_frame = None

for frame in range(frame_start, frame_end + 1):
    scene.frame_set(frame)
    loc = obj.matrix_world.translation.copy()   # p_i

    if prev_loc is None:
        vel_vec = Vector((0.0, 0.0, 0.0))
        speed   = 0.0
    else:
        dt      = (frame - prev_frame) / fps        # delta t_i
        disp    = loc - prev_loc                    # p_i - p_{i-1}
        vel_vec = disp / dt                         # v_i
        speed   = vel_vec.length                    # s_i

    write_per_frame_entry(frame, loc, vel_vec, speed)
    prev_loc   = loc
    prev_frame = frame
\end{minted}

For each frame \(i\) in the chosen range, this loop automatically prints one line of kinematic data into the text block, containing:
\begin{itemize}
  \item the frame index \(i\) and its timestamp \(t_i\) (relative to the render start frame);
  \item the world-space position \(p_i\);
  \item the full 3D velocity vector \(\mathbf{v}_i\);
  \item the scalar speed \(s_i\).
\end{itemize}
Any question that refers to “the speed at \(t\) seconds” or “the velocity at frame \(i\)” is answered by mapping the queried time to a frame index using the same time--frame conversion and then reading off the corresponding per-frame entry from this table, without any manual measurement or additional approximation.

In addition to per-frame velocities, we use the sequence of speeds \(\{s_i\}\) to characterize simple motion regimes at the clip level. Let \(I_v\) be the index set of frames for which a speed value is defined (all frames except the very first one), and let \(N_v = |I_v|\). We compute the mean speed
\[
\bar{s} = \frac{1}{N_v} \sum_{i \in I_v} s_i
\]
and the maximum relative deviation from this mean,
\[
\delta_{\text{speed}} =
\begin{cases}
0, & \text{if } \bar{s} = 0, \\[4pt]
\displaystyle \max_{i \in I_v} \frac{|s_i - \bar{s}|}{\bar{s}}, & \text{otherwise.}
\end{cases}
\]
Given a user-specified tolerance \(\tau_v\) (parameter \texttt{UNIFORM\_SPEED\_TOLERANCE}, e.g., \(\tau_v = 0.01\)), we classify the clip as:
\begin{itemize}
  \item \textbf{no effective motion} if \(\bar{s} \approx 0\) (numerically \(\bar{s} = 0\), indicating that the object is essentially stationary);
  \item \textbf{approximately uniform speed} if \(\bar{s} > 0\) and \(\delta_{\text{speed}} \le \tau_v\);
  \item \textbf{non-uniform speed} otherwise (speed fluctuations exceed the tolerance).
\end{itemize}
This logic matches the following code fragment:
\begin{minted}[
fontsize=\footnotesize,
linenos,
frame=lines,
framesep=2mm,
baselinestretch=1.0,
breaklines,
bgcolor=black!2
]{python}
avg_speed   = sum(speeds) / len(speeds)
max_dev     = max(abs(s - avg_speed) for s in speeds)
max_rel_dev = max_dev / avg_speed if avg_speed != 0 else 0.0

if avg_speed == 0:
    # almost no motion
    ...
else:
    if max_rel_dev <= UNIFORM_SPEED_TOLERANCE:
        # approximately uniform speed
        ...
    else:
        # not uniform-speed
        ...
\end{minted}

\noindent\textbf{Acceleration measurement and acceleration diagnostics.}
Starting from the third frame, we derive a scalar acceleration sequence that describes how quickly the object’s speed changes over time. Given the per-frame speeds \(\{s_i\}\) defined above, and the same temporal spacings \(\Delta t_i = t_i - t_{i-1}\), we define for all frames \(i \geq 2\)
\[
a_i = \frac{s_i - s_{i-1}}{\Delta t_i},
\]
so that \(a_i\) measures the finite-difference rate of change of speed between frames \(i-1\) and \(i\). The script computes these values alongside the speeds:
\begin{minted}[
fontsize=\footnotesize,
linenos,
frame=lines,
framesep=2mm,
baselinestretch=1.0,
breaklines,
bgcolor=black!2
]{python}
prev_speed = None
prev_frame = None

for frame in range(frame_start, frame_end + 1):
    scene.frame_set(frame)
    loc = obj.matrix_world.translation.copy()

    if prev_frame is None:
        dt    = 0.0
        speed = 0.0
        accel = None
    else:
        dt      = (frame - prev_frame) / fps
        disp    = loc - prev_loc
        vel_vec = disp / dt
        speed   = vel_vec.length

        if prev_speed is None:
            accel = None
        else:
            accel = (speed - prev_speed) / dt   # a_i

    write_per_frame_entry(frame, speed, accel)
    prev_loc   = loc
    prev_speed = speed
    prev_frame = frame
\end{minted}

For each frame \(i\) in the valid range, this produces a scalar acceleration value \(a_i\) (undefined at the first two frames), which we store in the same text table. Any question that refers to “the acceleration at frame \(i\)” or to “the acceleration over a given interval” is answered by mapping the queried time to a frame index and reading off the corresponding \(a_i\) entry.

To summarize acceleration behaviour over the entire clip, we perform an analogous diagnostic analysis on the set of defined accelerations. Let \(I_a\) be the index set of frames for which \(a_i\) is defined (starting from the third frame), and let \(N_a = |I_a|\). We compute the mean acceleration
\[
\bar{a} = \frac{1}{N_a} \sum_{i \in I_a} a_i
\]
and the maximum absolute deviation from this mean,
\[
\Delta_a^{\max} = \max_{i \in I_a} |a_i - \bar{a}|.
\]
If \(|\bar{a}|\) is larger than a small absolute threshold \(\varepsilon_{\min}\) (parameter \texttt{MIN\_ABS\_ACCEL}), we also measure the maximum relative deviation
\[
\delta_{\text{accel}} =
\begin{cases}
0, & \text{if } |\bar{a}| \le \varepsilon_{\min}, \\[4pt]
\displaystyle \frac{\Delta_a^{\max}}{|\bar{a}|}, & \text{otherwise.}
\end{cases}
\]
Given a user-specified tolerance \(\tau_a\) (parameter \texttt{UNIFORM\_ACCEL\_TOLERANCE}), the script then classifies the acceleration regime as:
\begin{itemize}
  \item \textbf{near-zero acceleration} if \(|\bar{a}| \le \varepsilon_{\min}\); in this case the overall acceleration is negligible and the clip is better interpreted as approximately constant-speed (or almost static);
  \item \textbf{approximately uniformly accelerated} if \(|\bar{a}| > \varepsilon_{\min}\) and \(\delta_{\text{accel}} \le \tau_a\); the sign of \(\bar{a}\) further distinguishes \emph{uniform acceleration} (\(\bar{a} > 0\)) from \emph{uniform deceleration} (\(\bar{a} < 0\));
  \item \textbf{irregular acceleration} otherwise, meaning that the accelerations fluctuate significantly around their mean.
\end{itemize}
This classification logic is implemented as:
\begin{minted}[
fontsize=\footnotesize,
linenos,
frame=lines,
framesep=2mm,
baselinestretch=1.0,
breaklines,
bgcolor=black!2
]{python}
avg_accel   = sum(accels) / len(accels)
max_dev_a   = max(abs(a - avg_accel) for a in accels)
if abs(avg_accel) > MIN_ABS_ACCEL:
    max_rel_dev_a = max_dev_a / abs(avg_accel)
else:
    max_rel_dev_a = 0.0

if abs(avg_accel) <= MIN_ABS_ACCEL:
    # acceleration is very small -> closer to uniform speed / almost static
    ...
else:
    if max_rel_dev_a <= UNIFORM_ACCEL_TOLERANCE:
        if avg_accel > 0:
            # approximately uniformly accelerating
            ...
        elif avg_accel < 0:
            # approximately uniformly decelerating
            ...
        else:
            # numerically close to zero
            ...
    else:
        # not uniformly accelerated
        ...
\end{minted}

These acceleration diagnostics never modify the underlying frame-level values \(\{a_i\}\), just as the velocity diagnostics never modify \(\{s_i\}\) or \(\{\mathbf{v}_i\}\). Instead, they provide a principled, threshold-based way to tag each clip as approximately constant-speed, uniformly accelerated, uniformly decelerated, or irregular, allowing us to phrase velocity- and acceleration-related questions at a level of precision that matches the actual motion regime present in each clip.

\subsubsection{Depth and Distance}

All geometric quantities used in our benchmark are computed directly inside Blender in world coordinates via a Python script that evaluates the scene frame by frame. The script reads the world-space transforms of selected entities, computes Euclidean distances, and logs a per-frame table into a Blender Text Editor text block. From this table, we derive two kinds of quantities: (i) depth metadata between objects and the camera, and (ii) inter-object distances in 3D and in projected 2D planes. 

\noindent\textbf{Depth metadata (object--camera).}
We extract depth values in Blender only as auxiliary geometric metadata,
not appearing in the inference question text or supervised targets. For any
object \(o\) with world-space position
\[
\mathbf{p}_t(o) = (x_t^o, y_t^o, z_t^o) \in \mathbb{R}^3
\]
and the active camera object with world-space origin
\[
\mathbf{p}_t^{\text{cam}} = (x_t^{\text{cam}}, y_t^{\text{cam}}, z_t^{\text{cam}}) \in \mathbb{R}^3,
\]
at time \(t\), we define the depth as the 3D Euclidean distance

\[
\begin{aligned}
d^{\text{depth}}_t(o)
&= \bigl\lVert \mathbf{p}_t(o) - \mathbf{p}_t^{\text{cam}} \bigr\rVert_2 \\
&= \sqrt{(x_t^o - x_t^{\text{cam}})^2 + (y_t^o - y_t^{\text{cam}})^2 + (z_t^o - z_t^{\text{cam}})^2}.
\end{aligned}
\]

During annotation, we treat \(d^{\text{depth}}_t(o)\) as a time series and often materialize it at a small set of shared reference time points (e.g., \(t_1 = 1\text{s}\), \(t_2 = 2\text{s}\)). Thus, for a moving object we record pairs such as
\[
\bigl(d^{\text{depth}}_{t_1}(o),\, d^{\text{depth}}_{t_2}(o)\bigr),
\]
and we use the same time points for other relevant entities in the scene (e.g., additional moving objects or candidate target objects). This temporal alignment ensures that when we expose depth values as priors, they are always comparable across objects at exactly the same timestamps. For static objects, the depth is constant over the clip, so the values at different time points are trivially identical.

Crucially, we never use these depth values as evaluation targets. We do not ask questions of the form “What is the distance between object \(o\) and the camera at time \(t\)?”, and we do not treat \(d^{\text{depth}}_t(o)\) as ground truth in any task. Instead, depth is used only as internal geometric metadata and, in some infer questions, as numeric priors that can help a VLM reason about the relative 3D configuration of the scene. Even in those infer questions, we do not necessarily expose the camera--target depth of the queried object itself; rather, we expose a subset of aligned depth values for selected objects. All exposed depth values are manually reviewed to ensure that, together with the visual information, they are logically sufficient to infer the correct answer.

\noindent\textbf{3D inter-object distances.}
We use the same Blender script to annotate 3D distances between pairs of entities. In our benchmark, these entities are either
(i) two moving objects, or
(ii) a moving object and a static reference object in the scene.
We do not treat object--camera distances as inter-object ground truth; camera-related distances only appear as depth metadata as described above.

Let $\mathbf{p}_t(a) = (x_t^a, y_t^a, z_t^a)
\quad\text{and}\quad
\mathbf{p}_t(b) = (x_t^b, y_t^b, z_t^b)$ denote the world-space positions of entities \(a\) and \(b\) at time \(t\). The 3D object--object distance is
\[
\begin{aligned}
d^{3\text{D}}_t(a,b)
&= \bigl\lVert \mathbf{p}_t(a) - \mathbf{p}_t(b) \bigr\rVert_2 \\
&= \sqrt{(x_t^a - x_t^b)^2 + (y_t^a - y_t^b)^2 + (z_t^a - z_t^b)^2}.
\end{aligned}
\]

For objects whose dimensions do not change over time, we take \(\mathbf{p}_t(o)\)to be the Blender object origin in world coordinates. For articulated skeletal models (e.g., humans), transformations are defined on a set of bones rather than a single rigid body. In these cases, we first enumerate all bones in the armature and select a semantically stable reference joint (for humans, typically the pelvis/hip bone) as the anchor. All distances involving that character are then defined using the world-space position of this reference bone, which provides a temporally stable and semantically meaningful notion of “where the character is”. For some non-humanoid articulated assets or composite rigs where no single point or bone has a clear semantic interpretation as the “character center” (e.g., certain vehicles or multi-part machines), we instead introduce an auxiliary helper object. Concretely, we create an Empty object \(h\) rigidly parented to the rig root, snap its origin to the root in world space, and log its world-space position \(\mathbf{p}_t(h)\) at each frame. In those scenes, all distances involving the articulated asset are computed with respect to \(\mathbf{p}_t(h)\) rather than a specific bone. This dual strategy (bone-based anchors for humanoid characters and helper-object anchors for other articulated assets) keeps the distance annotations consistent while accommodating the diversity of rig structures in our Blender scenes.

\noindent\textbf{Planar (2D) distances.}
Some tasks explicitly constrain distance reasoning to a 2D projection, for example ``horizontal distance in the ground plane'' or ``distance in the vertical cross-section''. To support such tasks, we also annotate \emph{planar distances} by projecting 3D positions onto a chosen coordinate plane \(\Pi \in \{\mathrm{XY}, \mathrm{XZ}, \mathrm{YZ}\}\). We define the projections
\[
\begin{aligned}
\Pi_{\mathrm{XY}}(x,y,z) &= (x,y),\\
\Pi_{\mathrm{XZ}}(x,y,z) &= (x,z),\\
\Pi_{\mathrm{YZ}}(x,y,z) &= (y,z).
\end{aligned}
\]
The planar distance between \(a\) and \(b\) at time \(t\) is then
\[
d^{\text{plane}}_t(a,b)
= \bigl\lVert \Pi\bigl(\mathbf{p}_t(a)\bigr) - \Pi\bigl(\mathbf{p}_t(b)\bigr) \bigr\rVert_2.
\]

Whether a distance question is treated as 2D or 3D in our benchmark is fully determined by (i) the infer question and (ii) the geometric category of the underlying video (2D vs.\ 3D sequence). Each video is assigned to a 2D or 3D split based on how it is generated and how its geometry is intended to be interpreted.

For videos in the 2D split, distances are interpreted in a single plane by construction. If a question in such a video simply asks for ``the distance between objects \(a\) and \(b\)'', the ground-truth target is the planar distance \(d^{\text{plane}}_t(a,b)\), and the task is labeled as 2D.

For videos in the 3D split, we have full world-space geometry. If the infer question explicitly restricts reasoning to a plane (e.g., ``horizontal distance'' or ``vertical distance''), we again use the planar distance \(d^{\text{plane}}_t(a,b)\) and label the task as 2D. Otherwise, distance questions on 3D videos default to the full 3D Euclidean distance \(d_t(a,b)\) defined above, and these tasks are labeled as 3D.

\noindent\textbf{Blender implementation and per-frame logging.}
All depth and distance quantities above are computed by a unified Blender Python script. For a specified frame range \([f_{\min}, f_{\max}]\), the script iterates over frames, queries world-space positions, computes distances, and writes a formatted table into a Blender Text Editor text block. The core of the script is:
\begin{minted}[
fontsize=\footnotesize,
linenos,
frame=lines,
framesep=2mm,
baselinestretch=1.0,
breaklines,
breakanywhere,
bgcolor=black!2
]{python}
import bpy
from mathutils import Vector

scene = bpy.context.scene
fps   = scene.render.fps

def world_pos(obj_or_bone):
    """Return world-space translation of an object or a specific bone."""
    # For regular objects:
    if hasattr(obj_or_bone, "matrix_world"):
        return obj_or_bone.matrix_world.to_translation()
    # For pose bones (e.g., armature.pose.bones["Hips"]):
    return obj_or_bone.matrix.to_translation()

for f in range(f_min, f_max + 1):
    scene.frame_set(f)

    # Example: distance between entities A and B
    p_a = world_pos(entity_a)
    p_b = world_pos(entity_b)

    # Full 3D distance in world coordinates
    dist_3d = (p_a - p_b).length

    # Example planar distance in the XY plane (horizontal separation):
    # other planes (XZ, YZ) are obtained analogously in the full script.
    p_a_xy = Vector((p_a.x, p_a.y, 0.0))
    p_b_xy = Vector((p_b.x, p_b.y, 0.0))
    dist_xy = (p_a_xy - p_b_xy).length

    # Timestamp (seconds), assuming t = 0 at f_min
    t = (f - f_min) / fps

    # Log one line (frame, time, 3D distance, XY-plane distance, ...)
    write_to_text_block(f, t, dist_3d, dist_xy)
\end{minted}

The same script is run with different choices of \texttt{entity\_a} and \texttt{entity\_b} to (i) record object--camera distances as depth metadata and (ii) generate object--object distance annotations. In the code listing above, we show the XY-plane case for concreteness. In practice, this block is used as a template: when planar distances in other coordinate planes (XZ or YZ) are
needed, we simply modify the projection lines accordingly (e.g.,\texttt{Vector((p\_a.x, p\_a.z, 0.0))} for XZ or
\texttt{Vector((0.0, p\_a.y, p\_a.z))} for YZ) and rerun the script for that scene.

All measurements are derived directly from Blender's world-space transforms, so the numeric values used in our questions and annotations match the underlying scene geometry exactly. Depth values are stored only as auxiliary geometric
metadata or optional numeric priors, whereas inter-object distances serve as the ground truth posterior.

\subsection{Lab Data Annotation}

Unlike Blender-based simulations, lab captures do not provide annotations that can be directly read out from software. Nevertheless, with our multi-stereo camera setup, we are able to reconstruct both the 3D scene and the 3D geometry of the target objects.

To obtain annotations for lab captures, we first attempted to use off-the-shelf 3D reconstruction models such as FoundationPose~\cite{wen2024foundationpose} to assist with annotation. However, in our experiments, we found that even state-of-the-art AI models struggled to localize objects stably in world coordinates and frequently failed under occlusions. 

Therefore, we instead rely on traditional geometry-based reconstruction with calibrated camera poses. We use the main camera together with manually selected metric depth values to recover the objects' world coordinates. Concretely, we use the UI-based annotation tool introduced earlier (see \autoref{fig:depth_annotator}) to manually click on the center of the target object across a sequence of frames in the video. The tool records the 2D coordinates of each click in image space together with the corresponding metric depth from the main camera. Using the calibrated camera intrinsics, we then back-project the annotated object centers into the world coordinate system.

To obtain static priors such as object size, we manually measure the shape and dimensions of each object to reduce measurement error. To obtain dynamic priors such as object velocities and accelerations, we annotate the object center across at least 5 adjacent frames (for smoothing purpose) and compute the instantaneous velocity and acceleration magnitudes from the resulting sequence of world coordinates, assuming a constant frame rate. Formally, the computation of dynamic priors in lab captured videos are:
\begin{align*}
v_k^{\mathrm{world}} 
  &\approx \frac{x_{k+1}^{\mathrm{world}} - x_{k}^{\mathrm{world}}}{\Delta t}, \\
a_k^{\mathrm{world}} 
  &\approx \frac{x_{k+1}^{\mathrm{world}} - 2 x_{k}^{\mathrm{world}} + x_{k-1}^{\mathrm{world}}}{\Delta t^2}, 
\end{align*}
where $\Delta t$ is determined from the frame rate of the captured videos.

Note that, based on our predefined questions and the priors required at specific time stamps, not all frames of the lab videos need to be annotated. This greatly reduces the annotation workload and helps prevent error accumulation over time.

When the object is occluded in the main camera view, we instead estimate its world coordinates by averaging the positions obtained by transforming its 2D annotations from the other three cameras into the world coordinate system.

\subsection{Internet Data Annotation}
\label{sec:internet_annotation}

For internet videos, we must derive kinematic ground truth from raw pixels rather than from simulator logs or calibrated sensors.
To keep the process systematic, we adopt a three-stage annotation workflow: (i) metadata and task specification, (ii) pixel-level measurement with a custom video tool, and (iii) conversion from pixel-space kinematics to real-world quantities.

\noindent\textbf{Metadata and task specification.}
Each collected internet clip is first assigned a unique identifier (e.g., \texttt{internet\_0001}).
Based on the scene content, annotators determine the physical prior available in the video (e.g., gravity $g=9.8\,\mathrm{m/s^2}$, the length of a credit card, lane width on a road, or conveyor-belt speed), and label the corresponding \textbf{\texttt{video\_type}} code (e.g., \texttt{S2MC}, \texttt{A2SS}) as described in \autoref{subsec:video_type}.
We then identify all feasible inference targets in the clip (e.g., vehicle speed, ball acceleration, object size) and enumerate kinematic inference questions that can be solved from the chosen prior.
For each question, we annotate the \textbf{\texttt{inference\_type}} (static vs.\ dynamic posterior) and write the final textual question so that the requested quantity, unit, and time reference (instantaneous vs.\ average) are explicit.

\begin{figure*}[ht!]
    \centering
    \includegraphics[width=0.99\linewidth]{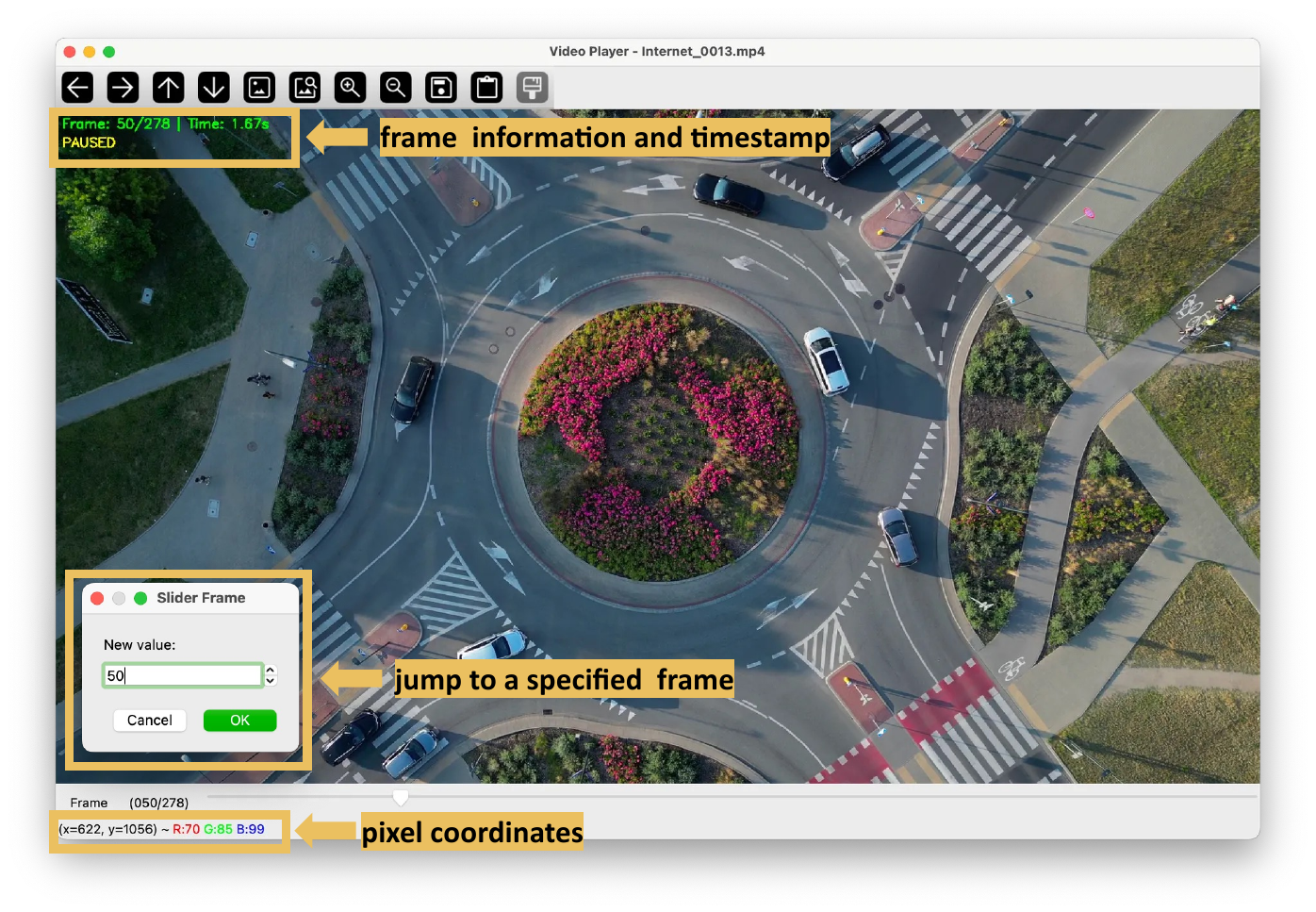}
    \caption{\textbf{Pixel-level measurement tool.}}
    \label{fig:internet_parsing}
\end{figure*}

\noindent\textbf{Pixel-level measurement tool.}
To read off pixel-space trajectories, we build a small annotation tool in Python using OpenCV.
The script takes as input the path to a video file and interactively queries the annotator for the correct frame rate.
It first shows the FPS detected by OpenCV, then allows the annotator to either accept it or manually enter a more reliable value (e.g., 24/25/30/60\,fps).
The tool verifies or manually counts the total frame number to avoid metadata errors, and reports basic properties such as resolution and duration.

After loading, the script launches an interactive player (\autoref{fig:internet_parsing}).
Each frame is overlaid with its index and timestamp, \texttt{ Frame: $k$/N | Time: $t_k$\,s}, and a play/pause status indicator.
A trackbar at the bottom allows direct jumping to any frame; dragging the slider or entering a frame index updates the display immediately.
Annotators can also step frame-by-frame with the left/right arrow keys or play the sequence in real time with the space bar.
In addition to these controls, we use a standard OpenCV mouse callback to display the current pixel coordinates $(x,y)$ of the cursor, which facilitates precise measurement of distances and object positions.

\noindent\textbf{(3) Measuring pixel-space kinematics.}
Given the verified frame rate $f$ (in fps), we set the time step as $\Delta t = 1/f$.
Annotators use the player to locate the relevant frames for each question:
for instantaneous quantities we jump to the frame closest to the target timestamp; for average quantities we select a sequence of frames spanning the interval of interest.
Along the main motion direction, we record
(i) the object’s pixel length $\mathbf S^{\mathrm{pixel}}$ (e.g., bumper-to-bumper for a car), and
(ii) the pixel coordinate $x_k^{\mathrm{pixel}}$ of the object’s reference point at discrete times
$t_k = k \Delta t$.

From these measurements we compute pixel-space kinematics using finite differences.
For a 1D trajectory $\{x_k^{\mathrm{pixel}}\}_{k=0}^{K}$, the (approximate) velocity and acceleration at time $t_k$ are
\begin{align*}
\mathbf V_k^{\mathrm{pixel}} 
  &\approx \frac{x_{k+1}^{\mathrm{pixel}} - x_{k}^{\mathrm{pixel}}}{\Delta t}, \\
\mathbf A_k^{\mathrm{pixel}} 
  &\approx \frac{x_{k+1}^{\mathrm{pixel}} - 2 x_{k}^{\mathrm{pixel}} + x_{k-1}^{\mathrm{pixel}}}{\Delta t^2}.
\end{align*}
When motion is measured in two image-plane directions, we apply the same formulas component-wise to $(x_k^{\mathrm{pixel}}, y_k^{\mathrm{pixel}})$.
In this way we obtain, for every prior object and inference target, the relevant pixel-space size $S^{\mathrm{pixel}}$, velocity $\mathbf V_k^{\mathrm{pixel}}$, and acceleration $\mathbf A_k^{\mathrm{pixel}}$.

\noindent\textbf{(4) Converting to real-world kinematics.}
Assuming planar motion (\autoref{sec:general_principles}), a single scalar scale factor $\gamma>0$ with units [world length / pixel] suffices along the motion direction.
Depending on which physical prior is available for a given clip, we estimate $\gamma$ via

\begin{equation*}
\gamma =
\begin{cases}
\dfrac{\mathbf S^{\mathrm{world}}}{\mathbf S^{\mathrm{pixel}}}, 
  & \text{if size prior is given} \\[0.6em]
\dfrac{\lvert \mathbf V^{\mathrm{world}}_{t_0} \rvert}{\lvert \mathbf V^{\mathrm{pixel}}_{t_0} \rvert},
  & \text{if velocity prior is given} \\[0.6em]
\dfrac{\lvert \mathbf A^{\mathrm{world}}_{t_0} \rvert}{\lvert \mathbf A^{\mathrm{pixel}}_{t_0} \rvert},
  & \text{if acceleration prior is given} \, .
\end{cases}
\label{eq:scale_factor}
\end{equation*}

where $t_0$ is the timestamp at which the prior is defined.

Once $s$ is determined, any kinematic quantity of the inference target can be expressed in world units by a simple rescaling:
\begin{align*}
(S_{\text{target}})^{\mathrm{world}} &= \gamma \, (S_{\text{target}})^{\mathrm{pixel}},\\
(\mathbf V_{\text{target}})^{\mathrm{world}}_{t_k} &= \gamma \, (\mathbf V_{\text{target}})^{\mathrm{pixel}}_{t_k},\\
(\mathbf A_{\text{target}})^{\mathrm{world}}_{t_k} &= \gamma \, (\mathbf A_{\text{target}})^{\mathrm{pixel}}_{t_k}.
\end{align*}
These world-space values are then used as the ground-truth priors and posteriors in \textsc{QuantiPhy}.

\subsection{Segmented Data}
Segmenting out target objects from the videos only changes the background and does not alter their original physical properties. Therefore, we reuse the original annotations of the videos for the segmented data without any modification.

\section{Vision-Language Models}
\label{appendix:models}

\begin{figure*}[ht]
    \captionsetup{name=Table}
    \centering
    \includegraphics[width=.88\linewidth]{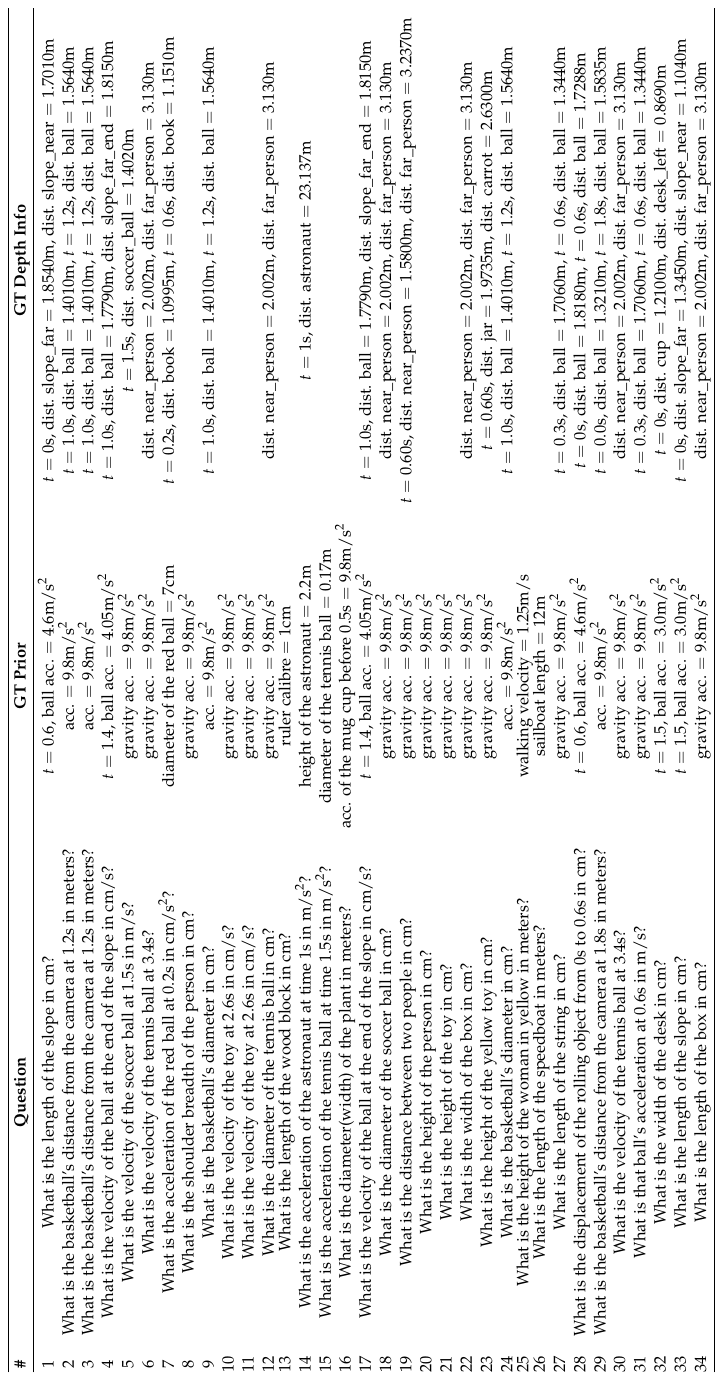}
    \caption{\textbf{Selected examples of text inputs and answers (part 1).} Question is the natural-language prompt presented to the VLM. GT Prior is the physical prior provided to the model. GT Depth Info is the depth annotation used for \texttt{3D} reasoning tasks. GT Posterior is the numeric ground-truth answer to the kinematic inference question. See \autoref{tab:record-schema-example} for detailed explanation. Raw Response shows the corresponding VLM output for each model, and Parsed Value shows the parsed value extracted from the Raw Response.}
    \label{tab:parsing-examples1}
    \end{figure*}

\begin{figure*}[ht]
    \captionsetup{name=Table}
    \centering
    \includegraphics[width=.86\linewidth]{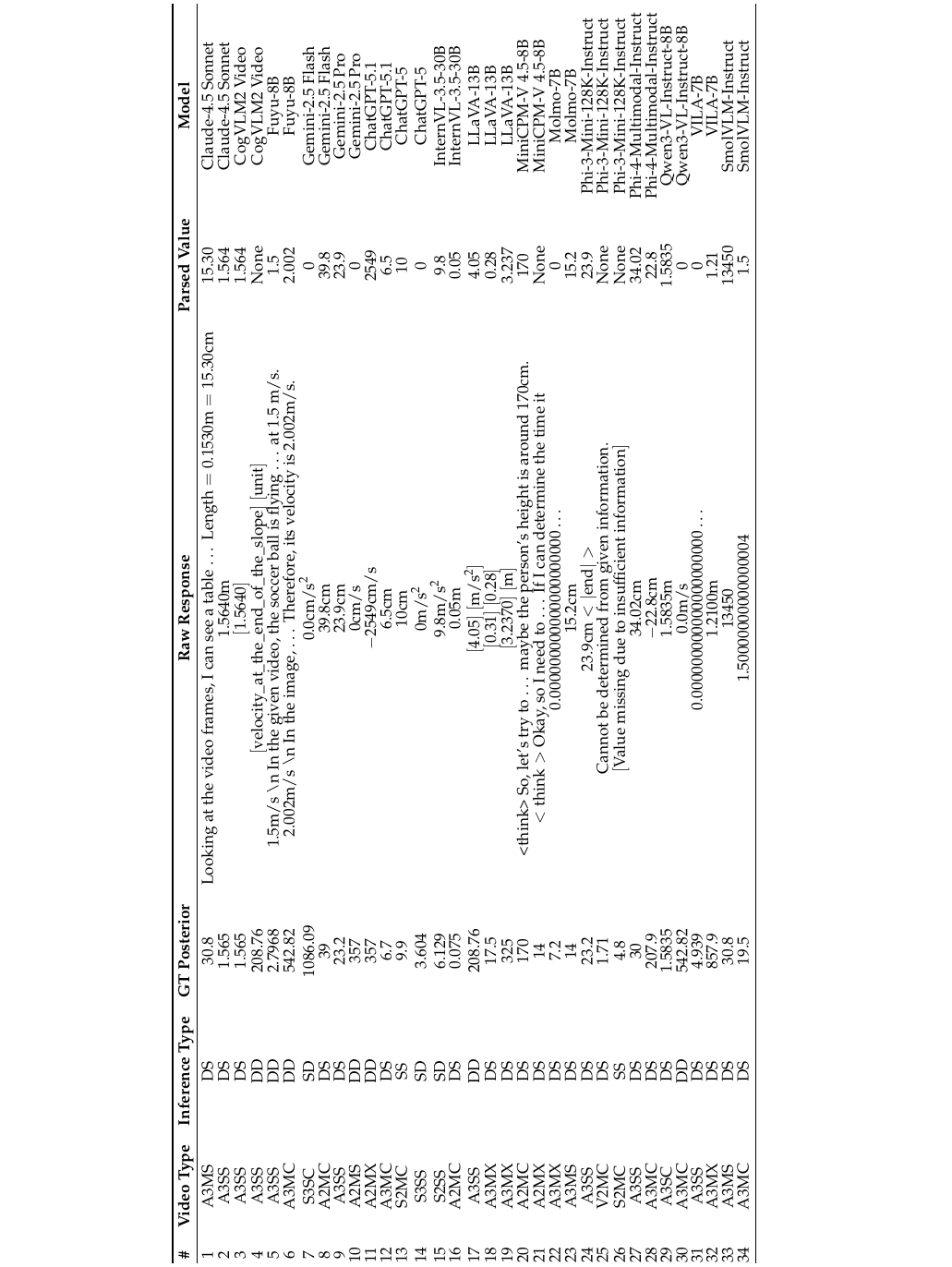}
    \caption{\textbf{Selected examples of text inputs and answers (part 2).} Question is the natural-language prompt presented to the VLM. GT Prior is the physical prior provided to the model. GT Depth Info is the depth annotation used for \texttt{3D} reasoning tasks. GT Posterior is the numeric ground-truth answer to the kinematic inference question. See \autoref{tab:record-schema-example} for detailed explanation. Raw Response shows the corresponding VLM output for each model, and Parsed Value shows the parsed value extracted from the Raw Response.}
    \label{tab:parsing-examples2}
\end{figure*}

We evaluate a diverse suite of 21 VLMs, spanning proprietary closed-source APIs, open-source models hosted via Replicate, and open-source models deployed by our own.

\noindent \textbf{Proprietary Models.} We select flagship models from four major providers, including standard multimodal capabilities and those utilizing inference-time reasoning.

\begin{itemize}
    \item \textbf{OpenAI.} We evaluate two models: \textbf{GPT-5.1}~\cite{OpenAIgpt5.1} (\texttt{gpt-5.1-2025-11-13}) and \textbf{GPT-5}~\cite{OpenAIgpt5} (\texttt{gpt-5-2025-08-07}), both are flagship multimodal models. The latter is a ``thinking model'' that leverages Chain-of-Thought (CoT) processing for complex reasoning tasks. Both are accessed via the OpenAI API, accepting video inputs as sequences of base64-encoded frames to enable temporal understanding.
    
    \item \textbf{Google.} We evaluate \textbf{Gemini 2.5 Pro}~\cite{google2025gemini25pro} (\texttt{gemini-2.5-pro}), a high-capacity model capable of processing large context windows of interleaved images and text, and \textbf{Gemini 2.5 Flash}~\cite{google2025gemini25flash} (\texttt{gemini-2.5-flash}), a lightweight variant optimized for low-latency tasks. Both models accept base64 inline data and are optimized for complex reasoning across modalities.
    
    \item \textbf{Anthropic.} We utilize \textbf{Claude Sonnet 4.5}~\cite{anthropic2025claude45sonnet} (\texttt{claude-sonnet-4-5-20250929}). This model accepts a structured list of content blocks (text and base64-encoded images). It is characterized by its detailed explanatory capabilities, often providing extensive textual rationale alongside numerical answers. (See \autoref{tab:parsing-examples2} for raw response examples).

    \item \textbf{xAI.} We evaluate \textbf{Grok 4.1 (Fast Reasoning)}~\cite{xai2025grok41}, a model that combines rapid inference with advanced reasoning capabilities, designed to handle complex multimodal tasks with reduced latency while maintaining high accuracy.

\end{itemize}

\noindent \textbf{Open-source Models.} We also evaluate 15 distinct open-source models, representing a spectrum of architectures, parameter sizes, and input modalities to systematically assess architectural variations and scaling effects.

\begin{itemize}
    \item We include \textbf{LLaVA-13B}~\cite{liu2023visual}, a foundational baseline combining a Vicuna LLM with a CLIP encoder; \textbf{VILA-7B}~\cite{lin2023vila}, pre-trained on interleaved image-text data; and \textbf{Qwen3-VL-8B}~\cite{bai2025qwen25vl}, the latest iteration of Alibaba's vision-language series.
    
    \item To examine scaling effects across model sizes, we deploy and evaluate the \textbf{Qwen3-VL-Instruct} series at three scales: \textbf{2B}, \textbf{8B}, and \textbf{32B} parameters~\cite{bai2025qwen25vl}, enabling direct comparison of performance improvements with increased capacity within a single architecture family.

    \item Similarly, we assess the \textbf{InternVL-3.5} series across three parameter scales: \textbf{2B}, \textbf{8B}, and \textbf{30B}~\cite{chen2024internvl}, providing additional insights into how architectural choices interact with model scale for vision understanding.

    \item  We evaluate \textbf{Fuyu-8B}~\cite{adept2023fuyu}, which utilizes a simplified architecture processing raw image patches without a separate visual encoder.
    
    \item We test \textbf{Molmo-7B}~\cite{allenai2025molmo} and \textbf{SmolVLM}~\cite{marafioti2025smolvlm} (1.6B), both designed for high-efficiency reasoning on consumer hardware.
    
    \item \textbf{Phi-4 Multimodal}~\cite{microsoft2025phi4multi} and the smaller \textbf{Phi-3-Mini-128K-Instruct}~\cite{microsoft2024phi3}, both are Microsoft suite and process text interleaved with image lists.
    
    \item Finally, to assess temporal analysis capabilities, we include \textbf{MiniCPM-V 4.5}~\cite{yao2024minicpmv} and \textbf{CogVLM2-Video}~\cite{wang2024cogvlm2}. Distinct from frame-based approaches, these models accept direct video file inputs for native temporal processing.
\end{itemize}

In total, our evaluation encompasses \textbf{21 models} (6 proprietary and 15 open-source), providing comprehensive coverage of the current state-of-the-art in multimodal reasoning across different scales, architectures, and deployment paradigms.

\section{Prompt Design} \label{appendix:prompt_design}
For quantitative evaluation, we use a constrained generation strategy designed for precise numerical outputs. The system prompt explicitly restricts the output space, instructing the model to ``provide ONLY the numerical answer with units'' and emphasizing ``No explanation or reasoning needed.'' The prompt structure includes the visual input, system instructions, and ground truth priors and/or depth info, followed by the specific query and a post-prompt reinforcement. (See \autoref{fig:case1} for the Text Prompt template).

To ensure reproducibility, we enforce deterministic generation where possible. We set \texttt{temperature=0} (greedy decoding) for all models supporting this parameter. For hosted models via Replicate where explicit temperature control is unavailable, we utilize the default inference parameters recommended by the model maintainers.

The input sequence for all models is programmatically structured as \texttt{[Video Frames][System Prompt][Ground Truth Prior and/or Depth Info][Question][Post-prompt]}. Detailed examples are provided in \autoref{tab:parsing-examples1}.

\begin{itemize}
    \item \texttt{[Video Frames]}. This segment contains the full sequence of frames extracted from the source video, base64-encoded. We normalize all videos to 480p resolution. Our preliminary exploration indicated that while lower spatial resolution (480p) has negligible impact on physics reasoning, temporal subsampling (dropping frames) significantly degrades the tracking of velocity and acceleration. Therefore, we prioritize {temporal fidelity} by retaining all frames from the source video. This approach also distinguishes our methodology from most prior works, which often prioritize spatial resolution at the expense of frame rate.
    
    \item \texttt{[System Prompt]}. This serves as the behavioral instruction, establishing the persona that ``You are an expert video analyst...'' We selected this formulation following a pilot study of five prompt variations on a subset of 15 videos, which revealed no significant difference and this persona yielded the relatively highest adherence to formatting constraints without altering reasoning accuracy.
    
    \item \texttt{[Ground Truth Prior and/or Depth Info]}. This includes physical constants and contextual priors (e.g., ``Given that acceleration $a = 9.8 m/s^2$''). For \texttt{3D} scenes, this also includes depth information (e.g., ``At $t=1.0s$, the distance of the ball to the camera is $1.779m$...'') for estimation.
    
    \item \texttt{[Question]}. The specific physics query (e.g., ``What is the total displacement in the y-axis?'').
    
    \item \texttt{[Post-prompt]}. A final instruction reinforcing the output format (e.g., ``Provide... ONLY the numerical answer with units'') to mitigate the tendency of models to generate verbose "Chain-of-Thought" explanations in the final output.
\end{itemize}

\section{Answer Retrieval and Parsing} \label{sec:parsing}

For each question, we query the model once. To ensure robustness against API instability or transient errors, we implement an automated retry mechanism: if a query fails (e.g., timeout or server error) or yields a non-parsable response, the request is re-submitted up to a maximum of five times.

To extract quantitative data from potentially noisy model outputs, we employ a hierarchical parsing function (\texttt{parse\_number}) designed to handle both concise and verbose responses. The logic proceeds as follows.

\begin{enumerate}
    \item \textbf{Exact Match Validation:} We first check if the raw response strictly matches a numerical format (with or without units). If the response is concise (containing only a value and a unit matching the requested physical quantity), the numerical value is extracted directly.
    
    \item \textbf{Delimiter Search:} If the response is verbose, we scan for explicit answer markers, including \texttt{``="}, \texttt{``Final Answer:"}, \texttt{``Answer:"}, \texttt{``=>"}, and \texttt{``:"}. If a delimiter is identified, we discard the preceding text and retain only the substring immediately following it. If multiple cases are shown, we take the last occurrence. 
    
    \item \textbf{Unit Sanitization:} The retained text is cleaned of common physical units (e.g., ``meters'', ``m/s'', ``kg'') to prevent string processing errors during numerical extraction.
    
    \item \textbf{Heuristic Extraction:} Finally, a regular expression is applied to the cleaned text to identify floating-point numbers. We also take absolute value of the numerical answer. Crucially, if multiple numbers are found, we extract the {last} valid number in the sequence. This heuristic assumes that in verbose Chain-of-Thought responses, the final conclusion is located at the end of the text.
    
    \item \textbf{Failure Handling:} If no valid number is identified after these steps, the response is recorded as a failure (parsed as \texttt{None}).
\end{enumerate}

This strict parsing pipeline is essential because our evaluation metric relies on exact numerical regression. If a model provides only a qualitative description (e.g., ``The ball is moving quickly'', ``Cannot be determined'', or API error) or fails to reach a numerical conclusion within the decoding limit, it is then treated as a failure.

While most models adhered to the format constraints (i.e., [number] [unit]), we observed that \textbf{Claude Sonnet 4.5}, \textbf{Fuyu-8B}, and \textbf{MiniCPM-V 4.5} frequently generated verbose, multi-sentence explanations despite instructions to the contrary. By targeting the post-delimiter text and the last numerical value, our parsing logic effectively retrieves the correct answer from these verbose outputs. Representative examples of raw responses and their parsed values are detailed in \autoref{tab:parsing-examples2}.

\section{Human Study Details.}

\begin{figure*}[t]
    \centering
    
    \begin{subfigure}{0.7\linewidth}
        \centering
        \includegraphics[width=\linewidth]{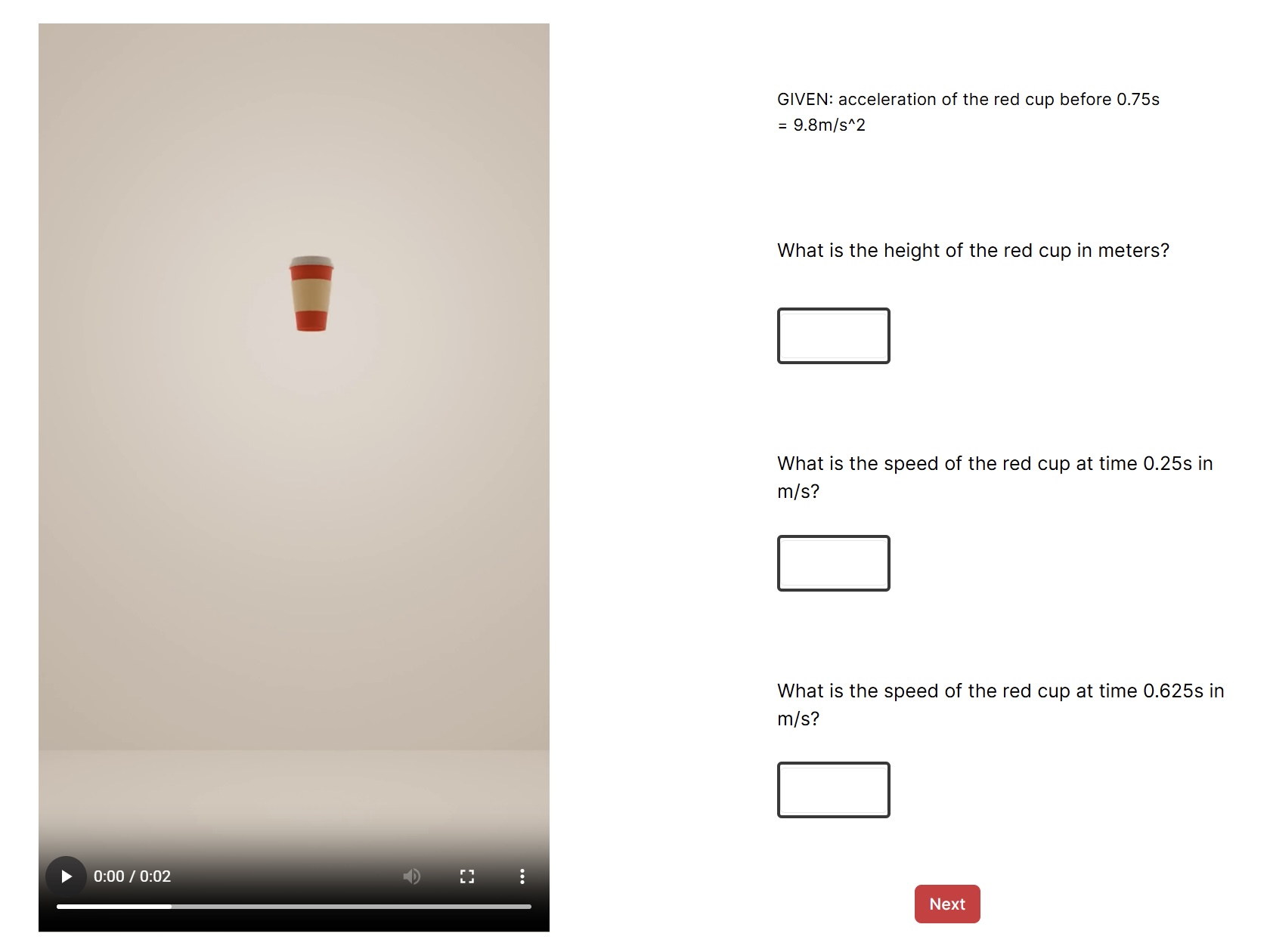}
        \caption{\texttt{2D} survey interface. Participants see the prior ground truth, question text, and a numeric answer box, and can replay or scrub the video.}
        \label{fig:ui-2d}
    \end{subfigure}
    \vspace{0.6em}
    
    \begin{subfigure}{0.8\linewidth}
        \centering
        \includegraphics[width=\linewidth]{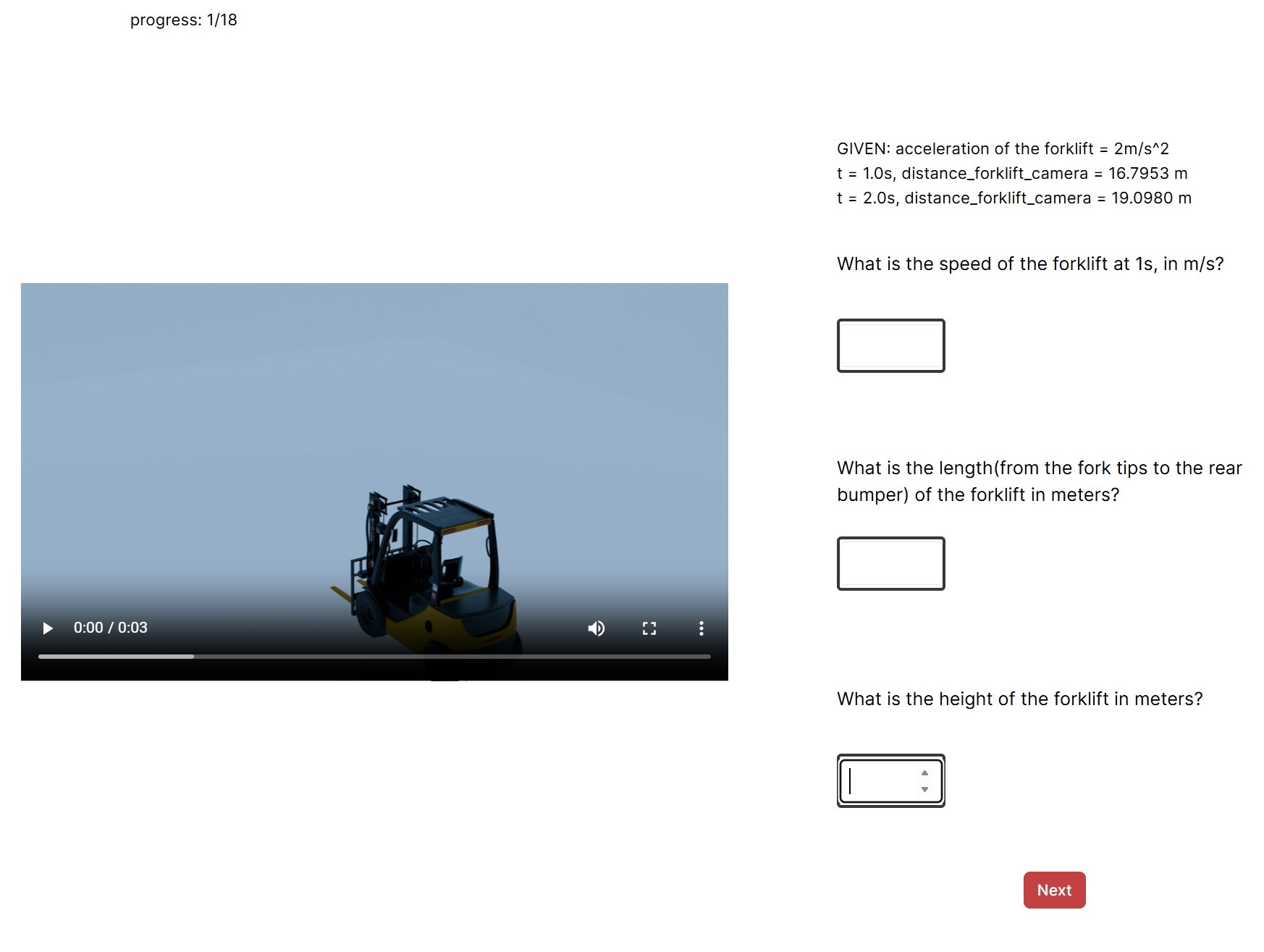}
        \caption{\texttt{3D} survey interface. The layout mirrors the \texttt{2D} condition, with additional depth prior ground truth.}
        \label{fig:ui-3d}
    \end{subfigure}
    
    \caption{\textbf{Human study interface.}
    Example screenshots of the \texttt{2D} (top) and \texttt{3D} (bottom) survey UIs.
    Both interfaces present the physical prior, quantitative question, and a numeric input field, while allowing participants to replay and scrub the video timeline.}
    \label{fig:ui-overview}
\end{figure*}

To contextualize model performance and establish an empirical upper bound on human quantitative physical reasoning, we conducted a survey study using the Gorilla Experiment Builder platform. The platform supported video presentation, question randomization, and structured data collection. Below, we describe the participant cohort, task construction, interface design, evaluation methodology, and resulting performance trends.

\subsection{Participants}
Participants were recruited from a mix of undergraduate, graduate, and PhD researchers, including individuals with advanced training in various fields including engineering, physics, and mathematics. This allowed us to approximate both typical human performance and an expert-level upper bound. Participants were informed that they could freely choose their reasoning strategy---including intuitive estimation, visual approximation, or explicit calculation---in order to reflect natural human reasoning rather than enforce a prescribed computation protocol.

Our survey is divided into two versions: a \texttt{2D} survey and a \texttt{3D} survey. Both the \texttt{2D} and \texttt{3D} surveys were completed by multiple participants, including several with substantial technical backgrounds. A subset of participants completed both tasks, enabling direct cross-dimensional comparisons of individual consistency.

\subsection{Task Construction and Experimental Design}
Following the 36 fine-grained video categories defined in \autoref{subsec:video_type}, we organized the full dataset into these category units for both model and human evaluation. Each category contains a small pool of representative videos, from which stimuli were sampled during the experiment.

Participants were randomly assigned to either the \texttt{2D} survey or the \texttt{3D} survey. Each participant viewed 18 videos, one sampled per category within their assigned dimensionality. Each video was followed by 1--3 quantitative kinematic questions, with the same priors and task formulations used in our VLM evaluation (e.g., estimating acceleration magnitude, inferring relative size changes, or recovering object velocity).

The UI interface was intentionally designed to be concise and unobtrusive. For each trial, participants were presented with:
\begin{itemize}
\item A video with standard playback controls;
\item The physical prior ground truth (\texttt{A/S/V}) and/or corresponding depth information for \texttt{3D} videos;
\item The quantitative question texts;
\item Numeric input boxes (numbers only).
\end{itemize}

Participants could freely replay, pause, and scrub the video timeline. Scrubbing resolution was restricted to whole seconds to match the temporal information available to VLMs, which process videos solely as pixel sequences without access to exact timestamps. The interface permitted unrestricted video replay to minimize memory effects and ensure that both humans and VLMs operated over comparable visual evidence.

\begin{figure}[t]
    \centering
    \includegraphics[width=0.95\linewidth]{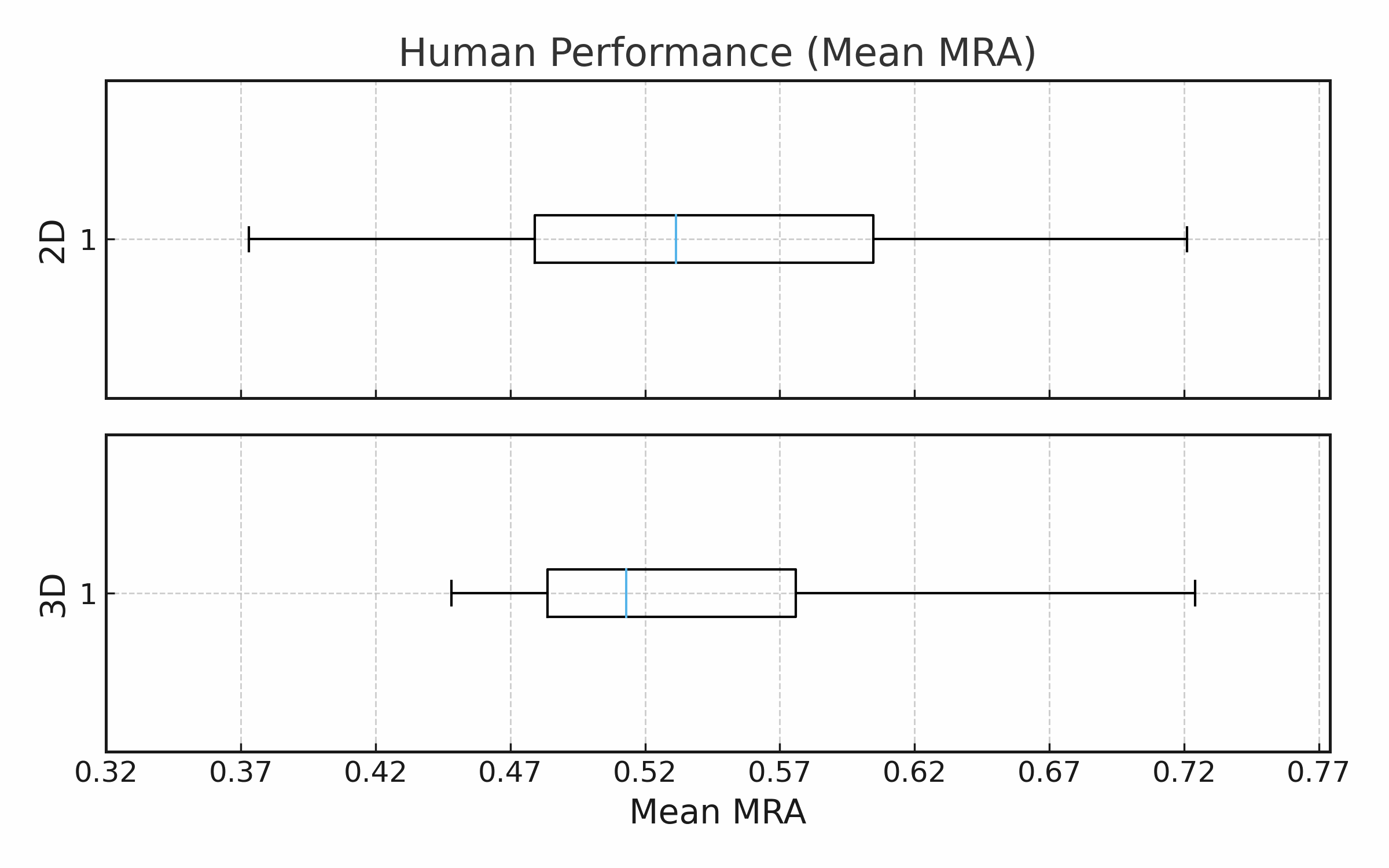}
    \caption{\textbf{Distribution of human quantitative reasoning performance.}
    Horizontal boxplots summarize participant-level mean MRA scores for the \texttt{2D} (top) 
    and \texttt{3D} (bottom) survey conditions.}
    \label{fig:human-boxplot}
\end{figure}

\subsection{Evaluation Metric}
Human responses were evaluated using the same MRA metric introduced in Section~\ref{sec:evaluation}. 
For each individual question, we compute MRA by averaging accuracy over ten relative-error tolerance thresholds, yielding a smooth, threshold-agnostic measure of quantitative precision. 
After computing an MRA score for every answered question, we aggregate performance at the participant level by taking the mean MRA across all questions completed by that participant. 
Figure~\ref{fig:human-boxplot} presents participants performance for the \texttt{2D} and \texttt{3D} surveys, respectively.

\subsection{Results and Observations}
We observe a few findings from the results.

\noindent\textbf{Strong cross-dimensional consistency.}
Across participants who completed both tasks, MRA scores for \texttt{2D} and \texttt{3D} surveys were highly correlated. High-performing participants in \texttt{2D} almost universally remained high-performing in \texttt{3D}, suggesting that human physical intuition is stable and transferrable across dimensional modalities.

\noindent\textbf{Human upper bound substantially exceeds VLM performance.}
Top human participants achieve MRA = \textbf{0.721} in \texttt{2D} and  MRA = \textbf{0.724} in \texttt{3D}, showing better performance than the evaluated VLMs. Although the average human performance is not higher than model performance, the human upper bound remains far above current VLM capabilities. This gap is particularly notable given the modest sample size of our study; even with limited data, humans exhibit strong physical reasoning competence that remains challenging for contemporary models.

\noindent\textbf{Practical implications.}
These findings indicate that while VLMs may approximate average-level human intuition in some settings, they remain far from achieving human-like precision or matching expert-level reasoning. The human study therefore provides a meaningful benchmark for assessing the gap between current VLM capabilities and the aspirational goal of physically grounded, human-level visual reasoning.

\section{Sketchfab Model Sources}

In \autoref{tab:sketchfab_models}, we list the Model Name, Author, and Model ID for all Sketchfab models used in the Blender-generated videos in our dataset.
\onecolumn
\begin{longtable}{|p{6cm}|p{4cm}|p{6cm}|}
\caption{Sketchfab 3D Models} \label{tab:sketchfab_models} \\
\hline
\textbf{Model Name} & \textbf{Author} & \textbf{Model ID} \\
\hline
\endfirsthead

\multicolumn{3}{c}{\tablename\ \thetable\ -- \textit{Continued from previous page}} \\
\hline
\textbf{Model Name} & \textbf{Author} & \textbf{Model ID} \\
\hline
\endhead

\hline
\multicolumn{3}{r}{\textit{Continued on next page}} \\
\endfoot

\hline
\endlastfoot
Supermarket trolley broken & kreems & 7f460877380349f8886280a596253034 \\
\hline
supermarket 1 & amogusstrikesback2 & ca1def2b7be544068def3cec5852c67e \\
\hline
Pool Table (Animation) & Yanez Designs & 0f2ae181a2dd4b00a6ec25073692037f \\
\hline
Bowling Pack (Bowling Pins \& Ball) & EverZax & 52743c4714c14211ac71d2fe1e5c8da3 \\
\hline
Bowling Club & tiunov.se & 0b8fae45fcda4fe78f93bb2a899401a6 \\
\hline
Elephant & planeta-elefante & f8778fc3d161481abba7ec23a8ddd1e8 \\
\hline
Walking Astronaut & Unknown Animaker & d9062a2003df422abdafdc02afdac085 \\
\hline
Mr Man Walking & Instinto Ideal Studio & 98ccac2b0e2845789b6f789978ca06ed \\
\hline
Jupiter \& satellites & Sakado & 379fd77b970c4821898c05c483913dec \\
\hline
Model 92A - Great White Shark & DigitalLife3D & 702e7b53637f4ded9ca479a8124e810d \\
\hline
Model 95A - Adult Leatherback Sea Turtle, "Mac" & DigitalLife3D & 4974c93644a24da280fde68cba74a12d \\
\hline
Model 69A - Striped Bass & DigitalLife3D & 35be3af9a7c4441c98109e5562d36c09 \\
\hline
Underground Parking Lot & Janis Zeps & 0ad5c221525b4bbba3a164c6235d28b8 \\
\hline
Airport Car & Jungle Jim & 29d1c6260b134a3baf5231f34de1b24f \\
\hline
Airport Catering Truck & rwy00 & 289f4e2cfa3f4722b0476b1fc37681d8 \\
\hline
Towel & TheSpacePunk & 6edc341457e44603ab351470ab800493 \\
\hline
Plain Mug & LightSwitch & 19c8fe5702b544d0a1409d3dac1cf90e \\
\hline
Flour ARIDLL & fwild & d40db94d92ab4e7a82a1c199312f3985 \\
\hline
Milk & Multipainkiller Studio & 5e8d71045e2040ba8f6619d86d204cf5 \\
\hline
Apple [Scan] & hoxsvl.scan & facb1aa6928c4f8f82d87a019b9f134e \\
\hline
PACKAGING EMPAQUE DE HUEVOS & willinando1w08 & cd51c5d25bbd4370881aebd3648cfe8c \\
\hline
Kitchen Pack & GRIP420 & a513f2e85bc94ae5b6d8cbd74909e3c4 \\
\hline
Kitchen Tools & anybody & 417e3873fcb34f7ab9744506d7bcc838 \\
\hline
Coffee Shop Cup & David Zerba & 37e6805f2b7a4158a1d61fe75f8e2a33 \\
\hline
Cup & Dmytro Nikonov & 7d450bb714034fceaa7b59a0e564f46b \\
\hline
Bird Animations Alex & ahitch3 & 081fa7f0cfd649b9b07babb4c619acc7 \\
\hline
Dove Bird Rigged & FourthGreen & 6b91be2a28fb4404a2d57d5ca98bd4dc \\
\hline
POLICE CAR - Belgian VW Transporter & Mickael Boitte & 35eaf9505e13464385404402ad865508 \\
\hline
Jo on Bike - Rigged \& Animated & NEEEU Spaces GmbH & 36ee5344e81149858a664cde9f98e835 \\
\hline
Simple Factory Scene & Pickeri & 804fbe0cd1fa44fa9ca86ae42c82d63d \\
\hline
Car & Paulius & a619dd25a6c04af0b2d8730aa1cb058b \\
\hline
Basketball Court & tiunov.se & 5faad7b528124907ab82732ed0c6b743 \\
\hline
Basketball NBA Championship Official Trophy & johnnokomis & 04f6f1135ffb48749c43c9c20c75fc19 \\
\hline
Bouncing Basketball & Maurice Svay & b8731a2fda6849c9a164d1966dc16ff8 \\
\hline
Zen Japanese Tea House with Go Board & winters810 & a1e95e5efee349a693f30eda32401aef \\
\hline
Chinese Chess & chung\_the\_artist & 7ad0f4f0ebbc455d9e9f829c956dda80 \\
\hline
paper airplane & vesicalsnail & 0967ab4a9c654a569a13ea1f8d9dca0c \\
\hline
3d soccer ball & BlenderMaster & 0400146e0e3c4d8f8b57bfa06d7dfb4c \\
\hline
Cardboard Box & Pricey1600 & 4e622ef1a09c43e28a49d9fa37f9eeee \\
\hline
Tennis Ball Low-poly PBR & MaX3Dd & e5c2b0e5860549acaa2dfe8b764d5f94 \\
\hline
Old Camera Bag .::RAWscan::. & Andrea Spognetta (Spogna) & 788f8b75874f417ebde498ffd231410c \\
\hline
Animated ROBOT SDC & SDC PERFORMANCE & 3d127f327a6c4033a32b810b5fb071ed \\
\hline
Gift Box & local.yany & 83296611584143a3afad6c0d0c0a4227 \\
\hline
Sci-fi Box & Igor\_K. & ca415724f32043489fcab2ec74582619 \\
\hline
Cat Walk & LostBoyz2078 & 915680200c064815bba75e008ba9efb5 \\
\hline
Deer walk & LostBoyz2078 & 229ba6ba0d1e4811ab89382f74601e16 \\
\hline
Horse & kenchoo & 86d47bdcd5ab41238ba44547e4d21f9c \\
\hline
Horse Walk & Amitesh Nandan & 93b53ddcec414592842753d1819f3133 \\
\hline
Rabbit Rigged & FourthGreen & e7213589744d436b9d96e2dbb31198a5 \\
\hline
phoenix bird & NORBERTO-3D & 844ba0cf144a413ea92c779f18912042 \\
\hline
First Aid DX2 - 300 Followers Celebration & re1monsen & c5ddfa9e6309403083bbce60bdcc3d71 \\
\hline
Nathan Animated 003 - Walking 3D Man & Renderpeople & 143a2b1ea5eb4385ae90a73657aca3bc \\
\hline
Chernovan Nemesis & Swiss\_Fox & c6c91c73e93444f4b72d6c24db778e73 \\
\hline
Dragon Fly & LostBoyz2078 & 1443a7efe5d5450b8db4c15d8ff5c343 \\
\hline
Borboleta Azul - Butterfly & Lancaster Modelagem 3D & ab9192b6bc8f49e3baed63e984c7073a \\
\hline
Blue Whale - Textured & Bohdan Lvov & d24d19021c724c3a9134eebcb76b0e0f \\
\hline
jellyfish & yanix & d06a5a553fe641ab92f720527b2278f3 \\
\hline
Koi Fish & 7PLUS & 236859b809984f52b70c94fd040b9c59 \\
\hline
Running Raccoon Animation & Santrez & bfce4d4815234c39bcf012352e52c27e \\
\hline
(FREE) Cyberpunk Hovercar & Lionsharp Studios & 3205b1075bb44ffc826bce0c2a04d74c \\
\hline
White Eagle Animation Fast Fly & GremorySaiyan & 30203bf39e5145f19c79e83c550139d3 \\
\hline
\end{longtable}

\end{document}